\pdfoutput=1
\documentclass{article}

\PassOptionsToPackage{numbers, sort&compress}{natbib}

\usepackage[main, final]{neurips_2026}

\usepackage{graphicx}
\usepackage[dvipsnames,table]{xcolor}
\usepackage{amsmath}
\usepackage{amssymb}
\usepackage{colortbl}
\usepackage{makecell}
\usepackage{hhline} %
\usepackage{multirow,multicol}
\usepackage{subcaption}
\usepackage{wrapfig,lipsum,booktabs}
\usepackage{bbm}
\usepackage{pifont}
\usepackage{lipsum}
\usepackage{tabu}

\usepackage[utf8]{inputenc} %
\usepackage[T1]{fontenc}    %
\definecolor{citecolor}{rgb}{0.12,0.49,0.85}
\usepackage[pagebackref,breaklinks,colorlinks,citecolor=citecolor]{hyperref}       %
\usepackage{url}            %
\usepackage{booktabs}       %
\usepackage{amsfonts}       %
\usepackage{nicefrac}       %
\usepackage{microtype}      %

\usepackage{enumitem}

\newlength{\affilwd} %
\newcommand{\red}[1]{\textcolor{red}{#1}}

\newcommand{\gray}[1]{\textcolor{gray}{#1}}

\def\eg{\emph{e.g.}}
\def\ie{\emph{i.e.}}

\def\etal{\emph{et al.}}

\def\ours{MOSS }

\newcolumntype{L}[1]{>{\raggedright\let\newline\\\arraybackslash\hspace{0pt}}m{#1}}
\newcolumntype{C}[1]{>{\centering\let\newline\\\arraybackslash\hspace{0pt}}m{#1}}
\newcolumntype{R}[1]{>{\raggedleft\let\newline\\\arraybackslash\hspace{0pt}}m{#1}}

\usepackage{amsmath,amsfonts,bm}

\def\eqref#1{equation~\ref{#1}}

\def\1{\bm{1}}

\def\mF{{\bm{F}}}

\def\mS{{\bm{S}}}

\DeclareMathAlphabet{\mathsfit}{\encodingdefault}{\sfdefault}{m}{sl}
\SetMathAlphabet{\mathsfit}{bold}{\encodingdefault}{\sfdefault}{bx}{n}

\title{Exploring Multi-Order Self-Similarity \\for Motion Understanding}

\author{%
  Manjin Kim$^{1}$\thanks{Equal contribution.} \hspace{7mm} Heeseung Kwon$^2$\footnotemark[1] \hspace{7mm} Karteek Alahari$^3$ \hspace{7mm} Minsu Cho$^{1,2}$   \vspace{3mm}\\ 
  \settowidth{\affilwd}{\tt http://cvlab.postech.ac.kr/research/MOSS}%
  \makebox[\affilwd][s]{$^1$POSTECH\hfill$^2$RLWRLD\hfill$^3$Inria\thanks{Univ.\ Grenoble Alpes, Inria, CNRS, Grenoble INP, LJK, France}\hphantom{\textsuperscript{\normalfont\textdagger}}}\vspace{3mm}\\
  {\tt \url{http://cvlab.postech.ac.kr/research/MOSS}}
}

\begin{document}

\maketitle


\begin{abstract}
Space-time self-similarity (STSS), which captures visual correspondences across frames, provides an effective way to represent temporal dynamics for video understanding.
In this work, we explore higher-order STSS and demonstrate how STSSs at different orders reveal distinct aspects of these dynamics.
We then introduce the Multi-Order Self-Similarity (MOSS) module, a lightweight neural module designed to learn and integrate multi-order STSS features.
It can be applied to a wide range of motion-centric tasks with only marginal computational cost and memory overhead.
Extensive experiments on video action recognition, motion-centric video VQA, and robotic tasks consistently demonstrate substantial improvements, validating the broad applicability of MOSS as a general motion modeling module.
\end{abstract}
    

\section{Introduction}
\label{sec:intro}

The real world is dynamic, not static. The most prominent characteristic that distinguishes videos from images lies in the presence of such temporal dynamics, \ie, changes of visual patterns over time. 
Without a proper grasp of those features, \eg, motion information, video understanding models often become biased toward static contextual cues, limiting their generalization in out-of-context scenarios~\cite{devias,choi2019can,hat,li2018resound}.
Scaling alone does not resolve this; even recent video multimodal large language models (MLLMs) still fall short in fine-grained temporal understanding~\cite{yashima2026remora,imam2025can,wang2025videostr,shi2026causality}, calling for representations that explicitly capture temporal dynamics.


\begin{figure*}[t!]
    \centering
    \begin{subfigure}[t]{0.55\linewidth}
        \centering
        \includegraphics[width=\linewidth]{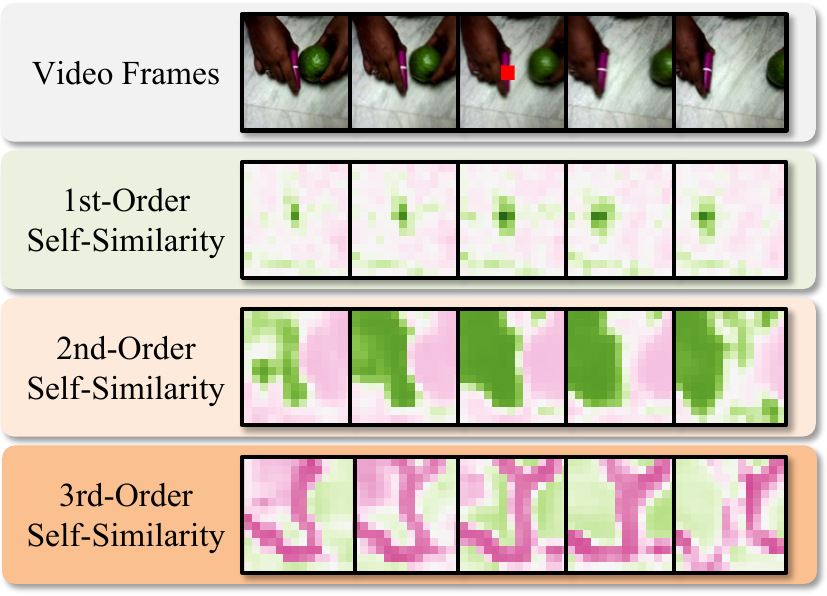}
        \vspace{-3mm}
        \subcaption{STSS maps across different orders.}
        \label{fig:teaser_a}
    \end{subfigure}
    \hfill
    \begin{subfigure}[t]{0.425\linewidth}
        \centering
        \includegraphics[width=\linewidth]{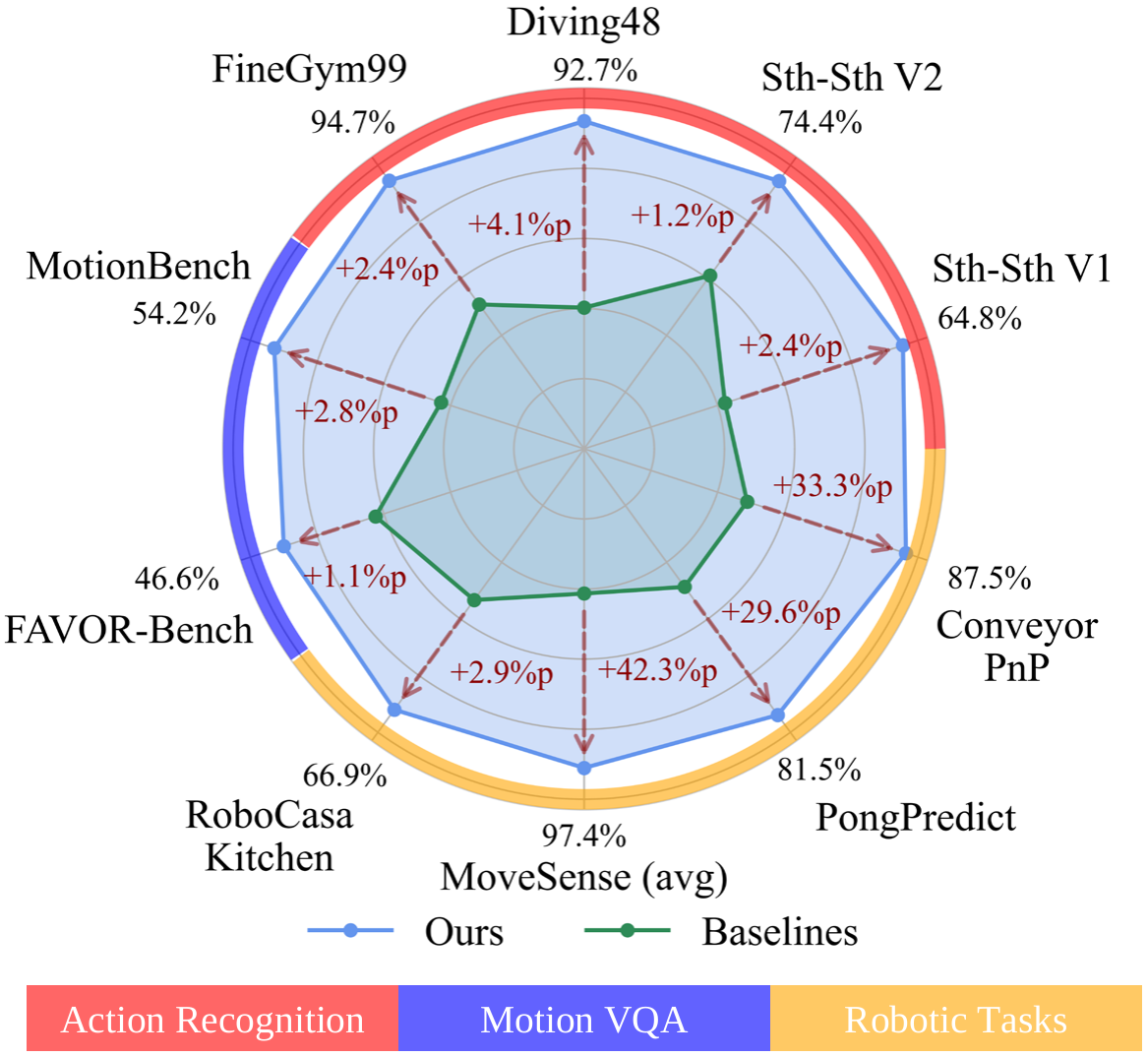}
        \vspace{-3mm}
        \subcaption{Overall improvements.}
        \label{fig:teaser_b}
    \end{subfigure}
    \vspace{1mm}
    \caption{\textbf{Multi-order space-time self-similarities (STSS) for motion understanding.} (a) Given a \textcolor{red}{red} query, the \mbox{1st-}, \mbox{2nd-}, and 3rd-order STSS effectively identify \textit{motion flows, motion segments}, and \textit{the layout of motion segments}, respectively. (b) Leveraging multi-order STSS captures diverse spatio-temporal dynamics, resulting in significant performance improvements across video action recognition, motion-centric video VQA, and robotic tasks.}
    \label{fig:teaser}
\end{figure*}

Temporal dynamics in general can be represented as structural patterns of how visual elements interact with each other in space and time. While the most popular and explicit form of it would be motion fields or optical flows~\cite{dosovitskiy2015flownet,ng2018actionflownet,sun2018pwc,raft}, the seminal work by Shechtman and Irani~\cite{shechtman2005space,selfsimilarity} has shown that the space-time self-similarity (STSS), \ie, a correlation volume over a local window of a video in space and time, effectively reveals temporal dynamics suppressing irrelevant appearance variations. 
Recent studies~\cite{bian2022learning,rsa,selfy,son2022contrastive,corrnet,atm} also demonstrate that learning self-similarity features on latent feature maps enables neural networks to understand motion in videos better, improving performance.

In this work, we explore higher-order self-similarities, \eg, \textit{self-similarity of self-similarity} in space and time as the 2nd-order STSS, and investigate what kinds of distinct temporal dynamics emerge.
Self-similarity operations reveal structured correlation patterns~(Fig.~\ref{fig:teaser_a}); given a base feature map describing appearance for each position in space and time, the conventional 1st-order STSS computes similarities of appearances, revealing \textit{motion flows}, \eg, the leftward translation of the queried pen across frames (2nd row).
In the same vein, given the 1st-order STSS map describing motion flows, the 2nd-order STSS computes similarities of motion, recognizing \textit{motion segments}; the 2nd-order STSS maps highlight regions of both the hand and pen that share similar motion patterns regardless of their distinct appearances (3rd row).
Extending this hierarchy, the 3rd-order STSS captures similarities over motion segments, reflecting the \textit{layout of motion segments} (4th row).
This hierarchical progression to higher-order STSS provides useful cues for comprehensive video analysis in complex scenarios.

From these insights, we design a novel neural module, dubbed \ours(\textbf{M}ulti-\textbf{O}rder \textbf{S}elf-\textbf{S}imilarity), that learns and integrates STSS representations across multiple orders into holistic motion features.
The proposed module is lightweight and can be easily integrated into existing video architectures, enhancing temporal modeling capabilities across various domains (Fig.~\ref{fig:teaser_b}).
We first evaluate our method on diverse action recognition benchmarks, \ie, Something-Something V1 \& V2~\cite{goyal2017something,mahdisoltani2018effectiveness}, Diving48~\cite{li2018resound}, and FineGym~\cite{shao2020finegym}, achieving significant performance gains with marginal computation and memory overhead.
We further incorporate MOSS into video MLLMs, improving fine-grained motion understanding and yielding substantial gains on MotionBench~\cite{motionbench} and FAVOR-Bench~\cite{favorbench}.
Finally, we incorporate MOSS into Vision-Language-Action (VLA) models to enhance their temporal perception in robotic settings.
We validate its effectiveness on the RoboCasa-Kitchen benchmark~\cite{robocasa} and further evaluate it on three real-world robotic tasks, \textit{MoveSense}, \textit{PongPredict}, and \textit{Conveyor Pick-and-Place}, which involve interactions in dynamic environments, showing improved temporal understanding and more robust interaction performance.

Our contributions are summarized as:
\vspace{-2mm}
\begin{itemize}[leftmargin=*]
     \item We provide an in-depth analysis of high-order space-time self-similarities and find that combining 1st- and 2nd-order STSS yields an effective representation for temporal dynamics.    
    \item We propose MOSS, a novel lightweight neural module that learns integrated STSS features at multiple orders for comprehensive temporal understanding.
    \item We demonstrate the versatility of MOSS across diverse motion-related tasks, including video action recognition, motion-centric video VQA, and robotic tasks.
\end{itemize}


\section{Related Work}
\label{sec:related_work}

\noindent\textbf{Self-Similarity in Video Understanding.}
The pioneering work by Shechtman and Irani~\cite{shechtman2005space,selfsimilarity} shows that self-similarity, \ie, a correlation over a local window of an image or a video in space and time, reveals structural layouts and suppresses appearance variations, and has been leveraged by Junejo \etal~\cite{junejo2010view,junejo2008cross} to develop robust temporal descriptors for action recognition under view changes.
Recently, several methods~\cite{motionsqueeze,selfy,corrnet} employ self-similarities within a video clip to learn motion features.
CorrNet~\cite{corrnet} and MSNet~\cite{motionsqueeze} obtain short-term motions via cross-frame similarities, SELFY~\cite{selfy} learns STSS-based bi-directional motion features, and ATM~\cite{atm} combines STSS with frame-wise differences.
In parallel, diverse transformer architectures model spatio-temporal correlations through attention mechanisms~\cite{vivit,mvit,uniformerv2,uniformer,videoswin}, and some methods incorporate STSS into attention for better video representation learning~\cite{rsa,structvit,kwon2025lightweight}.
To the best of our knowledge, no prior work has explored higher-order STSS, \ie, self-similarity of self-similarity in space-time. We investigate higher-order STSS and demonstrate its effectiveness for modeling temporal dynamics.

\noindent\textbf{Temporal Modeling in Video Action Recognition.}
Video action recognition has long served as a fundamental task for evaluating the temporal modeling capability of video models.
Early approaches capture temporal dynamics via explicit motion representations such as optical flows~\cite{simonyan2014two} or through 3D convolutional networks~\cite{i3d,feichtenhofer2020x3d,feichtenhofer2019slowfast,c3d,tran2018closer}, while more recent methods extend vision transformers~\cite{vit} to the video domain using spatio-temporal attention~\cite{vivit,timesformer,mvit,uniformer,mvitv2,videoswin}.
With the emergence of large vision foundation models~\cite{clip,evaclip,sun2024eva,siglip}, several methods further explore efficient image-to-video transfer by adapting frozen image encoders either with lightweight temporal modules~\cite{omniclip,stadapter,dualpath,m2clip,aim} or side networks~\cite{evl,dist,side4video}, avoiding costly end-to-end fine-tuning.
In this work, we follow this paradigm and demonstrate that multi-order STSS features enable effective image-to-video transfer, achieving strong performance with marginal computational overhead.

\noindent\textbf{Temporal Understanding in Video MLLMs.}
Recent advances in multimodal large language models (MLLMs)~\cite{internvl25, qwen2, videollava, videollama3} have extended large language models to video processing, enabling comprehensive video understanding beyond video action recognition.
These models demonstrate strong capabilities in event-level and story-level understanding~\cite{videomme, mvbench, lvbench, nextqa}, benefiting from large-scale video-text pretraining and powerful language reasoning abilities.
However, recent studies~\cite{motionbench, favorbench} show that video MLLMs still struggle with fine-grained motion, \eg, object motion or repetition counting.
In this work, we integrate the MOSS module into video MLLMs and enhance motion understanding with minimal computational and training overhead.

\noindent\textbf{Temporal Understanding in VLAs.}
Vision-Language-Action (VLA) models extend vision-language models by equipping them with action generation heads to learn robot action policies~\cite{bjorck2025gr00t,black2024pi_0,intelligence2025pi_}. 
Most existing VLAs operate on single-frame visual observations, which inherently limit their ability to interact with dynamic environments. 
Recently, a few VLAs~\cite{jang2025contextvla,li2026matters,team2024octo} incorporate multi-frame sequences to learn long-term policies, yet their ability to capture temporal dynamics has not been systematically examined. 
We here introduce several tasks for evaluating temporal understanding in these types of VLAs and show that integrating the proposed MOSS module significantly improves their ability to reason about temporal dynamics in robotic manipulation.
\section{Our Approach}

We first revisit the concept of \textit{space-time self-similarity} (STSS), then extend it to higher orders, and discuss the distinct information captured at each order.
We then introduce our MOSS module that exploits the STSS representations across the orders and integrates them into holistic motion features.

\subsection{Revisiting Space-Time Self-Similarity (STSS)}
\label{sec:revisiting_stss}
\noindent \textbf{STSS Transformation.}
Self-similarity~\cite{selfsimilarity} reveals geometric structures of correlations between visual entities while suppressing their visual content, allowing us to understand relational patterns in visual data.
In the video domain, STSS computes pair-wise correlations between a query and its local spatio-temporal neighbors, describing spatio-temporal dynamics of the query across frames~\cite{selfy,kim2026tempered}.
We define an STSS transformation function $f$ that maps input feature maps to a 6D tensor as,
\begin{align}
    f: \mathbb{R}^{T \times H \times W \times C} \rightarrow \mathbb{R}^{T\times H \times W \times L \times U \times V},
\end{align}
where $(T, H, W)$ are the spatio-temporal dimensions, and $(L, U, V)$ denote the size of the local spatio-temporal window.
Given input feature maps of $T$ frames \mbox{$\mathbf{\mF} \in \mathbb{R}^{T\times H \times W \times C}$,} each element of the STSS tensor \mbox{$\mathbf{\mS} = f(\mathbf{F})\in \mathbb{R}^{T \times H \times W \times L \times U \times V}$} is computed as,
\begin{align}
    \mathbf{\mS}_{t,h,w,l,u,v} = \phi(\mathbf{\mF}_{t,h,w}, \mathbf{\mF}_{t+l,h+u,w+v}),
\end{align}
where $(t, h, w)$ are the 3D coordinates of a query and $(l,u,v)$ are the offsets of the local spatio-temporal window of the query, with $(l,u,v) \in [-\lfloor \frac{L}{2}\rfloor, \lfloor \frac{L}{2}\rfloor] \times [-\lfloor \frac{U}{2} \rfloor, \lfloor \frac{U}{2} \rfloor] \times [-\lfloor \frac{V}{2} \rfloor, \lfloor \frac{V}{2} \rfloor]$. The function $\phi$ computes the similarity, \eg, cosine similarity, between two feature vectors.

\begin{figure*}[t]
    \centering
    \begin{subfigure}{0.645\linewidth}
        \includegraphics[width=\textwidth]{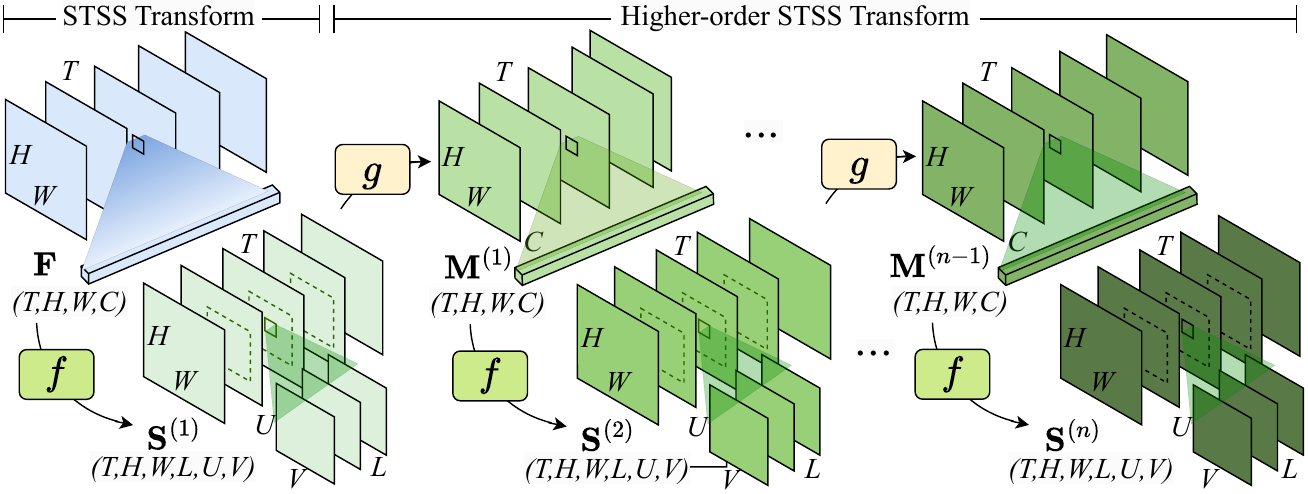}
        \caption{STSS \& Higher-order STSS Transformation}
    \label{fig:main_a}
    \end{subfigure}%
    \hspace{0.02\linewidth}%
    \begin{subfigure}{0.33\linewidth}
        \includegraphics[width=\linewidth]{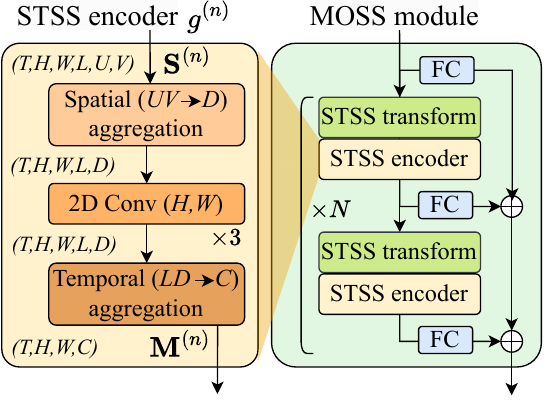}
        \subcaption{MOSS module}
    \label{fig:main_b}
    \end{subfigure}%
    \caption{\textbf{High-order STSS transformation \& Multi-Order Self-Similarity (MOSS) module.} (a) depicts a recursive process for high-order STSS transformation. (b) shows the overall process of the MOSS module for learning multi-order STSS representations.} 
    \label{fig:main}
\end{figure*}

\noindent \textbf{Characteristics of STSS.}
The STSS tensor $\mathbf{S}$ effectively captures appearance-based correspondences across different frames, presenting diverse temporal information throughout the video sequence.
For $l=0$, it is spatial self-similarity~\cite{selfsimilarity} showing object layouts or similar objects within the same frame.
For $l \ne 0$, it becomes spatial cross-similarity between two different frames, presenting a displacement map of the query, commonly used in motion feature learning~\cite{motionsqueeze, corrnet} or optical flow estimation~\cite{bian2022learning,ng2018actionflownet,sun2018pwc,raft}.
By connecting the regions across the $L$ frames, the tensor turns out to reveal the \textit{motion flows} of the query over time.

\subsection{Generalization to Higher-Order STSS}
\label{sec:higher_order_stss}

\noindent \textbf{High-Order STSS Transformation.}
Unlike the conventional STSS, higher-order STSS explores the \textit{similarity of similarity} patterns themselves, providing a deeper understanding of motion dynamics.
However, recursively applying $f$ is impractical since the tensor dimension increases exponentially.
To address this, we introduce an STSS encoding function $g: \mathbb{R}^{T\times H \times W \times L \times U \times V} \rightarrow  \mathbb{R}^{T \times H \times W \times C}$, which abstracts features from the STSS tensor while mapping the high-dimensional tensor to the original feature space.
Note that $g$ can be an arbitrary function including vectorization, pooling operations, parametrized learnable encoders, or their compositions.
By composing $f$ and $g$, we can define a recursive process for computing higher-order STSS tensors while keeping the feature dimensions consistent (Fig.~\ref{fig:main_a}).
Considering the original STSS is presented as the 1st-order STSS tensor $\mathbf{\mS}^{(1)}$, we define the $n$-th order STSS tensor $\mathbf{\mS}^{(n)}$ recursively as,
\begin{align}
    \mathbf{S}^{(n)} = \begin{cases}
        f(\mathbf{F}), & \text{if } n = 1 \\
        f \circ g\left( \mathbf{S}^{(n-1)} \right), & \text{if } n \geq 2.
    \end{cases}
    \label{eq:high_order_stss}
\end{align}

\noindent \textbf{Advantages and Key Features of High-Order STSS.}
Higher-order STSSs play distinct roles compared to the 1st-order STSS in understanding spatio-temporal dynamics.
While 
\begin{wrapfigure}{r}{0.42\linewidth}
    \vspace{-3mm}
    \centering
    \includegraphics[width=\linewidth]{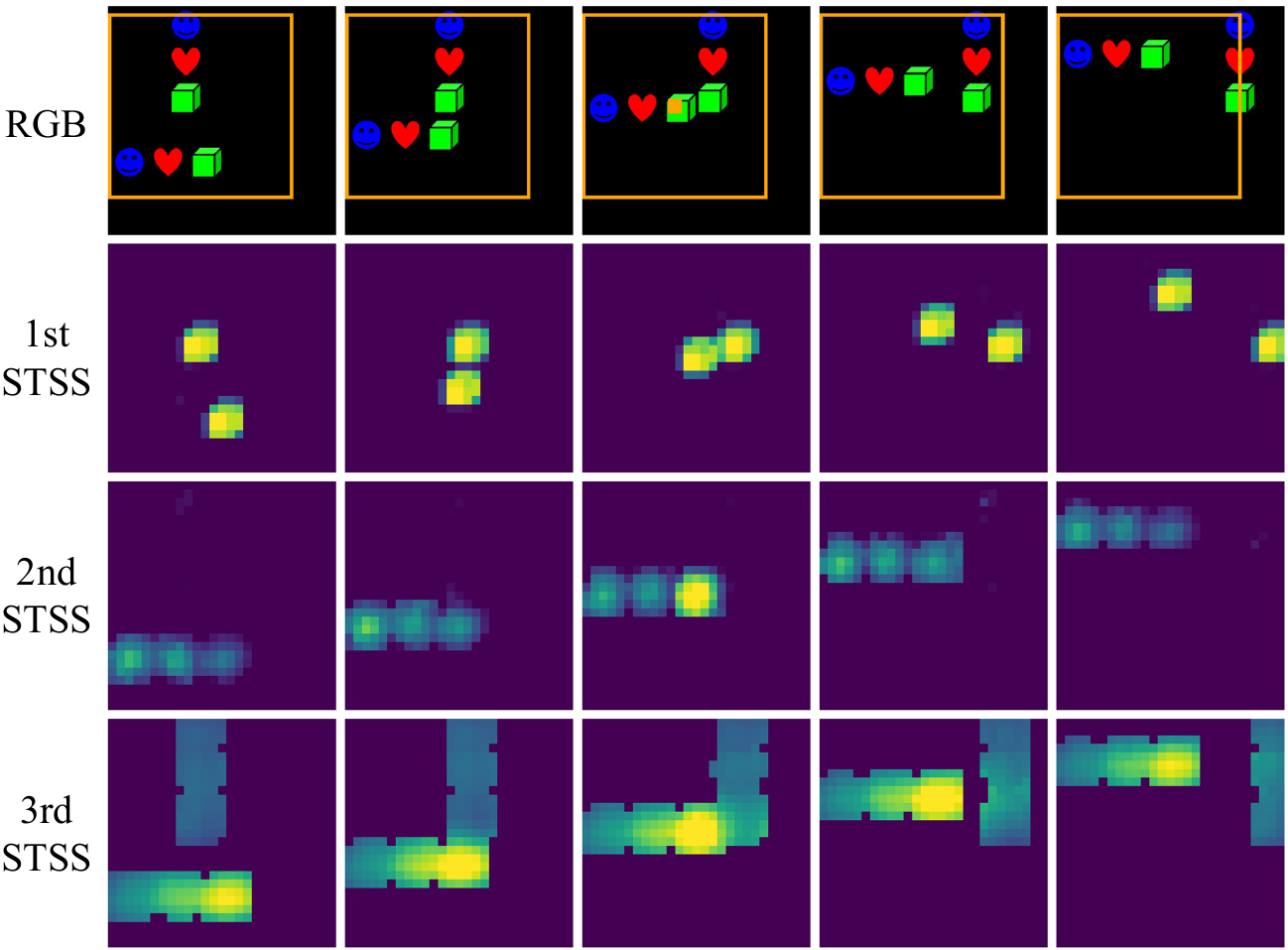}
    \vspace{-3mm}
    \caption{\textbf{STSS map visualizations on a toy video clip}. RGB frames and \mbox{1st-}, \mbox{2nd-}, and 3rd-order STSS maps of the \textcolor{brown}{brown} query by setting STSS encoding function $g$ as vectorization over $(L,U,V)$. The maps progressively capture \textit{motion flow}, \textit{motion segments}, and \textit{overall motion layouts}.}
    \label{fig:main_toy_visualization}
    \vspace{-7mm}
\end{wrapfigure}the 1st-order STSS reveals basic motion flows (\eg~existence of motion, directions) by computing similarities based on appearance across frames,  the 2nd-order STSS computes motion-based similarities and reveals regions with coherent motion patterns, akin to \textit{motion segments}.
Since pixels within the same object share similar motion flows, these motion segments effectively highlight moving objects and their motion trajectories from complex scenes.
This distinct nature of the 2nd-order STSS provides complementary temporal cues in scenarios where the 1st-order STSS fails; the 1st-order STSS struggles to distinguish pure motion of the query from motions of other visually similar objects (2nd row, Fig.~\ref{fig:main_toy_visualization}), whereas the 2nd-order STSS separates the query's motion from other visually similar objects while highlighting other objects with similar motion patterns (3rd row, Fig.~\ref{fig:main_toy_visualization}).

The 3rd-order STSS further extends this hierarchy by computing similarities based on motion segments, demonstrating the \textit{overall layouts of these segments.} 
Unlike the 2nd-order STSS that identifies individual motion segments, the 3rd-order STSS captures how these segments interact with each other, revealing the overall motion patterns (4th row, Fig.~\ref{fig:main_toy_visualization}). This can benefit understanding complex actions involving multiple simultaneous motions (\eg~group actions).
This progression from 1st- to higher-order STSS unveils diverse aspects of motion dynamics for comprehensive video understanding.

While our framework theoretically supports higher-order STSSs ($n \geq 4$), our empirical analysis suggests that STSS beyond 3rd-order does not provide significant benefits for temporal modeling.

\subsection{Learning Multi-Order STSS Representations}
We here introduce the \ours(Multi-Order Self-Similarity) module, a lightweight neural module that transforms multi-order STSS tensors into neural motion features.
We first explain our STSS encoder $g$ that effectively exploits structural patterns of the STSS tensor at each order and then describe \ours that combines multi-order STSS features into a deeper motion representation.

\noindent \textbf{STSS Encoder.}
To obtain the $n$-th order STSS representation $\mathbf{M}^{(n)}$, we express the computation as,
\begin{align}
    \vspace{-1mm}
    \mathbf{M}^{(n)} = g^{(n)} \left( \mathbf{S}^{(n)} \right),
    \vspace{-1mm}
\end{align}
where we employ an independent STSS encoder $g^{(n)}$ for each order.
We design $g$ in a late-fusion manner, \ie, encoding spatial structures first then fusing temporal information~\cite{selfy}, as illustrated in Fig.~\ref{fig:main_b}.
This order follows from the tensor itself: each $(U,V)$ slice is a displacement map between two frames and the $L$ axis indexes these maps over time, so motion flow emerges only once the per-offset structures are explicit.
We first transform the structural patterns of each spatial similarity map across $L$ frames into a $D$-dimensional vector by flattening the $(U,V)$ dimensions and applying a fully connected layer, resulting in a tensor of size $\mathbb{R}^{T \times H \times W \times L \times D}$.
While the previous methods~\cite{selfy,atm} used a series of 2D convolutions for $(U,V)$ extraction, we found that a simple linear layer achieves competitive performance while reducing memory overhead.
Next, we refine the spatial similarity features by applying a series of 2D convolutions over $(H,W)$ dimensions. Each convolution block consists of $\mathtt{Conv2d}\text{--}\mathtt{BatchNorm}\text{--}\mathtt{GeLU}$, maintaining $D$ channels.
Finally, we concatenate $L$ refined similarity features along the channel dimension and apply a fully connected layer to integrate features across temporal offsets, resulting in a tensor of size $\mathbb{R}^{T \times H \times W \times C}$.

\noindent \textbf{\ours Module.} The final output feature maps are obtained by combining the multi-order STSS feature maps with the original visual feature maps as,
\begin{align}
    \vspace{-2mm}
    \texttt{MOSS}(\mathbf{F}) = \texttt{FC} \left(\mathbf{F}\right) + \sum_{n=1}^N \texttt{FC} \left(\mathbf{M}^{(n)}\right).
    \vspace{-2mm}
\end{align}
This combination allows our model to leverage both the original visual features and the diverse motion patterns captured by multi-order STSS features.


\section{Experiments}
\label{sec:experiments}
    The proposed MOSS module is generic, so it can be integrated into a wide range of neural architectures and tasks. We first validate the effectiveness of MOSS on video action recognition, and then further demonstrate its versatility in improving temporal understanding for video MLLMs and VLAs.

\subsection{Video Action Recognition}
\label{sec:video_action_recognition}
\vspace{-2mm}
\noindent \textbf{Datasets.}
{\em Something-Something V1 \& V2}~\cite{goyal2017something,mahdisoltani2018effectiveness} contain 108K and 220K video clips, respectively, focusing on fine-grained actions.
{\em Diving48}~\cite{li2018resound} is a human diving action dataset consisting of 18K videos with 48 classes.
{\em FineGym}~\cite{shao2020finegym} is a fine-grained action benchmark containing 33K gymnastics videos.
These datasets emphasize temporal relationships through motion-centric action categories, where success depends on accurate modeling of complex spatio-temporal dynamics.

\begin{wrapfigure}{rt}{0.48\linewidth}
    \centering
    \includegraphics[width=\linewidth]{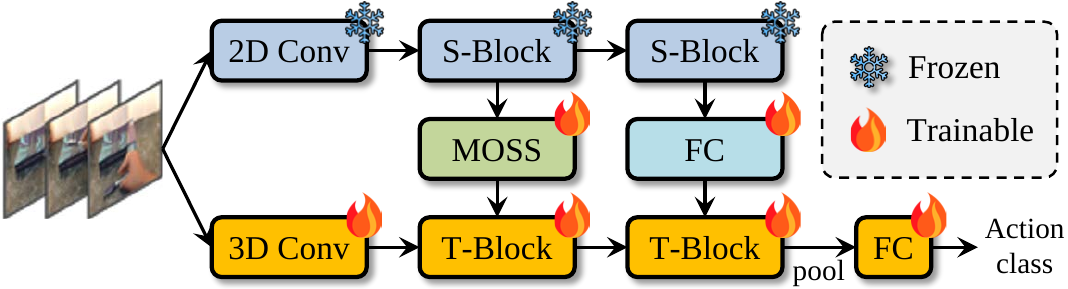}
    \vspace{-3mm}
    \caption{\textbf{Overall video architecture}.}
    \label{fig:video_model}
    \vspace{-2mm}
\end{wrapfigure}

\noindent \textbf{Implementation Details.}
We integrate the \ours module into Side4Video~\cite{side4video} as illustrated in Fig.~\ref{fig:video_model}.  We insert a single MOSS module between the spatial and temporal encoders, so it extracts multi-order STSS features from the visual features and feeds them to the temporal encoder as motion cues.
Please refer to Appendix~\ref{sec:supp_impl_ar} for detailed model and training configurations.


\begin{table*}[t]
  \caption{\textbf{Results on motion-centric video benchmarks.} $^\dag$ trained with text supervision. * reproduced with our setup. ``Input'' indicates \# frames$\times$\# crops$\times$\# clips. VM, CLIP400M, and Merged-2B indicate the pre-training corpora of V-JEPA~\cite{vjepa,vjepa2}, CLIP~\cite{clip}, and EVA-CLIP~\cite{evaclip}, respectively.}
  \vspace{-1mm}
  \setbox0=\hbox{\begin{minipage}[t]{0.67\linewidth}
    \begin{subtable}[t]{\columnwidth}
    \subcaption{Something-Something V1 \& V2}
    \vspace{-1mm}
\scalebox{0.74}{
\setlength{\tabcolsep}{4pt}
\begin{tabular}{lcccccc}
  \Xhline{0.8pt}
  \multirow{2}{*}{method} & \multirow{2}{*}{backbone} & \multirow{2}{*}{pre-train} & \multirow{2}{*}{input} & \multirow{2}{*}{TFLOPs} & SSV1 & SSV2 \\
   &  &  &  &  & top1 & top1 \\
  \Xhline{0.2pt}
  \textit{Full fine-tuning} & & & &\\
  \gray{ViViT~\cite{vivit}} & \gray{L/16$\times$2 FE} & \gray{IN21K,K400} & \gray{32$\times$12} & \gray{1.0$\times$12} & - & \gray{65.9} \\
  \gray{UniFormerV2~\cite{uniformerv2}} & \gray{ViT-L/14}& \gray{CLIP400M} & \gray{32$\times$3} & \gray{1.73$\times$3} & \gray{62.7} & \gray{73.0} \\
  \gray{ATM~\cite{atm}} & \gray{ViT-L/14} & \gray{CLIP400M} & \gray{16$\times$6} & \gray{0.84$\times$6} & \gray{64.0} & \gray{73.5} \\
  \Xhline{0.2pt}
  \textit{Frozen backbone} & & & &\\
  V-JEPA~\cite{vjepa} & ViT-H/16 & VM2M & 16$\times$6 & - & - & 74.3 \\
  V-JEPA 2~\cite{vjepa2} & ViT-H/16 & VM22M & 16$\times$6 & - & - & 74.0 \\
  M$^2$CLIP$^\dag$~\cite{m2clip} & ViT-B/16 & CLIP400M & 16$\times$12 & - & - & 67.3 \\
  OmniCLIP$^\dag$~\cite{omniclip} & ViT-B/16 & CLIP400M & 32$\times$3 & 0.8$\times$3 & - & 69.1 \\
  EVL~\cite{evl} & ViT-L/14 & CLIP400M & 32$\times$3 & 3.21$\times$3 & - & 66.7 \\
  ST-Adapter~\cite{stadapter} & ViT-L/14 & CLIP400M & 32$\times$3 & 2.75$\times$3 & - & 72.3 \\
  DualPath~\cite{dualpath} & ViT-L/14 & CLIP400M & 48$\times$3 & 0.72$\times$3 & - & 72.2 \\
  AIM~\cite{aim} & ViT-L/14 &  CLIP400M & 32$\times$3 & 3.84$\times$3 & - & 70.6 \\
  DiST$^\dag$~\cite{dist} & ViT-L/14 & CLIP400M & 32$\times$3 & 2.83$\times$3 & - & 73.1 \\ 
  MoTED$^\dag$~\cite{moted} & ViT-L/14 & CLIP400M & 32$\times$3 & 2.89$\times$3 & - & 73.8 \\ 
  Qian \etal~\cite{qian2025rethinking} & ViT-L/14 & CLIP400M & 32$\times$3 & 1.69$\times$3 & - & 73.6 \\
  Side4Video~\cite{side4video} & ViT-B/16 & CLIP400M & 16$\times$6 & 0.36$\times$6 & 60.7 & 71.5 \\
  Side4Video~\cite{side4video} & ViT-L/14 & CLIP400M & 16$\times$6 & 1.74$\times$6 & 62.4 & 73.2 \\
  Side4Video~\cite{side4video} & ViT-E/14 & Merged-2B & 16$\times$6 & 15.96$\times$6 & \textbf{67.3} & 75.2 \\
  \rowcolor{gray!15} MOSS-B (ours) & ViT-B/16 & CLIP400M & 16$\times$6 & 0.36$\times$6 & 61.8 & 72.4 \\
  \rowcolor{gray!15} MOSS-L (ours) & ViT-L/14 & CLIP400M & 16$\times$6 & 1.67$\times$6 & 64.8 & 74.4 \\
  \rowcolor{gray!15} MOSS-L (ours) & ViT-L/14 & Merged-2B & 16$\times$6 & 1.67$\times$6 & \textbf{67.3} & \textbf{75.3} \\
  \Xhline{0.8pt}
\end{tabular}}
\label{tab:motion_benchmarks_ss_v1_v2}
\end{subtable}
\hfill
\end{minipage}}%
\edef\tabonehgt{\the\dimexpr\ht0+\dp0\relax}%
\leavevmode\box0\ %
\begin{minipage}[t][\tabonehgt][s]{0.32\linewidth}
\begin{subtable}[t]{\columnwidth}
\subcaption{Diving48}
\vspace{-1mm}
\scalebox{0.74}{
\setlength{\tabcolsep}{2pt}
    \begin{tabular}{L{\dimexpr3.8cm-2pt\relax}|C{1.5cm}}
    \Xhline{0.8pt}
    method & top1 \\
    \Xhline{0.2pt}
    TimeSformer-L~\cite{timesformer}    & 81.0 \\
    V-JEPA~\cite{vjepa} & 87.9 \\
    ORViT~\cite{herzig2022object} & 88.0 \\
    StructViT-B-4-1~\cite{structvit} & 88.3 \\
    Side4Video-B*~\cite{side4video} & 88.6 \\
    V-JEPA 2~\cite{vjepa2} & 89.8 \\
    AIM ViT-L~\cite{aim} & 90.6 \\
    Video-FocalNet-B~\cite{videofocalnet} & 90.8 \\
    \rowcolor{gray!15} MOSS-B (ours) & 91.2 \\
    \rowcolor{gray!15} MOSS-L (ours) & \textbf{92.7} \\
    \Xhline{0.8pt}
    \end{tabular}
}
\label{tab:motion_benchmarks_diving48}
\end{subtable}

\vfill

\begin{subtable}[t]{\columnwidth}
\subcaption{FineGym}
\vspace{-1mm}
\scalebox{0.74}{
\setlength{\tabcolsep}{1pt}
    \begin{tabular}{L{2.9cm}|C{1.2cm}C{1.2cm}}
    \Xhline{0.8pt}
    method & gym99 & gym288 \\
    \Xhline{0.2pt}
    TSM$_{\textrm{two-stream}}$~\cite{lin2019tsm}  & 81.2 & 46.5 \\
    RSANet~\cite{rsa}                       & 86.4 & 50.9 \\
    StructViT-B-4-1~\cite{structvit}               & 89.5 & 54.2 \\
    TQN~\cite{zhang2021temporal} & 90.6 & 61.9 \\
    VT-CE~\cite{vtce} & 91.4 & 62.6 \\
    Side4Video-B*~\cite{side4video} & 92.3 & 69.1 \\
    \rowcolor{gray!15} MOSS-B (ours) & 93.9 & 70.2 \\
    \rowcolor{gray!15} MOSS-L (ours) & \textbf{94.7} & \textbf{71.1} \\
    \Xhline{0.8pt}
    \end{tabular}
}
\label{tab:motion_benchmarks_finegym}
\end{subtable}
\vspace{0pt}%
\end{minipage}
\vspace{-2mm}
\label{tab:motion_benchmarks}
\end{table*}

\noindent \textbf{Results.}
In Tab.~\ref{tab:motion_benchmarks_ss_v1_v2}, we present the results on Something-Something V1 and V2.
Using 16 frames, MOSS-B attains top-1 accuracies of 61.8\% and 72.4\% on V1 and V2, respectively, achieving performance comparable to adapter-based PEFT methods~\cite{stadapter,dualpath,aim} with the larger ViT-L/14 backbone.
Scaling up to MOSS-L, we achieve strong performances of 64.8\% on V1 and 74.4\% on V2, significantly surpassing all the CLIP-based methods at the same ViT-L scale and even competing with video foundation models~\cite{vjepa,vjepa2} using larger ViT-H backbones.
We then replace the spatial encoder with EVA-CLIP~\cite{evaclip} and obtain 67.3\% on V1 and 75.3\% on V2, competitive with Side4Video with larger ViT-E backbone while requiring 10$\times$ fewer FLOPs.
These results demonstrate the effectiveness of high-order STSSs in understanding temporal dynamics.

We summarize the results on Diving48 and FineGym in Tabs.~\ref{tab:motion_benchmarks_diving48} and~\ref{tab:motion_benchmarks_finegym}, respectively, which contain more complex motion patterns than Something-Something datasets. For both benchmarks, MOSS-B outperforms all other methods~\cite{vjepa2,vjepa,timesformer,herzig2022object,rsa,structvit,vtce,lin2019tsm,aim,side4video,zhang2021temporal,videofocalnet}; MOSS-L obtains 92.7\% on Diving48, 94.7\% on gym99, and 71.1\% on gym288, achieving state-of-the-art by large margins.

We also validate the consistent effectiveness of our method on Kinetics-400 and other video tasks such as temporal action detection and generic event boundary detection (Appendices~\ref{sec:experiments_k400}--\ref{sec:experiments_gebd}).

\subsection{Ablation Studies}
\label{sec:ablation}
We here provide in-depth analyses of high-order STSS on Something-Something V1 and Diving48 using MOSS-S with 8 and 32 frames, respectively.

\noindent \textbf{Effect of High-Order STSS.}
In Tab.~\ref{tab:ablation_stss}, we show the individual effects of incorporating STSS at different orders; in every row the lower orders are still computed, and only their features are excluded from the fusion.
Compared to the baseline without any STSS, the 1st-order STSS improves the top-1 accuracy by 2.1\%p on Something-Something V1, highlighting the effectiveness of the motion information it captures.
The 2nd- \& 3rd-order STSS also demonstrate distinct improvements in accuracy by 1.8\%p and 1.4\%p, respectively.
Notably, combining the 1st-order and 2nd-order STSS leads to additional performance gains over using each order independently, suggesting that they provide complementary temporal cues, while incorporating the 3rd-order does not lead to additional improvements.
Diving48 exhibits a similar trend, confirming the consistent benefits of multi-order STSS in temporal modeling.


\begin{table*}[t]
    \centering
    \setlength\tabcolsep{4pt}
    \captionsetup{width=\linewidth}
    \captionsetup[sub]{size=footnotesize}
    \caption{\textbf{Ablation studies on Something-Something V1 and Diving48 datasets.} All the experiments are conducted with MOSS-S using 8- and 32-frame inputs on Something-Something V1 and Diving48, respectively. ``FLOPs'', ``TP'', and ``Mem'' respectively indicate FLOPs (G), trainable parameters (M), and memory footprint (GB) using 8 frames. Memory footprint is measured with a batch size of 32 on a single GPU. Rows in \colorbox{gray!15}{gray} indicate our default configurations.}
    \begin{subtable}[t]{0.493\linewidth}
        \centering
        \subcaption{Effect of High-Order STSS}
        \vspace{-1mm}
        \setlength\tabcolsep{2.5pt}
        \scalebox{0.77}{
        \begin{tabular}{L{2.8cm}|C{1.1cm}C{0.7cm}C{0.8cm}|C{0.9cm}C{0.7cm}}
             \Xhline{0.8pt}
              STSS orders   & FLOPs & TP & Mem & SSV1 & D48  \\
              \Xhline{0.3pt}
               - &  148.4 & 4.5 & 8.0 & 56.9 & 85.0\\
              \Xhline{0.3pt}
               $+$ 1st  & 150.0 & 5.1 & 9.0 & 59.0 & 86.3\\
               $+$ 2nd & 151.5 & 5.6 & 9.9 & 58.7 & 86.1\\
               $+$ 3rd  & 152.9 & 6.1 & 10.7 & 58.3 & 85.7\\
               \cellcolor{gray!15}$+$ MOSS (1st$+$2nd) & \cellcolor{gray!15}151.5 & \cellcolor{gray!15}5.6 & \cellcolor{gray!15}9.9 & \cellcolor{gray!15}\textbf{60.0} & \cellcolor{gray!15}\textbf{87.7}\\
              \Xhline{0.8pt}
        \end{tabular}
        }%
    \label{tab:ablation_stss}
    \end{subtable}
    \hfill
    \begin{subtable}[t]{0.493\linewidth}
        \centering
        \subcaption{High-Order STSS Transformation}
        \vspace{-1mm}
        \setlength\tabcolsep{2.5pt}
        \scalebox{0.77}{
        \begin{tabular}{L{\dimexpr2.0cm+4\tabcolsep+\arrayrulewidth\relax}C{3.4cm}|C{0.9cm}C{0.7cm}}
        \Xhline{0.8pt}
        fusion & $\mathbf{S}^{(2)}$ & SSV1 & D48 \\
        \Xhline{0.3pt}
        1st STSS only & - & 59.0 & 86.3 \\
        \Xhline{0.3pt}
        MLP & $f(\texttt{MLP}([\mathbf{F}, \mathbf{M}^{(1)}]))$& 59.1 & 86.6 \\
        conv & $f(\texttt{Conv2d}([\mathbf{F}, \mathbf{M}^{(1)}]))$& 59.2 & 86.6 \\
        addition & $f(\mathbf{F}+\mathbf{M}^{(1)})$& 59.4 & 86.8 \\
        \cellcolor{gray!15}no-fusion & \cellcolor{gray!15}$f(\mathbf{M}^{(1)})$ & \cellcolor{gray!15}\textbf{60.0} & \cellcolor{gray!15}\textbf{87.7}\\
        \Xhline{0.8pt}
        \end{tabular}
        }%
        \label{tab:ablation_connection}
    \end{subtable}

    \vspace{2mm}

    \begin{subtable}[t]{0.493\linewidth}
        \centering
        \subcaption{Different Image Encoders}
        \vspace{-1mm}
        \setlength\tabcolsep{2.5pt}    
        \scalebox{0.77}{
        \begin{tabular}[t]{L{2.8cm}|C{1.1cm}C{0.7cm}C{0.8cm}|C{0.9cm}C{0.7cm}}
        \Xhline{0.8pt}
        ViT-B & FLOPs & TP & Mem & SSV1 & D48 \\
        \Xhline{0.3pt}
        MAE~\cite{mae} & 148.4 & 4.5 & 8.0 & 53.1 & 83.6 \\
        $+$1st-order & 150.0 & 5.1 & 9.0 & 54.9 & 86.3\\
        $+$MOSS (ours) & 151.5 & 5.6 & 9.9 & \textbf{56.0} & \textbf{87.2}\\
        \Xhline{0.3pt}
        DINO~\cite{dino} & 148.4 & 4.5 & 8.0 & 53.5 & 84.1\\
        $+$1st-order & 150.0 & 5.1 & 9.0 & 55.3 & 85.0\\
        $+$MOSS (ours) & 151.5 & 5.6 & 9.9 & \textbf{56.6} & \textbf{86.3}\\
        \Xhline{0.8pt}
        \end{tabular}
        }%
        \label{tab:ablation_pretrain}
    \end{subtable}
    \hfill
    \begin{subtable}[t]{0.493\linewidth}
        \centering
        \subcaption{Comparison to Other Temporal Modules}
        \setlength\tabcolsep{2.5pt}
        \vspace{-1mm}
        \scalebox{0.77}{
        \begin{tabular}{L{2.8cm}|C{1.1cm}C{0.7cm}C{0.8cm}|C{0.9cm}C{0.7cm}}
        \Xhline{0.8pt}
        method & FLOPs & TP & Mem & SSV1 & D48 \\
        \Xhline{0.3pt}
        baseline & 148.4 & 4.5 & 8.0 & 56.9 & 85.0\\
        \Xhline{0.3pt}
        R(2+1)D & 151.4 & 6.4 & 9.6 & 57.5 & 86.1\\
        Fact Attn. & 151.8 & 6.6 & 11.5 & 57.4 & 85.9\\
        Local Attn. & 150.6 & 5.6 & 11.0 & 57.5 & 85.5\\
        SELFY~\cite{selfy} & 151.1 & 5.1 & 11.2 & 59.2 & 87.0\\
        ATM~\cite{atm} & 153.2 & 6.0 & 12.1 & 59.6 & 87.2\\
        \cellcolor{gray!15}MOSS (ours) & \cellcolor{gray!15}151.5 & \cellcolor{gray!15}5.6 & \cellcolor{gray!15}9.9 & \cellcolor{gray!15}\textbf{60.0} & \cellcolor{gray!15}\textbf{87.7}\\
        \Xhline{0.8pt}
        \end{tabular}
        }%
        \label{tab:ablation_temporal_module}
    \end{subtable}

    \vspace{2mm}

    \begin{subtable}[t]{0.493\linewidth}
        \centering
        \subcaption{Other STSS Learning Methods}
        \vspace{-1mm}
        \setlength\tabcolsep{2.5pt}
        \scalebox{0.77}{
        \begin{tabular}{L{2.8cm}|C{1.1cm}C{0.7cm}C{0.8cm}|C{0.9cm}C{0.7cm}}
        \Xhline{0.8pt}
        method & FLOPs & TP & Mem & SSV1 & D48 \\
        \Xhline{0.3pt}
        1st STSS only & 150.0 & 5.1 & 9.0 & 59.0 & 86.3\\
        \Xhline{0.3pt}
        + R(2+1)D & 151.1 & 5.8 & 9.7 & 59.1 & 86.7\\
        + Fact Attn. & 151.2 & 5.8 & 11.0 & 59.2 & 86.4\\
        + Diff. & 151.5 & 5.6 & 9.7 &  59.2 & 86.5\\
        \Xhline{0.3pt}
        \cellcolor{gray!15}+ 2nd STSS & \cellcolor{gray!15}151.5 & \cellcolor{gray!15}5.6 & \cellcolor{gray!15}9.9 & \cellcolor{gray!15}\textbf{60.0} & \cellcolor{gray!15}\textbf{87.7}\\
        \Xhline{0.8pt}
        \end{tabular}
        }%
        \label{tab:ablation_matching_cost}
    \end{subtable}
    \hfill
    \begin{subtable}[t]{0.493\linewidth}
        \centering
        \subcaption{Fine-tuning Methods}
        \setlength\tabcolsep{1.7pt}
        \vspace{-1mm}
        \def\ftbody{%
             \Xhline{0.8pt}
             FT & method & FLOPs & TP & Mem & SSV1 & D48 \\
              \Xhline{0.3pt}
              full & ViT-B~\cite{clip} & 140.7 & 86.4 & 28.3 & 51.9 & 84.2\\
              FT & $+$ MOSS (ours) & 144.1 & 87.6 & 30.3 & \textbf{59.6} & \textbf{87.8}\\
              \Xhline{0.3pt}
              \multirow{2}{*}{PEFT} & AIM ViT-B~\cite{aim} & 207.9 & 14.3 & 38.6 & 54.8 & 87.3\\
              & $+$ MOSS (ours) & 211.3 & 15.6 & 40.9 & \textbf{57.4} & \textbf{89.4}\\
              \Xhline{0.3pt}
              \multirow{4}{*}{LST} & Side4Video~\cite{side4video} & 148.4 & 4.5 & 8.0 & 56.9 & 85.0 \\
              & \cellcolor{gray!15}$+$ MOSS (ours) & \cellcolor{gray!15}151.5 & \cellcolor{gray!15}5.6 & \cellcolor{gray!15}9.9 & \cellcolor{gray!15}\textbf{60.0} & \cellcolor{gray!15}\textbf{87.7}\\
               & DiST~\cite{dist} & 163.1 & 19.0 & 11.1 & 55.6 & 86.3 \\
                & $+$ MOSS (ours) & 165.5 & 20.3 & 12.8 & \textbf{58.5} & \textbf{88.9} \\
              \Xhline{0.8pt}}
        \edef\ftwD{\the\dimexpr0.9cm+1.6pt\relax}\edef\ftwE{\the\dimexpr0.7cm+1.6pt\relax}
        \setbox0=\hbox{\begin{tabular}{c|l|ccc|C{\ftwD}C{\ftwE}}\ftbody\end{tabular}}
        \dimen0=\dimexpr(7cm+30pt+2\arrayrulewidth-\wd0)/3\relax
        \ifdim\dimen0<0pt \dimen0=0pt \fi
        \def\colw#1#2{\settowidth{\dimen2}{#2}\edef#1{\the\dimexpr\dimen2+\dimen0\relax}}
        \colw\ftwA{FLOPs}\colw\ftwB{86.4}\colw\ftwC{Mem}
        \scalebox{0.77}{
        \begin{tabular}{c|l|C{\ftwA}C{\ftwB}C{\ftwC}|C{\ftwD}C{\ftwE}}\ftbody\end{tabular}
        }%
        \label{tab:ablation_finetuning}
    \end{subtable}%
\label{tab:ablation}
\end{table*}

\noindent \textbf{High-Order STSS Transformation.}
We compare transformation methods for computing the 2nd-order STSS $\mathbf{S}^{(2)}$: `MLP', `conv', `addition', and `no-fusion' (Eq.~\ref{eq:high_order_stss}).
The first three methods combine the original visual feature $\mathbf{F}$ with the 1st-order STSS feature $\mathbf{M}^{(1)}$ before calculating $\mathbf{S}^{(2)}$, while the last computes $\mathbf{M}^{(2)}$ purely from $\mathbf{M}^{(1)}$ without visual input.
Intuitively, one might expect that augmenting the 1st-order STSS features with visual features would provide richer context for the 2nd-order STSS, leading to better performance.
However, Tab.~\ref{tab:ablation_connection} shows the opposite: the `no-fusion' significantly outperforms others.
This suggests that the 1st- and 2nd-order STSS features capture complementary dynamics. The gain therefore comes from keeping the recursion within correspondence space, not from stacking more blocks or spending more computation.

\noindent \textbf{Spatial Encoders.}
In Tab.~\ref{tab:ablation_pretrain}, we evaluate MOSS on two other spatial encoders pretrained on different objectives: DINO~\cite{dino} and MAE~\cite{mae}.
The results show that both 1st- and 2nd-order STSS consistently improve performance across all pre-training objectives, demonstrating generalizability across various backbones, with gains of 2.9--3.1\%p on Something-Something V1.

\noindent 
\textbf{Comparison to Existing Temporal Modules.}
We compare our method to other temporal modeling modules~\cite{vivit,selfy,tran2019video,tran2018closer,atm} with similar computational costs.
We replace the MOSS module with different modules, including spatio-temporal convolution~\cite{tran2018closer}, factorized spatio-temporal attention~\cite{vivit}, local attention with a spatio-temporal window $(L, U, V)$, and the other STSS learning blocks~\cite{selfy,atm}.
SELFY and ATM extract 1st-order STSS features using convolutions, with ATM additionally performing frame-wise subtraction for richer dynamics.
The results in Tab.~\ref{tab:ablation_temporal_module} show that transforming STSS directly into motion features~\cite{selfy,atm} is more effective at capturing temporal dynamics than convolution or attention, consistent with prior work~\cite{selfy}.
However, the effectiveness of SELFY~\cite{selfy} and ATM~\cite{atm} is overshadowed by excessive memory overhead when applying a series of convolutions to the large STSS tensor $\mathbf{S}$.
In contrast, we use a simple FC layer to directly reduce the volume of $\mathbf{S}$. 
This enables memory-efficient processing of multi-order STSSs, leading to superior performance with less memory consumption.

\noindent
\textbf{Comparison to Other STSS Learning Methods.}
In Tab.~\ref{tab:ablation_matching_cost}, we delve into the effectiveness of high-order STSS by comparing different STSS learning methods.
We keep the 1st STSS encoder fixed and replace the 2nd STSS encoder with R(2+1)D convolution~\cite{tran2018closer}, factorized self-attention~\cite{vivit}, and frame-wise difference calculation~\cite{atm}.
The results show that other STSS learning methods provide marginal gains while our 2nd-order STSS encoder significantly boosts accuracy.
This indicates that the 2nd-order STSS encoder captures complementary temporal dynamics distinct from the 1st-order.

\noindent \textbf{Fine-tuning Methods.}
Although we integrate our MOSS module into the ladder side-tuning (LST) framework~\cite{side4video} for efficient action recognition in previous experiments,
MOSS is also compatible with various fine-tuning scenarios including full fine-tuning and parameter-efficient fine-tuning (PEFT)~\cite{stadapter,aim}.
Here we conduct experiments in such scenarios. 
For full fine-tuning, we add temporal convolution blocks and a single MOSS module to the CLIP-pretrained ViT-B~\cite{clip} and train the entire network following ATM~\cite{atm}.
For PEFT, we integrate MOSS into AIM~\cite{aim} and train the module and adapters keeping the backbone frozen.
For LST, we additionally conduct experiments on DiST~\cite{dist} by inserting a single MOSS module between the spatial encoder and the integration branch.
The results in Tab.~\ref{tab:ablation_finetuning} show that MOSS substantially improves performance with marginal overhead in all settings, demonstrating its flexibility in different image-to-video transfer methods.

\begin{figure*}[t]
    \centering
\begin{subfigure}[t]{0.49\linewidth}
\centering
    \includegraphics[width=\textwidth]{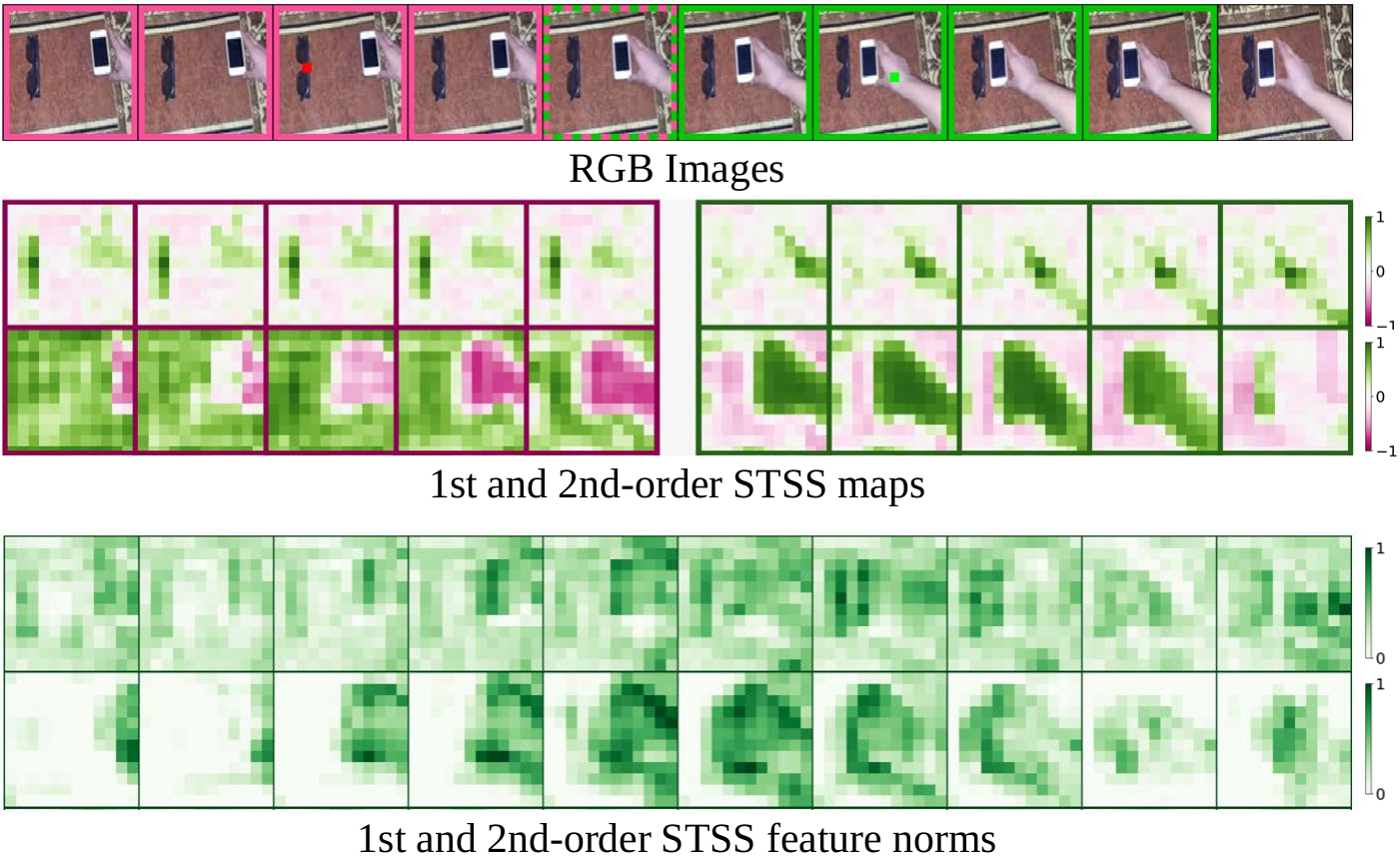}
    \vspace{-3mm}
    \caption{Moving something closer to something}
    \label{fig:stss_a}
\end{subfigure}
    \hspace{0.005\linewidth}%
\begin{subfigure}[t]{0.49\linewidth}
\centering
    \includegraphics[width=\textwidth]{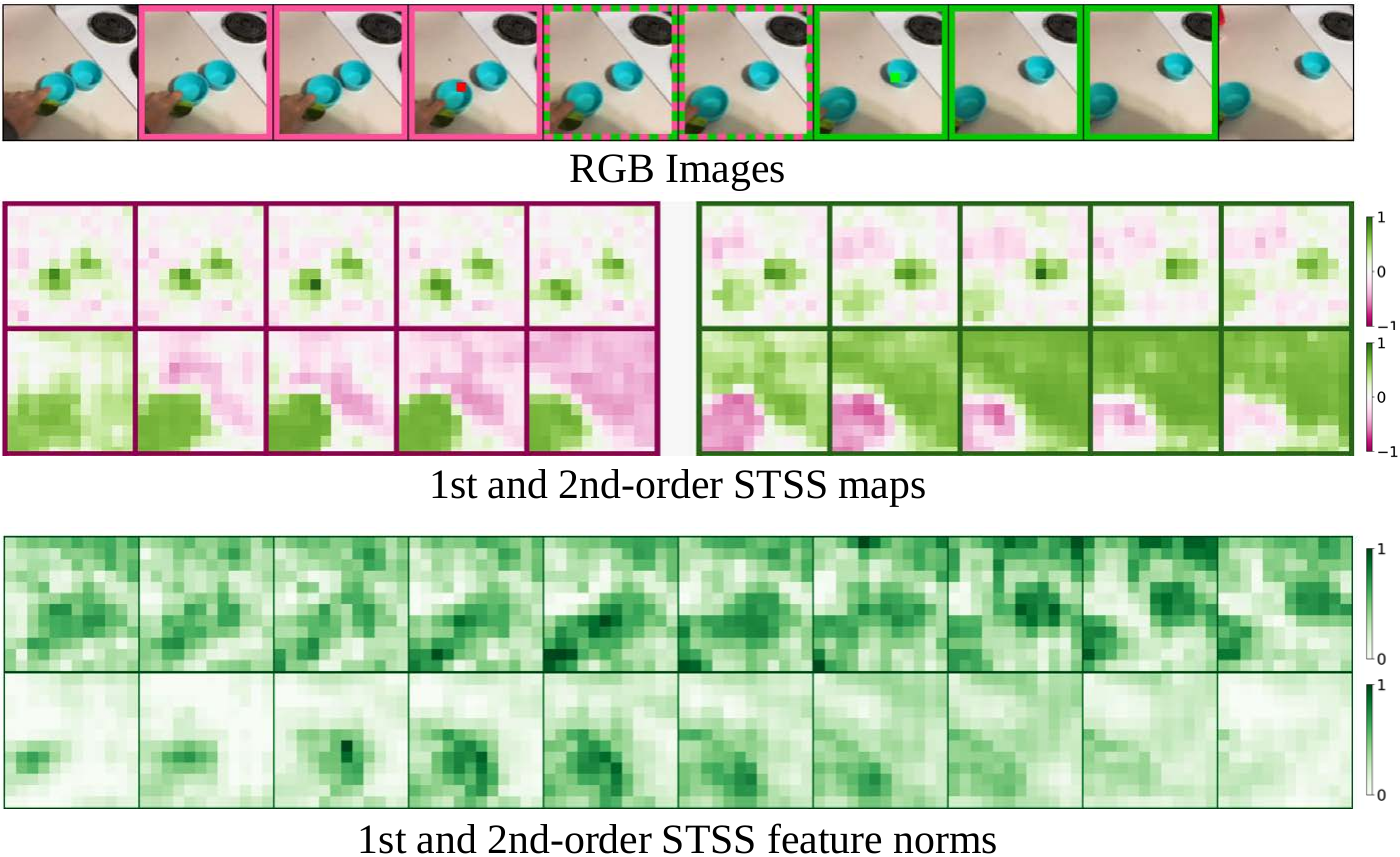}
    \vspace{-3mm}
    \caption{Moving something away from something}
    \label{fig:stss_b}
\end{subfigure}
    \vspace{1mm}
    \caption{\textbf{STSS visualization}. RGB frames are at the top, with two queries and their spatio-temporal matching regions marked in \red{red} and \textcolor{Green}{green}, respectively. The rows below show the STSS maps of both queries and the L2-norms of 1st- and 2nd-order feature maps. Best viewed in pdf.}
    \label{fig:stss}
\end{figure*}

\noindent \textbf{Visualization of Multi-Order STSS.}
In Fig.~\ref{fig:stss}, we present visualization results of 1st- and 2nd-order STSSs on Something-Something V1 to analyze their distinct contributions in temporal understanding.
We observe that the 1st-order STSS identifies basic motions of objects (2nd row, Fig.~\ref{fig:stss_a}), but struggles to distinguish between visually similar objects with different motion patterns (2nd row, Fig.~\ref{fig:stss_b}).
In contrast, the 2nd-order STSS overcomes this limitation by segmenting regions based on their motion patterns, effectively distinguishing visually similar objects and background regions (3rd row in Fig.~\ref{fig:stss_b}).
We also visualize the L2-norm of STSS feature maps across different orders and observe that the 2nd-order STSSs effectively suppress static regions while highlighting moving foreground objects (5th row).
These complementary contributions of multi-order STSS beyond basic motion enable comprehensive video understanding.

Please refer to Appendices~\ref{sec:supp_ablation} and \ref{sec:supp_perclass} for additional ablation studies, Appendix~\ref{sec:supp_self_attention} for an in-depth comparison with self-attention, Appendix~\ref{sec:supp_visualization} for qualitative results, and Appendix~\ref{sec:limitation} for discussions on limitations and future directions.

\subsection{MOSS in Video MLLMs}

\noindent \textbf{Datasets.} {\em FAVOR-Bench}~\cite{favorbench} evaluates fine-grained motion-level reasoning in video MLLMs, comprising 1,776 videos and 8,184 multiple-choice QA pairs across 6 motion-related tasks. We use 15K samples from the publicly released training set, {\em FAVOR-Train}, for fine-tuning.
{\em MotionBench}~\cite{motionbench} 
\begin{wrapfigure}{r}{0.44\linewidth}
    \centering
    \includegraphics[width=\linewidth]{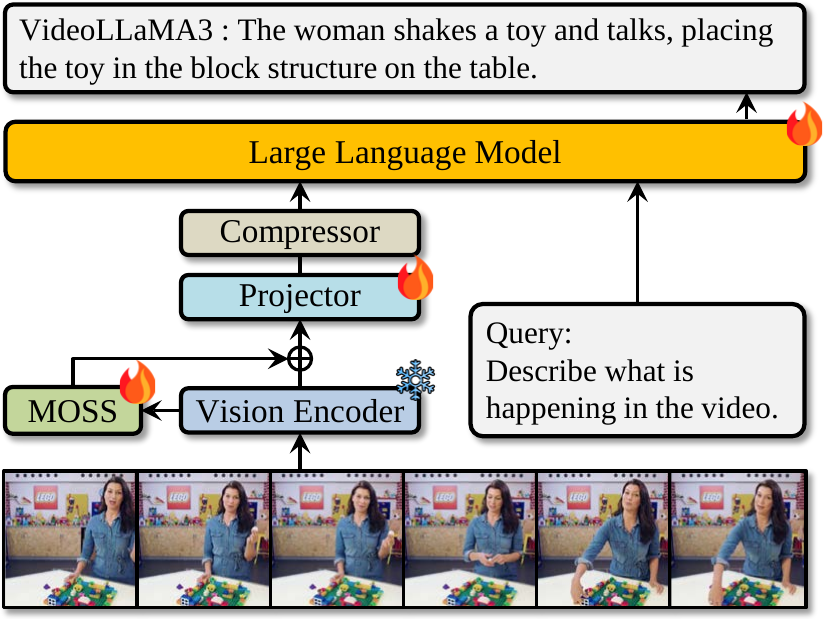}
    \vspace{-5.5mm}
    \caption{\textbf{VideoLLaMA3 with MOSS}.}
    \label{fig:videollm}
    \vspace{-7mm}
\end{wrapfigure}is another recent benchmark designed for measuring fine-grained motion understanding, consisting of 5,385 videos and 8,052 QA pairs across 6 tasks. We evaluate the FAVOR-Train fine-tuned model on its dev split (4,018 questions) in a zero-shot manner.

\noindent \textbf{Implementation Details.}
As our baseline, we adopt VideoLLaMA3-2B~\cite{videollama3}.
A single MOSS module is inserted after the 6th layer of the SigLIP vision encoder~\cite{siglip} to extract multi-order STSS features. The extracted STSS features are added before the projector to inject early motion cues into the LLM (Fig.~\ref{fig:videollm}), providing primitive temporal cues for motion-level video reasoning.
Please refer to Appendix~\ref{sec:supp_impl_videollm} for detailed training configurations.

\begin{table*}[t]
\centering
\caption{\textbf{Results on FAVOR-Bench and MotionBench-Dev.} 
FAVOR-Bench consists of six tasks: Action Sequence (AS), Camera Motion (CM), Holistic Action Classification (HAC), Multiple Action Details (MAD), Non-subject Motion (NSM), and Single Action Detail (SAD). MotionBench-Dev includes six tasks as well: Motion Recognition (MR), Location-related Motion (LM), Camera Motion (CM), Motion-related Objects (MO), Action Order (AO), and Repetition Count (RC).}
\scalebox{0.78}{
\setlength{\tabcolsep}{2pt}
\begin{tabular}{l|ccccccc|ccccccc}
\Xhline{0.8pt}
\multirow{2}{*}{method} 
    & \multicolumn{7}{c|}{FAVOR-Bench~\cite{favorbench}} 
    & \multicolumn{7}{c}{MotionBench-Dev~\cite{motionbench}} \\
    & all & AS & CM & HAC & MAD & NSM & SAD 
    & all & MR & LM & CM & MO & AO & RC \\
    \Xhline{0.2pt}
VideoLLaMA3-2B~\cite{videollama3} 
    & 42.2 & 43.8 & 27.3 & 44.4 & 48.1 & 45.3 & 42.8 
    & 50.2 & 54.9 & 54.2 & 36.1 & 68.3 & 37.0 & 27.3 \\
$+$ FAVOR-Train~\cite{favorbench} 
    & 45.5 & 44.8 & 27.8 & 54.5 & 51.3 & \textbf{54.7} & 45.3 
    & 51.4 & 55.8 & 54.4 & 36.4 & \textbf{68.8} & 38.3 & 33.0 \\
\rowcolor{gray!15} $+$ FAVOR-Train $+$ MOSS 
    & \textbf{46.6} & \textbf{46.8} & \textbf{28.8} & \textbf{55.0} & \textbf{52.2} & 53.3 & \textbf{45.7}
    & \textbf{54.2} & \textbf{59.7} & \textbf{55.7} & \textbf{48.3} & 68.1 & \textbf{38.5} & \textbf{34.0} \\
    \Xhline{0.8pt}
\end{tabular}
}
\label{tab:videollm}
\end{table*}

\noindent \textbf{Results.}
We summarize the results in Tab.~\ref{tab:videollm}. Compared to VideoLLaMA3-2B fine-tuned on the same data, incorporating MOSS improves overall accuracy by 1.1\%p on FAVOR-Bench.
We further evaluate on MotionBench without fine-tuning and observe that MOSS yields a notable 2.8\%p gain in overall accuracy, demonstrating generalizable improvements in motion understanding.
Notably, these improvements are achieved with only 13.7M additional parameters (0.6\% of the total) and $\sim$8 GPU-hours of training, making MOSS substantially more efficient than existing approaches~\cite{btadapter, slowfocus, staveq2} that require large-scale retraining on massive video datasets.
See Appendix~\ref{sec:supp_long_video} for further analyses.

\subsection{MOSS in VLAs}
\noindent \textbf{Simulation Benchmark.} \textit{RoboCasa-Kitchen}~\cite{robocasa} comprises 24 tasks in simulated kitchen environments, with over 2,500 scenes and 150 objects, enabling evaluation of policy generalization. We use 300 demos per task for training and evaluate with 50 trials per task.

\noindent \textbf{Real-World Robotic Tasks.} Since most existing robotic benchmarks~\cite{libero,simpler,robocasa} evaluate policies in static environments, we design three real-world robotic tasks, MoveSense, PongPredict, and Conveyor Pick-and-Place, in dynamic environments to evaluate the temporal understanding of VLAs (Fig.~\ref{fig:robot_task}).
\begin{itemize}[leftmargin=*]
\item \textbf{MoveSense}.
The robot observes a virtual object moving on the screen and must identify its motion pattern by pointing to the corresponding bowl. The object moves in one of two motion patterns: (1) left vs.\ right (translational motion) and (2) clockwise vs.\ counterclockwise (rotational motion). This task measures how accurately a VLA can {\em perceive object motion}.

\item \textbf{PongPredict}.
The robot observes a ping-pong ball traveling across the screen and must predict which region the ball will hit by pointing to the corresponding cup. The right wall is divided into three colored regions, red, green, and blue, and the robot must anticipate the target region before the ball reaches it. This task measures how accurately a VLA can {\em predict future object motion}.

\item \textbf{Conveyor Pick-and-Place (PnP)}.
The humanoid robot observes an object moving on a conveyor and must pick and place it onto a shelf in front. The conveyor operates at varying speeds, and evaluation includes unseen speeds beyond those used during training. This task measures how effectively a VLA can \textit{interact with moving objects} across diverse speeds.

\begin{figure}[t]
    \centering
    \begin{subfigure}[b]{0.63\linewidth}
        \centering
        \includegraphics[width=\columnwidth]{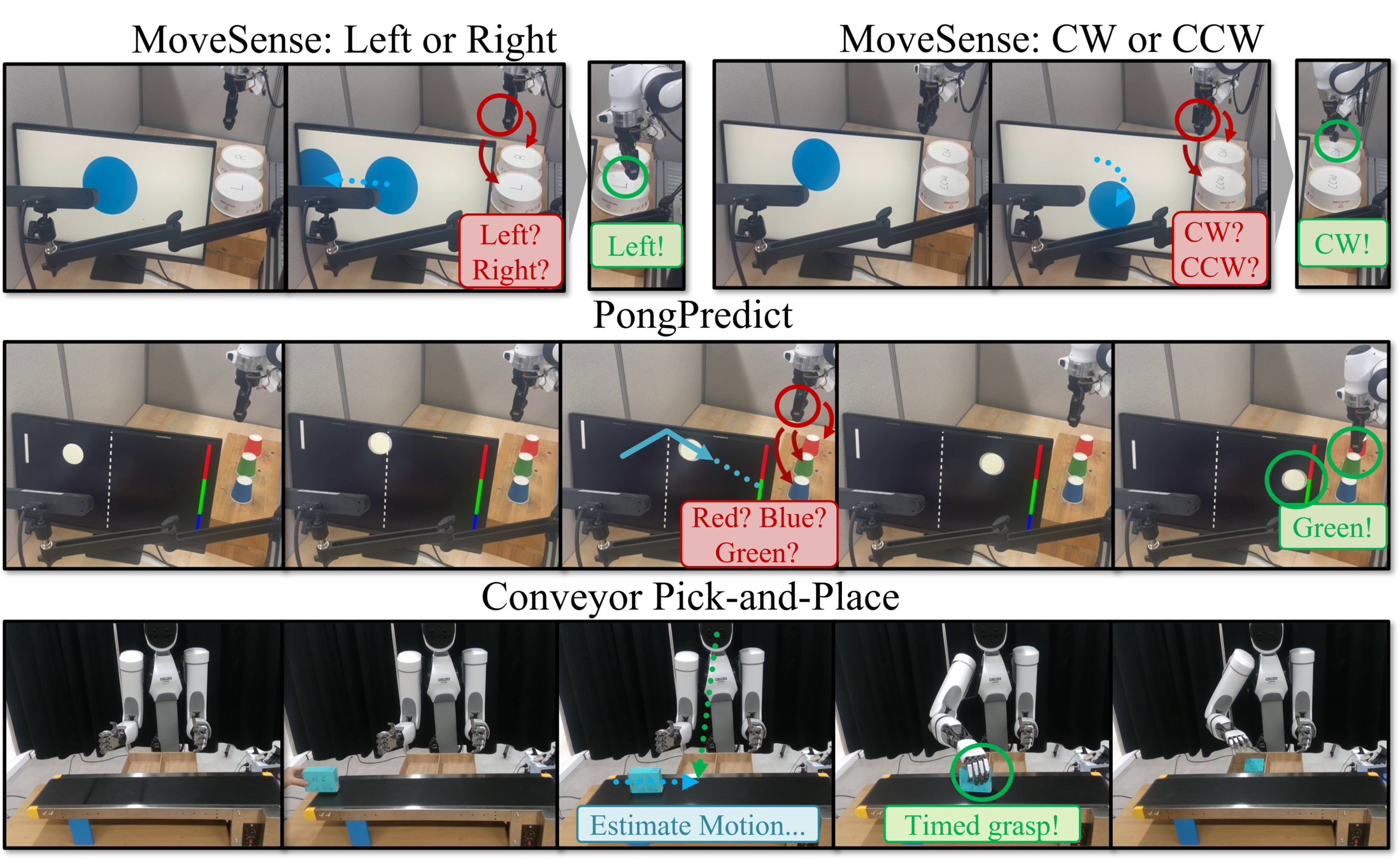}
        \vspace{-3mm}
        \subcaption{Real-world robotic tasks.}
        \label{fig:robot_task}
    \end{subfigure}
    \hfill
    \begin{subfigure}[b]{0.35\linewidth}
        \centering
        \includegraphics[width=\columnwidth]{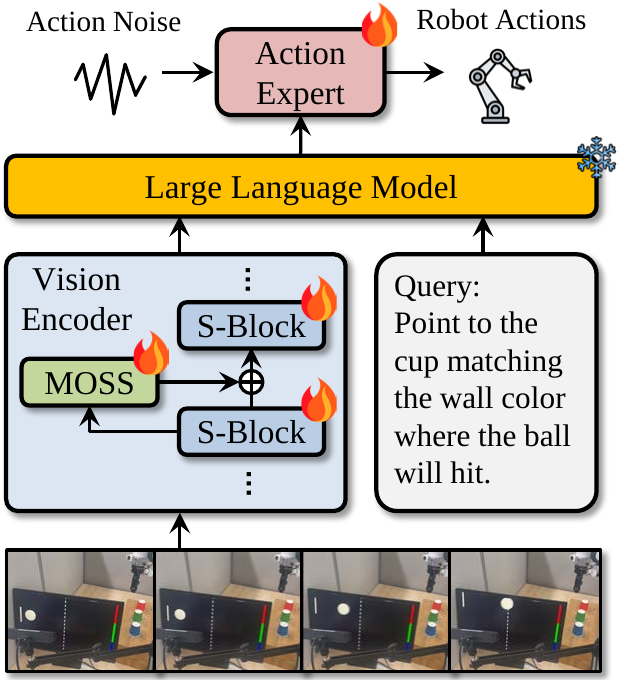}
        \vspace{-3mm}
        \subcaption{ContextVLA with MOSS.}
        \label{fig:robot_model}
    \end{subfigure}
    \caption{\textbf{Proposed real-world robotic tasks (MoveSense, PongPredict, Conveyor PnP) and our VLA model.} (a) visualizes the real-world robotic tasks in dynamic environments. (b) illustrates the overall VLA architecture. MOSS is inserted in the vision encoder and provides motion cues.}
    \label{fig:robot_task_model}
\end{figure}

\end{itemize}

\noindent \textbf{Implementation Details.}
We adopt GR00T-N1.5~\cite{bjorck2025gr00t} and ContextVLA~\cite{jang2025contextvla} as the single-frame and multi-frame baselines, respectively. On top of ContextVLA, we insert a single MOSS module after the 6th layer of the SigLIP vision encoder~\cite{siglip}, and inject its output features into the subsequent layers via a residual connection (Fig.~\ref{fig:robot_model}).

\noindent \textbf{Results.}
In Tab.~\ref{tab:vla_moss}, MOSS achieves notable gains on RoboCasa, a popular simulation benchmark, indicating its effectiveness in general robot manipulation scenarios.
We next evaluate real-world tasks requiring motion understanding, where MOSS improves performance by a significant margin.
The single-frame baseline (GR00T-N1.5) performs the worst, while the multi-frame baseline (ContextVLA) performs somewhat better due to the availability of richer temporal cues.
However, the multi-frame baseline often fails at motion perception tasks, such as motion direction (left vs.\ right) or rotation (CW vs.\ CCW), and frequently misses motion onset and becomes stuck, whereas MOSS achieves near-perfect performance.
Similar trends are observed in the motion prediction task (PongPredict), where baseline models often react too late or struggle under large vertical movements, whereas MOSS performs robustly.
In Conveyor PnP, baseline models tend to collapse to a single speed (GR00T-N1.5) or behave inconsistently across speeds (ContextVLA), whereas MOSS adjusts its action tempo and adapts robustly to varying speeds, generalizing well to unseen speeds.
Considering the multi-frame baseline relies entirely on a few VLM layers for temporal modeling, these results suggest that such implicit temporal modeling is insufficient for capturing temporal dynamics, while highlighting the effectiveness of MOSS.
See Appendix~\ref{sec:supp_impl_vla} for task setups and implementation, Appendix~\ref{sec:supp_visualization} for qualitative results, and Appendices~\ref{sec:supp_vla_single_frame} and~\ref{sec:supp_efficiency} for single-frame VLAs and latency.

\begin{table*}[t]
    \centering
    \caption{\textbf{Results on RoboCasa and real-world robotic tasks.} We report the success rates (\%) of VLAs fine-tuned on the training dataset using official implementations.}
    \scalebox{0.85}{
        \setlength{\tabcolsep}{3pt}
        \begin{tabular}{lc ccccc}
            \toprule
            \multirow{2}{*}{method} & \multirow{2}{*}{\#frames} & \multirow{2}{*}{\makecell{RoboCasa\\Kitchen}}&\multicolumn{2}{c}{MoveSense} & \multirow{2}{*}{\makecell{Pong\\Predict}} &  \multirow{2}{*}{\makecell{Conveyor\\PnP}} \\
             \cmidrule(lr){4-5}
             &  &  &Left/Right & CW/CCW &  & \\
            \Xhline{0.3pt}
            GR00T-N1.5~\cite{bjorck2025gr00t} & 1 & 62.4 &60.0 & 35.4 & 40.7 & 41.7 \\
            ContextVLA~\cite{jang2025contextvla} & 8 & 64.0 &67.5 & 42.7 & 51.9 & 54.2 \\
            \cellcolor{gray!15}$+$ MOSS (ours) & \cellcolor{gray!15}8 & \cellcolor{gray!15}\textbf{66.9} &\cellcolor{gray!15}\textbf{95.8} & \cellcolor{gray!15}\textbf{99.0} & \cellcolor{gray!15}\textbf{81.5} & \cellcolor{gray!15}\textbf{87.5} \\
            \bottomrule
        \end{tabular}
    }
    \label{tab:vla_moss}
    \end{table*}
\vspace{-3mm}
\section{Conclusion}
\vspace{-2mm}

We have provided an in-depth analysis of high-order space-time self-similarities and demonstrated that each order reveals diverse aspects of temporal dynamics.
We introduced MOSS, a lightweight neural module that learns and integrates multi-order STSS features as multi-faceted motion representations.
Through extensive experiments, we validated the broad applicability of MOSS across diverse video understanding benchmarks, motion-centric video VQA, and robotic tasks.
These results highlight the potential of high-order STSS in capturing complex motion patterns, demonstrating its role in comprehensive video understanding.

\newpage
\begin{ack}
This work was supported by the Institute of Information \& communications Technology Planning \& Evaluation (IITP) grants funded by the Korean government (MSIT) (RS-2022-II220959: Few-Shot Learning of Causal Inference in Vision and Language for Decision Making (40\%), RS-2022-II220264: Comprehensive Video Understanding and Generation with Knowledge-based Deep Logic Neural Network (40\%), RS-2024-00457882: National AI Research Lab Project (20\%)).
\end{ack}

{
    \small
    \bibliographystyle{plain}
    \bibliography{reference}
}

\newpage
\appendix

\begin{figure*}[h!]
    \centering
    \begin{subfigure}{\linewidth}
        \centering
        \includegraphics[width=0.99\linewidth]{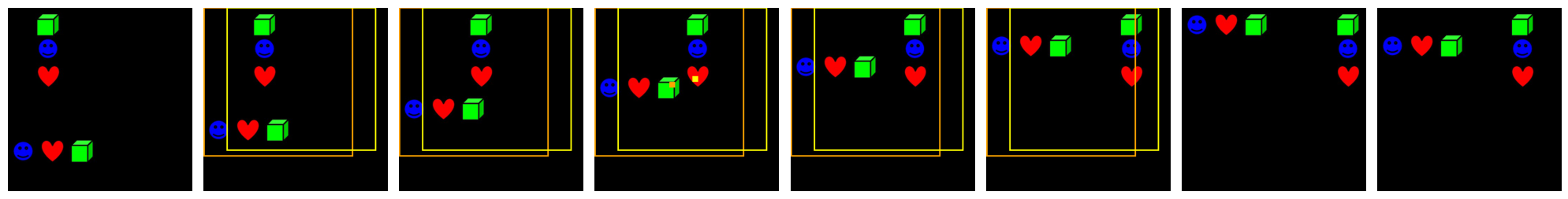}
        \subcaption{RGB frames}
        \label{fig:supp_toy_a}
    \end{subfigure}

    \vspace{2mm}
    
    \begin{subfigure}{0.49\linewidth}
        \centering
        \includegraphics[width=\linewidth]{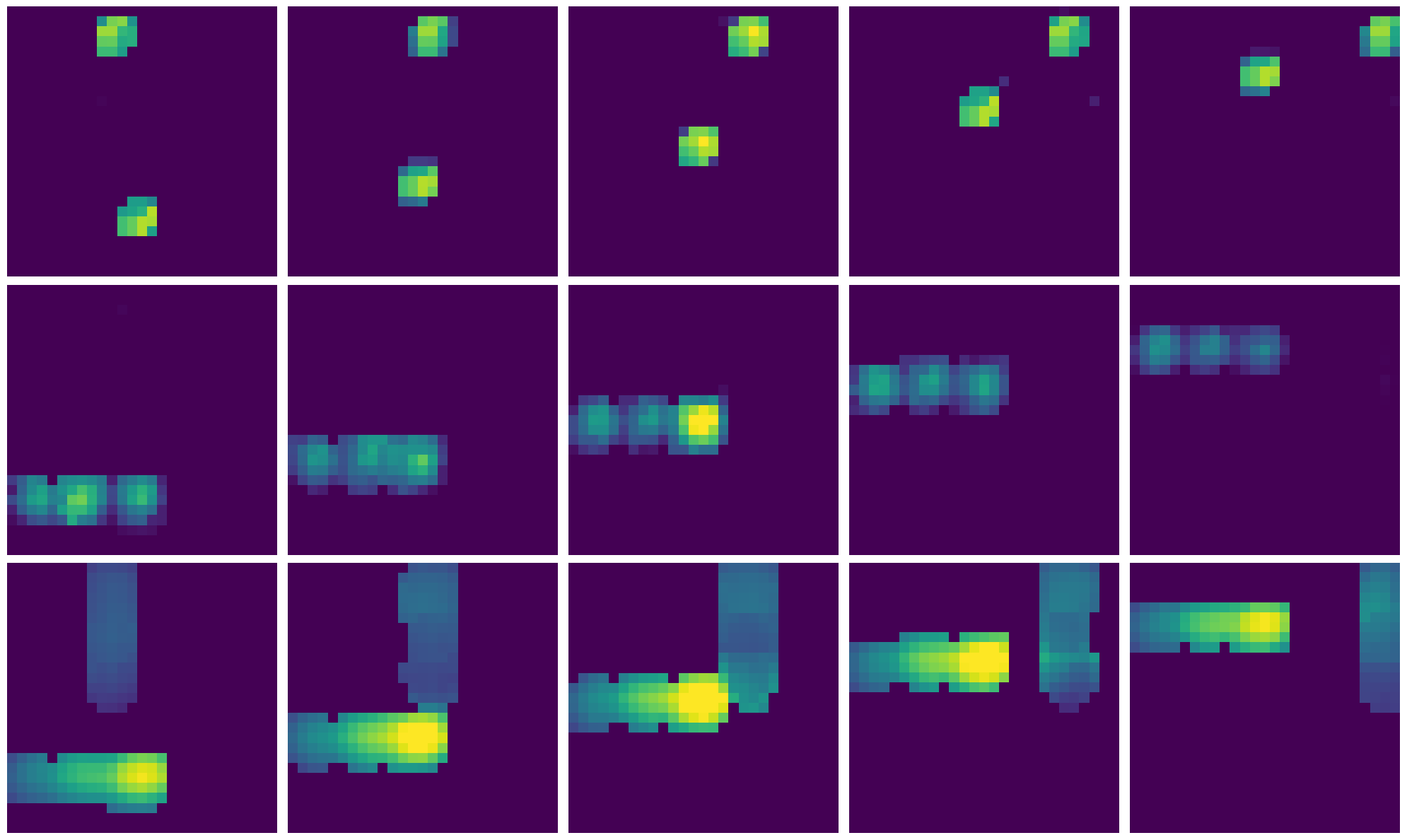}
        \vspace{-4mm}
        \subcaption{STSS of query 1}
        \label{fig:supp_toy_b}
    \end{subfigure}
    \hfill
    \begin{subfigure}{0.49\linewidth}
        \centering
    \includegraphics[width=\linewidth]{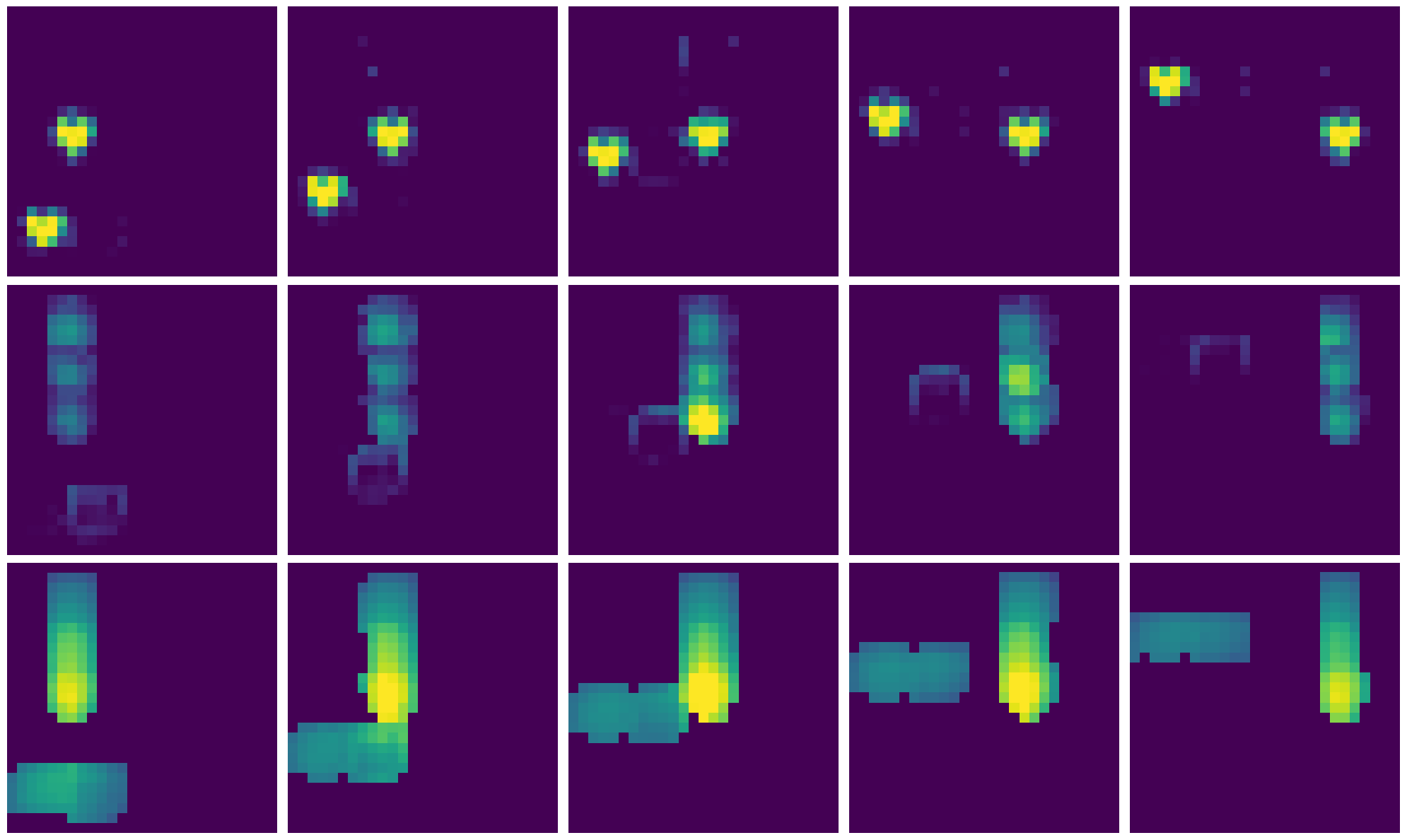}
        \vspace{-4  mm}
        \subcaption{STSS of query 2}
        \label{fig:supp_toy_c}
    \end{subfigure}
    
    \caption{\textbf{Visualization of STSS tensors.} (a) Input RGB frames, where two different queries and their spatio-temporal matching regions. (b) 1st- to 3rd-order STSS maps of the \textcolor{brown}{brown} query. (c) 1st- to 3rd-order STSS maps of the \textcolor{yellow}{yellow} query.}
    \label{fig:sup_toy_visualization}
\end{figure*}

\section{Illustration of High-Order STSS}
We present a toy example with a simplified video clip to clarify the characteristics of high-order STSS in modeling temporal dynamics, as described in Secs.~\ref{sec:revisiting_stss} and \ref{sec:higher_order_stss}.

\noindent \textbf{Experimental Setup.}
We synthesize a controlled video clip with two sets of moving objects, each comprising three objects: a blue circle, a red heart, and a green cube (Fig.~\ref{fig:supp_toy_a}).
Objects within each set share the same motion—either horizontal or vertical movement.
We select two green cubes as queries and visualize their STSS maps from the 1st- to the 3rd-order (Figs.~\ref{fig:supp_toy_b} and~\ref{fig:supp_toy_c}).
Here, we define the STSS encoding function $g$ as vectorization over $(L,U,V)$ dimensions, \ie, $g: \mathbb{R}^{T \times H \times W \times L \times U \times V} \rightarrow \mathbb{R}^{T \times H \times W \times (LUV)}$.

\noindent \textbf{Characteristics of STSS at Different Orders.}
The 1st-order STSS captures the motion flow of the query by establishing correspondences based on appearance across frames.
However, it struggles to distinguish between visually similar objects in different sets, capturing the motion of other unintended objects.
The 2nd-order STSS addresses this limitation by computing similarities based on motion patterns rather than appearance.
It effectively identifies the set of objects that share similar motion with the query, distinguishing visually similar objects moving differently.
The 3rd-order STSS extends this by grouping regions based on these motion segments, highlighting overall motion patterns across all object sets and enabling a higher-level understanding of motion.
Interestingly, despite the potential of modeling group dynamics, we observe that our trained models, in practice, learn to leverage the 3rd-order STSS to highlight motion boundaries of multiple motion segments where distinct motion patterns naturally emerge near (dis)occlusions, as opposed to continuous motion regions (Figs.~\ref{fig:stss_sup1}-\ref{fig:stss_sup5}).
This progression—from capturing motion flows to identifying motion segments to understanding motion at the object set level—reveals diverse aspects of temporal dynamics.

\section{Implementation Details}
\label{sec:implementation_detail_supp}

\subsection{Video Action Recognition.}
\label{sec:supp_impl_ar}

\noindent \textbf{Implementation Details.} Tabs.~\ref{tab:model_config} and \ref{tab:supp_implementation} list the model configurations and training hyperparameters for each model scale and dataset.
All models are trained on 8 NVIDIA RTX 6000 Ada GPUs.

\begin{table}[t]
    \centering
    \captionsetup{width=\columnwidth}
    \setlength\tabcolsep{8.1pt}
    \caption{\textbf{Model configurations}. ``TP" indicates the number of trainable parameters. FLOPs are measured using 8 frames.}
    \scalebox{0.90}{
    \begin{tabular}{l|ccc|ccc|cc}
      \Xhline{0.8pt}
      \multirow{2}*{methods} & \multicolumn{3}{c|}{image encoder} & \multicolumn{3}{c|}{temporal encoder} & FLOPs & TP \\
      ~ & layer & dim & head & layer & dim & head & (G) & (M) \\
      \Xhline{0.2pt}
      MOSS-S & 12 & 768 & 12 & 6 & 192 & 3 & 151 & 6\\
      MOSS-B & 12 & 768 & 12 & 12 & 320 & 5 & 179 & 22\\
      MOSS-M & 24 & 1024 & 16 & 12 & 320 & 5 & 707 & 24 \\
      MOSS-L & 24 & 1024 & 16 & 24 & 448 & 7 & 833 & 82\\
      \Xhline{0.8pt}
    \end{tabular}
    }
    \label{tab:model_config}
\end{table}
  
\begin{table*}[t]
  \centering
  \caption{\textbf{Training configurations on Something-Something V1\&V2, Diving48, FineGym, and Kinetics-400 for different model scales}.}
  \scalebox{0.8}{
  \setlength{\tabcolsep}{1pt}
  \begin{tabular}{@{}lcccccccccc@{}}
    \toprule
    \multirow{2}*{Setting} & \multicolumn{4}{c}{Something V1 \& V2} & \multicolumn{2}{c}{Diving48} & \multicolumn{2}{c}{FineGym} & \multicolumn{2}{c}{Kinetics-400} \\
    ~ & S & B & M & L & B & L & B & L & B & L \\
    \midrule
    \textit{Optimization} \\
    batch size & \multicolumn{3}{c}{128} & 96 & 128 & 80 & 128 & 80 & 128 & 96 \\
    epochs & \multicolumn{3}{c}{40} & 30 & \multicolumn{2}{c}{30} & \multicolumn{2}{c}{30} & \multicolumn{2}{c}{30} \\
    learning rate & \multicolumn{3}{c}{1e-3} & 2e-4 & 1e-3 & 2e-4 & 1e-3 & 2e-4 & 1e-3 & 2e-4 \\
    lr schedule & \multicolumn{10}{c}{cosine decay} \\
    optimizer & \multicolumn{10}{c}{AdamW ($\beta_{1}$ = 0.9, $\beta_{2}$ = 0.999)} \\
    weight decay & \multicolumn{10}{c}{0.15}\\
    warmup epochs & \multicolumn{10}{c}{4} \\
    pre-train & \multicolumn{4}{c}{CLIP400M} & \multicolumn{2}{c}{CLIP400M+K400} & \multicolumn{2}{c}{CLIP400M+K400} & \multicolumn{2}{c}{CLIP400M} \\
    \midrule
    \textit{Augmentation} \\
    sampling & \multicolumn{4}{c}{uniform} & \multicolumn{2}{c}{uniform} & \multicolumn{2}{c}{uniform} & \multicolumn{2}{c}{dense (stride=8)} \\
    resize & \multicolumn{10}{c}{RandomResizedCrop} \\
    RandAugment & \multicolumn{4}{c}{\small{rand-m7-n4-mstd0.5-inc1}} & \multicolumn{2}{c}{-} & \multicolumn{2}{c}{-} & \multicolumn{2}{c}{-}\\
    random flip & \multicolumn{10}{c}{0.5}\\ 
    label smoothing & \multicolumn{10}{c}{0.1}\\
    repeated Aug. & \multicolumn{4}{c}{2} & \multicolumn{2}{c}{1} & \multicolumn{2}{c}{1} & \multicolumn{2}{c}{2} \\
    gray scale & \multicolumn{4}{c}{-} & \multicolumn{2}{c}{-} & \multicolumn{2}{c}{-} & \multicolumn{2}{c}{0.2}\\
    \midrule
    \textit{MOSS module} \\
    window $(L,U,V)$ & \multicolumn{10}{c}{$(5,9,9)$}\\
    \# channels $D$ & \multicolumn{2}{c}{64} & \multicolumn{2}{c}{96} & 64 & 96 & 64 & 96 & 64 & 96 \\
    \# Enc. Blocks & \multicolumn{10}{c}{3}\\
    position $k$ & \multicolumn{2}{c}{4} & \multicolumn{2}{c}{8} & 4 & 8 & 4 & 8 & 4 & 8 \\
    \bottomrule
  \end{tabular}}
  \vspace{-1mm}
  \label{tab:supp_implementation}
\end{table*}

\noindent \textbf{License.}
We implement our model based on Side4Video~\cite{side4video} in PyTorch\footnote{\url{https://github.com/HJYao00/Side4Video}} under MIT license. The datasets used in our experiments are publicly available for academic research. Kinetics-400 is available under CC BY 4.0 license, Something-Something uses a custom academic license\footnote{\url{https://www.qualcomm.com/content/dam/qcomm-martech/dm-assets/documents/jester_something_something_exercise_research_license_final_qti_28jul2022.pdf}}, Diving48's license is unknown, and FineGym is under CC BY-NC 4.0 license.

\subsection{Motion Understanding in Video MLLMs.}
\label{sec:supp_impl_videollm}
\noindent \textbf{Implementation Details.} We adopt VideoLLaMA3-2B~\cite{videollama3} as our baseline and insert a single MOSS module after the 6th layer of the SigLIP vision encoder~\cite{siglip} to extract multi-order STSS features. The resulting features are added before the projector to inject early motion cues into the LLM.
We set $(L, U, V)=(7, 11, 11)$, and initialize the final FC layers to zeros to stabilize early training.
The model is fine-tuned using LoRA with learning rates of 1e-3, 1e-4, and 1e-5 for MOSS, projector, and LLM LoRA weights, respectively, over 1,000 iterations.
For both training and testing, all frames are resized such that the shorter side is 224 and sampled at 2 FPS. All models are trained on 8 NVIDIA RTX 6000 Ada GPUs.

\noindent \textbf{License.} Our implementation is built upon VideoLLaMA3-2B~\cite{videollama3} codebase\footnote{\url{https://github.com/DAMO-NLP-SG/VideoLLaMA3/tree/main}} under the Apache 2.0 license.
The benchmarks used in our evaluation, MotionBench~\cite{motionbench} and FAVOR-Bench~\cite{favorbench}, are also released under the Apache 2.0 license.

\subsection{Motion Understanding in VLAs.}
\label{sec:supp_impl_vla}
\noindent \textbf{Detailed RoboCasa Results.}
We provide the full RoboCasa results in Tab.~\ref{tab:robocasa}.

\noindent \textbf{Implementation Details.} We adopt GR00T-N1.5~\cite{bjorck2025gr00t} as the single-frame baseline model.
ContextVLA~\cite{jang2025contextvla} extends a single-frame VLA to a multi-frame setting without additional parameters by compressing multi-frame embeddings within the VLM. We adopt ContextVLA built on GR00T-N1.5 as our multi-frame baseline model.
On top of ContextVLA, we insert a single MOSS block after the 6th layer of the SigLIP vision encoder~\cite{siglip}, and inject its output features into the subsequent layers via a residual connection.
We set $(L,U,V)=(5,9,9)$.
In the multi-frame setting, we use 8 consecutive frames sampled at 10 FPS.
For RoboCasa, we train all models for 60K iterations using the AdamW optimizer with a batch size of 32.
For real-world tasks, we train all models for 30K iterations using the AdamW optimizer with a batch size of 64.
We adopt cosine learning rate schedule with 5\% warmup.
For training, we collect 64 episodes for MoveSense, each containing three interactions.
We collect 50 episodes for PongPredict, each containing at least five interactions.
For Conveyor PnP, we collect 40 episodes across two different speeds, with two interactions per episode.
For evaluation, we assess each model over 96 trials for MoveSense and 54 trials for PongPredict.
For Conveyor PnP, we evaluate each model over 24 trials across three speeds (S1, S2, S3), with 8 trials per speed. S1 and S3 are seen during training, while S2 is an unseen speed between them.

\begin{figure}[t]
    \centering
        \centering
        \includegraphics[width=\columnwidth]{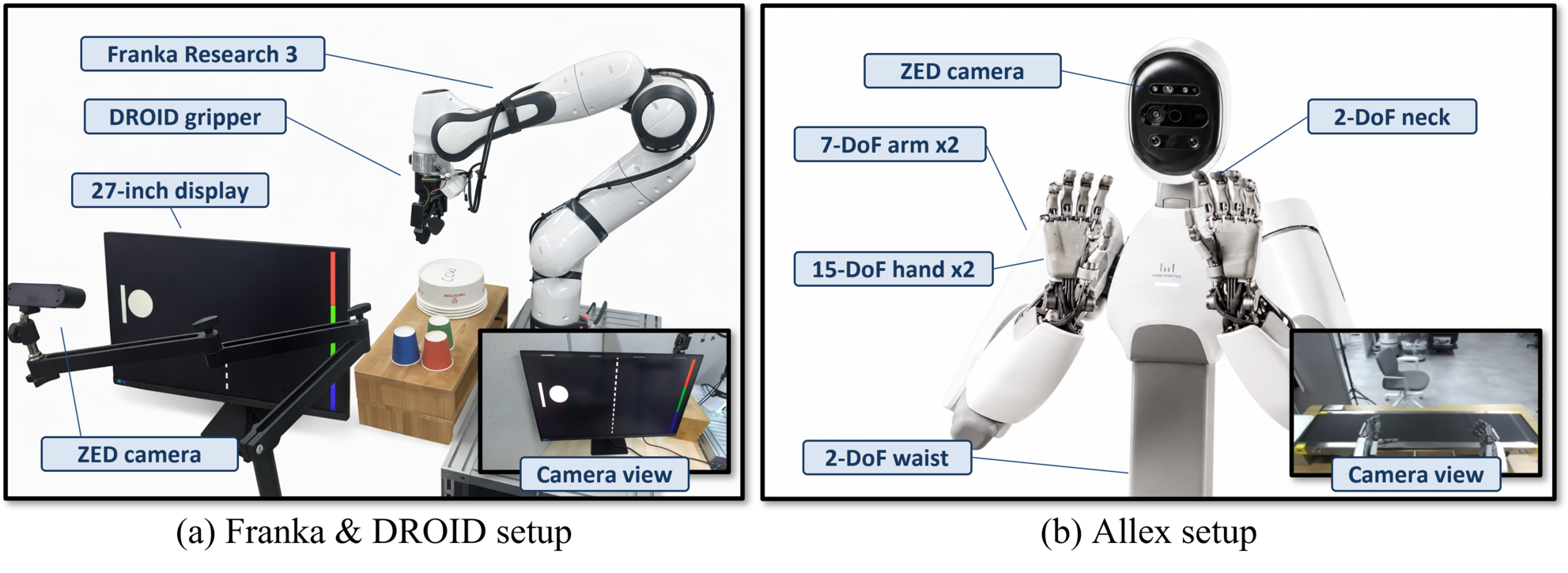}
    \caption{\textbf{Real-robot platform.} We specify the robot specifications and other environment settings.}
    \label{fig:robot_setup}
\end{figure}

\noindent \textbf{Real-World Robot Setup.} In Fig.~\ref{fig:robot_setup}, we provide the detailed setup for real-world robotic tasks. For \textit{MoveSense} and \textit{PongPredict}, we use a Franka Research 3 robot arm and a DROID gripper. A single ZED camera with a resolution of 720$\times$1280 is mounted on the side, facing the 27-inch display. We use the visual observation after resizing it to 224$\times$224 resolution. The VLA model is provided with a view containing only the display, with the robot’s motion intentionally excluded. We use absolute joint position as action space.
For \textit{Conveyor Pick-and-Place}, we use an Allex humanoid, a high-DoF upper-body robot.
We utilize 48 joints for manipulation, consisting of dual 7-DoF arms, two 15-DoF hands, a 2-DoF waist, and a 2-DoF neck.
A head-mounted ZED camera provides an egocentric view.

The text instructions used for each task are as follows.
\begin{itemize}
    \item MoveSense (Left/Right): Point left when the circle on the display moves left, and point right when it moves right.
    \item MoveSense (CW/CCW): Point right when the circle on the display moves clockwise, and point left when it moves counterclockwise.
    \item PongPredict: Point to the cup matching the color of the wall the ping pong ball will reach on the right.
    \item Conveyor Pick-and-Place: When the box appears on the conveyor belt, pick it up with your right hand and place it on the shelf in front.
\end{itemize}

\noindent \textbf{License.} Our implementation is built on GR00T-N1.5~\cite{bjorck2025gr00t}\footnote{\url{https://github.com/NVIDIA/Isaac-GR00T/tree/main}}, which is released under the Apache 2.0 license.
The license of the public ContextVLA~\cite{jang2025contextvla} implementation is unspecified.
RoboCasa, used as our simulation benchmark, is released under the MIT license.

\section{Additional Experiments}
\label{sec:experiments_supp}
This section presents additional results, including temporal action detection, generic event boundary detection, ablation studies on action recognition, a comparison with self-attention, evaluations on longer videos and sparse frame inputs, integration into single-frame VLAs, and inference latency.

\subsection{Video Action Recognition on Kinetics-400}
\label{sec:experiments_k400}

\begin{table}[t]
  \centering
  \caption{\textbf{Results on Kinetics-400}. $^\dag$ trained with text supervision. ``Input" indicates \#frames$\times$\#crops$\times$\#clips.}
  \vspace{1mm}
  \scalebox{0.9}{
  \setlength{\tabcolsep}{12pt}
  \begin{tabular}{lccc}
    \Xhline{0.8pt}
    method & input & TFLOPs  & top1 \\
    \Xhline{0.2pt}
    \textit{Full finetuning} & & & \\
    \gray{ViViT-H~\cite{vivit}} & \gray{32$\times$12} & \gray{3.98$\times$12} & \gray{84.9} \\
    \gray{MTV-H~\cite{mtv}} & \gray{32$\times$12} & \gray{3.71$\times$12} & \gray{85.8} \\
    \gray{$\text{XCLIP-L}^\dag$~\cite{xclip}} & \gray{16$\times$12} & \gray{3.09$\times$12}  & \gray{87.7} \\
    \gray{$\text{Text4Vis-L}^{\dag}$~\cite{wu2024transferring}} & \gray{32$\times$12} &\gray{ 1.66$\times$12}  & \gray{87.6} \\
    \gray{ATM ViT-L~\cite{aim}} & \gray{32$\times$12} & \gray{1.68$\times$12} & \gray{88.0} \\
    \Xhline{0.2pt}
    \textit{Frozen backbone} & & & \\
    V-JEPA ViT-H~\cite{vjepa} & 16$\times$6 & - & 84.5 \\
    V-JEPA 2 ViT-H~\cite{vjepa2} & 16$\times$6 & - & 85.3 \\
    M$^2$CLIP ViT-B~\cite{m2clip} & 32$\times$12 & 0.8$\times$12 & 84.1 \\
    OmniCLIP ViT-B~\cite{omniclip} & 8$\times$12 & 0.1$\times$12 & 84.1 \\
    ST-Adapter ViT-L~\cite{stadapter} & 32$\times$3 & 2.75$\times$3 & 87.2 \\
    $\text{CLIP4Vis ViT-L}^{\dag}$~\cite{clip4vis}& 8$\times$12 & 0.42$\times$12 & 87.4 \\
    AIM ViT-L~\cite{aim} & 32$\times$3 & 3.74$\times$3 & 87.5 \\
    $\text{DiST ViT-L}^{\dag}$~\cite{dist} & 32$\times$3 & 2.83$\times$3 & \textbf{88.0} \\
    Side4Video-B~\cite{side4video} & 32$\times$12 & 0.72$\times$12 & 84.2 \\
    Side4Video-L~\cite{side4video} & 16$\times$12 & 1.74$\times$12 & 87.0 \\
    \rowcolor{gray!15} MOSS-B (ours) & 32$\times$12 & 0.72$\times$12 & 85.2 \\   
    \rowcolor{gray!15} MOSS-L (ours) & 16$\times$12 & 1.67$\times$12 & \underline{87.7} \\
    \Xhline{0.8pt}
  \end{tabular}
  }
  \label{tab:k400}
\end{table}

In Tab.~\ref{tab:k400}, our method also demonstrates its effectiveness on Kinetics-400, which is an appearance-centric benchmark.
MOSS-B and MOSS-L achieve 85.2\% and 87.7\% top-1 accuracies, improving over the baseline by 1.0\%p and 0.7\%p, respectively, competitive with other methods~\cite{vivit,mtv,xclip,atm,evl,stadapter,aim,dist,vjepa,side4video,m2clip,omniclip,wu2024transferring,vjepa2}.
This validates the generalizability of our high-order STSS features, which can be effectively leveraged across diverse video domains.

\subsection{Temporal Action Detection}
\label{sec:experiments_tad}
\noindent \textbf{Datasets.}
\textit{THUMOS-14}~\cite{thumos} is a widely used benchmark for temporal action detection (TAD), containing 200 and 213 untrimmed videos for training and testing, annotated with 20 action classes.

\noindent \textbf{Implementation Details.}
We adopt AdaTAD~\cite{adatad} as our baseline, which adapts VideoMAE to a TAD model with \textit{TIA adapters}.
We insert a single MOSS module between the 6th VideoMAE block and the TIA adapter. Following \cite{adatad}, we train only the MOSS module and TIA adapters.
We use 768 input frames with a resolution of 224$\times$224 for training.
For evaluation, we report the mean Average Precision (mAP) at different temporal Intersection over Union (tIoU) thresholds.

\begin{table}[t]
    \centering
    \caption{\textbf{Results of temporal action detection on THUMOS-14}. We report mAP with tIoU thresholds from 0.3 to 0.7 and their average. * reproduced by out setup.}
    \vspace{1mm}
    \setlength{\tabcolsep}{4pt}
    \scalebox{0.90}{
    \begin{tabular}{l|c|cccccc}
    \Xhline{0.8pt}
    \multirow{2}{*}{method} & \multirow{2}{*}{backbone} & \multicolumn{6}{c}{THUMOS-14} \\
    & & 0.3 & 0.4 & 0.5 & 0.6 & 0.7 & Avg. \\
    \Xhline{0.2pt}
    TALLFormer~\cite{tallformer} & VSwin-B & 76.0 & - & 63.2 & - & 34.5 & 59.2 \\
    ActionFormer~\cite{actionformer} & I3D & 82.1 & 77.8 & 71.0 & 59.4 & 43.9 & 66.8 \\
    TriDet~\cite{tridet} & I3D & 83.6 & 80.1 & 72.9 & 62.4 & 47.4 & 69.3 \\
    AdaTAD~\cite{adatad} & VMAE-B & 87.0 & 82.4 & 75.3 & 63.8 & 49.2 & 71.5 \\
    AdaTAD~\cite{adatad} & VMAE-L & 87.7 & 84.1 & 76.7 & 66.4 & 52.4 & 73.5 \\
    \Xhline{0.2pt}
    AdaTAD*~\cite{adatad} & VMAE-B & 85.7 & 82.3 & 75.3 & 63.4 & 50.1 & 71.4 \\
    \rowcolor{gray!15}$+$MOSS & VMAE-B & 87.3 & 82.9 & 75.7 & 64.8 & 50.3 & \textbf{72.2} \\
    AdaTAD*~\cite{adatad} & VMAE-L & 87.5 & 83.7 & \textbf{77.6} & 66.5 & 51.9 & 73.5 \\
    \rowcolor{gray!15}$+$MOSS & VMAE-L & \textbf{88.1} & \textbf{84.5} & \textbf{77.6} & \textbf{67.7} & \textbf{52.8} & \textbf{74.1} \\
    \Xhline{0.8pt}
    \end{tabular}
    }
    \label{tab:tad}    
\end{table}
\begin{table}[t]
    \centering
    \caption{\textbf{Results of GEBD on Kinetics-GEBD and TAPOS}. * reproduced by our setup. ``Avg. F1" is an averaged F1 score with relative distance thresholds from 0.05 to 0.5. with 0.05 interval.}
    \vspace{1mm}
    \setlength{\tabcolsep}{4pt}
    \scalebox{0.90}{
    \begin{tabular}{l|cc|cc}
    \Xhline{0.8pt}
    \multirow{2}{*}{method} & \multicolumn{2}{c|}{Kinetics-GEBD} & \multicolumn{2}{c}{TAPOS} \\
    & F1@0.05 & Avg. F1 & F1@0.05 & Avg. F1 \\
    \Xhline{0.2pt}
    Temporal Perceiver~\cite{temporal_perceiver} & 74.8 & 86.0 & 55.2 & 73.2 \\
    DDM-Net~\cite{ddmnet} & 76.4 & 87.3 & 60.4 & 72.8 \\
    SC-Transformer~\cite{sctransformer} & 77.7 & 88.1 & 61.8 & 74.2 \\
    BasicGEBD~\cite{efficientgebd} & 76.8 & 86.6 & 60.0 & 71.0 \\
    EfficientGEBD~\cite{efficientgebd} & 78.3 & 88.3 & 63.1 & 74.8 \\
    \Xhline{0.2pt}
    BasicGEBD*~\cite{efficientgebd} & 76.9 & 86.4 & 60.1 & 71.2 \\
    \rowcolor{gray!15}BasicGEBD$+$MOSS & \textbf{79.4} & \textbf{89.0} & \textbf{68.5} & \textbf{76.6} \\
    \Xhline{0.8pt}
    \end{tabular}}
    \label{tab:gebd}    
\end{table}

\noindent \textbf{Results.}
In Tab.~\ref{tab:tad}, we summarize the TAD results on THUMOS-14. 
Our MOSS-enhanced model consistently outperforms the AdaTAD baseline across all tIoU thresholds, reaching 72.2\% and 74.1\% average mAP with VideoMAE-B and L backbones respectively.
These results demonstrate the versatility of the MOSS module for longer video understanding.

\subsection{Generic Event Boundary Detection}
\label{sec:experiments_gebd}
Generic event boundary detection (GEBD) aims to localize event boundaries in a video, such as changes in subject appearance or motion patterns, segmenting it into distinct and meaningful chunks.
Accurately detecting these instantaneous changes is crucial for effective event segmentation.

\noindent \textbf{Datasets.}
\textit{Kinetics-GEBD}~\cite{kineticsgebd} is the largest GEBD dataset, consisting of 55K videos with 1.3M taxonomy-free event boundaries including action and object changes. 
\textit{TAPOS}~\cite{tapos} comprises 15K Olympic sports videos with 21 distinct action classes. Following~\cite{kineticsgebd}, we re-design TAPOS for the GEBD task by trimming each action instance.

\noindent \textbf{Implementation Details.}
We employ BasicGEBD-L4~\cite{efficientgebd} as the backbone and add a single MOSS module after the 2nd stage of ResNet-50 and train the entire network end-to-end following the training protocols in~\cite{efficientgebd}. For evaluation, we measure F1 score with relative distance 0.05 and average F1 score from 0.05 to 0.5 with 0.05 interval.

\noindent \textbf{Results.}
In Tab.~\ref{tab:gebd}, we present the results on Kinetics-GEBD and TAPOS.
Compared to our baseline (BasicGEBD), our MOSS module substantially improves performance at F1@0.05 scores increasing by 2.5\%p and 8.4\%p on Kinetics-GEBD and TAPOS, respectively, achieving new state-of-the-art results.
These improvements demonstrate that our proposed module effectively captures fine-grained temporal changes in both motion and objects, which is crucial for accurate event boundary detection.

\begin{table*}[t]
  \centering
  \setlength\tabcolsep{3pt}
  \captionsetup{width=\linewidth}

  \caption{\textbf{Additional ablation studies on Something-Something V1 and Diving48.} All experiments are conducted with MOSS-S taking 8 and 32 frames as input, respectively. ``FLOPs", ``TP", and ``Mem" respectively indicate FLOPs (G), trainable parameters (M), and memory footprint (GB) using 8 frames. Memory footprint is measured using a batch size of 32 for a single GPU machine.}

\begin{subtable}[t]{0.493\linewidth}
    \centering
    \subcaption{Temporal window size $L$}
    \setlength\tabcolsep{2pt}
    \vspace{-2mm}
      \scalebox{0.88}{
      \begin{tabular}{L{0.9cm}L{0.9cm}|C{1.1cm}C{0.7cm}C{0.8cm}|C{0.9cm}C{0.7cm}}
      \Xhline{0.8pt}
      1st & 2nd & FLOPs & TP & Mem & SSV1 & D48 \\
      \Xhline{0.3pt}
      3 & 5 & 150.9 & 5.4 & 9.8 & 59.5 & 87.0\\
      \cellcolor{gray!15}5 & \cellcolor{gray!15}5 & \cellcolor{gray!15}151.5 & \cellcolor{gray!15}5.6 & \cellcolor{gray!15}9.9 & \cellcolor{gray!15}\textbf{60.0} & \cellcolor{gray!15}87.7 \\
      7 & 5 & 152.0 & 5.7 & 10.5 & 59.9 & 87.8 \\
      5 & 3 & 150.9 & 5.4 & 9.8 & 59.2 & 86.5 \\
      5 & 7 & 152.0 & 5.7 & 10.5 & 59.7 & \textbf{88.0} \\
      \Xhline{0.8pt}
      \end{tabular}
      }
      \label{tab:ablation_window}
\end{subtable}
\hfill
\begin{subtable}[t]{0.493\linewidth}
    \centering
    \subcaption{Module positions}
    \setlength\tabcolsep{2pt}
    \vspace{-2mm}
    \centering
    \scalebox{0.88}{
    \begin{tabular}{L{2.0cm}|C{1.1cm}C{0.7cm}C{0.8cm}|C{0.9cm}C{0.7cm}}
    \Xhline{0.8pt}
    pos & FLOPs & TP & Mem & SSV1 & D48 \\
    \Xhline{0.3pt}
    2 & 151.5 & 5.6 & 9.9 & 59.1 & 85.4\\
    \cellcolor{gray!15}4 & \cellcolor{gray!15}151.5 & \cellcolor{gray!15}5.6 & \cellcolor{gray!15}9.9 & \cellcolor{gray!15}60.0 & \cellcolor{gray!15}87.7 \\
    6 & 151.5 & 5.6 & 9.9 & 58.9 & 86.5 \\
    8 & 151.5 & 5.6 & 9.9 & 58.2 & 86.4 \\
    4,6 & 154.5 & 6.7 & 11.9 & \textbf{60.1} & 88.1 \\
    \Xhline{0.8pt}
    \end{tabular}
    }
    \label{tab:ablation_position}
\end{subtable}

    \label{tab:ablation_supp}
\end{table*}

\subsection{Additional Ablation Experiments}
\label{sec:supp_ablation}
We present additional ablation studies on Something-Something V1 and Diving48.
Unless otherwise specified, we follow the experimental settings in Secs.~\ref{sec:ablation} and \ref{sec:implementation_detail_supp}.

\noindent \textbf{Temporal Window Size $L$.}
In Tab.~\ref{tab:ablation_window}, we examine the effect of the size of temporal window $L$ for STSS transformation while keeping the spatial window size fixed as $(U,V)=(9,9)$.
We first vary the temporal window size of the 1st-order STSS keeping that of the 2nd-order STSS constant.
We observe that increasing $L$ from 3 to 5 improves performance by capturing longer-range temporal dynamics.
However, performance saturates when $L$ exceeds 5, providing no significant additional gains.
Similarly, varying the temporal window size of the 2nd-order STSS while fixing that of the 1st-order yields comparable results. 
Based on these results, we set the temporal window size $L=5$ for both the 1st- and 2nd-order STSS.

\noindent \textbf{Module Position.}
In Tab.~\ref{tab:ablation_position}, we examine the effect of different positions and the numbers of the MOSS module.
The results show that the MOSS module is beneficial for all the cases but the performance depends on the position of the module.
The MOSS module inserted after the 4th image encoder block performs the best.
We interpret these results as a trade-off between the robustness of the STSS transformation and the effectiveness of temporal modeling;
inserting the module too early may lead to noisy STSS transformation due to insufficient visual semantics in the feature maps, whereas inserting the module too late limits the capacity for temporal modeling because fewer temporal blocks remain to process the enriched features.
Given the marginal gain of using multiple modules, we add a single module after the 4th block by default for efficiency.

\subsection{Difference from Self-Attention}
\label{sec:supp_self_attention}
Self-attention and MOSS both compute STSS, but they differ in how the computed STSS is used, and this leads to heterogeneous output representations that learn fundamentally different information for temporal modeling.
The STSS itself suppresses appearance and emphasizes the structure of correspondences, which directly reveals motion (Sec.~\ref{sec:revisiting_stss}, Figs.~\ref{fig:main_toy_visualization} and \ref{fig:stss}).
MOSS transforms this structural pattern directly into a feature, so the representation remains in the space of correspondence structures, and stacking it composes structure upon structure---motion flows into motion segments, and motion segments into their layouts.
Self-attention, in contrast, decomposes the same STSS into attention weights to aggregate value embeddings.
The intermediate representation therefore returns to appearance features, and the next layer recomputes STSS from these appearance features rather than progressively composing correspondence structures.
Stacking self-attention thus deepens appearance relations rather than producing a \textit{similarity of a similarity}, which is precisely why increasing depth does not reproduce what emerges across orders of STSS.

\begin{figure}[t]
    \centering
    \setbox0=\vbox{\hsize=0.47\linewidth \linewidth=\hsize \parskip=0pt \parindent=0pt
        \captionsetup{position=top}%
        \captionof{table}{\textbf{Effect of using STSS as a feature vs.\ as attention weights}, with all other components fixed (MOSS-S, 8 frames, Something-Something V1). ``\# encoders'' denotes the number of STSS encoders, and ``FLOPs'' and ``TP'' indicate FLOPs (G) and trainable parameters (M), respectively.}%
        \label{tab:supp_stss_usage}\par
        \vfill
        {\centering
        \setlength\tabcolsep{3pt}
        \scalebox{0.85}{%
        \begin{tabular}{l|c|cc|c}
        \Xhline{0.8pt}
        STSS usage & \# encoders & FLOPs & TP & SSV1 \\
        \Xhline{0.3pt}
        -- (baseline) & -- & 148.4 & 4.5 & 56.9 \\
        \Xhline{0.3pt}
        \multirow{2}{*}{as attention weights} & 1 & 150.1 & 5.1 & 57.6 \\
         & 2 & 151.7 & 5.7 & 57.7 \\
        \Xhline{0.3pt}
        \multirow{2}{*}{as a feature (ours)} & 1 & 150.0 & 5.1 & 59.0 \\
         & \cellcolor{gray!15}2 & \cellcolor{gray!15}151.5 & \cellcolor{gray!15}5.6 & \cellcolor{gray!15}\textbf{60.0} \\
        \Xhline{0.8pt}
        \end{tabular}%
        }\par}
        \vskip 0pt}
    \setbox2=\vbox{\hsize=0.49\linewidth \linewidth=\hsize \parskip=0pt \parindent=0pt
        {\centering
        \begin{minipage}[b]{0.37\linewidth}
            \centering
            \includegraphics[width=\linewidth]{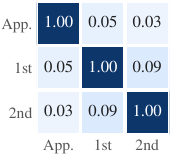}\\[0.5mm]
            {\footnotesize (a) Feature (ours)}
        \end{minipage}
        \hspace{0.08\linewidth}%
        \begin{minipage}[b]{0.37\linewidth}
            \centering
            \includegraphics[width=\linewidth]{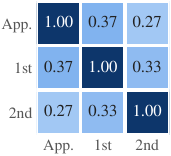}\\[0.5mm]
            {\footnotesize (b) Attention weights}
        \end{minipage}\par}
        \vfill
        \captionsetup{position=bottom}%
        \caption{\textbf{CKA between the encoder outputs} of the two-level models in Tab.~\ref{tab:supp_stss_usage}, with STSS used (a) as a feature (ours) and (b) as attention weights. ``App.'' denotes the appearance features $\mathbf{F}$ fed into the module.}%
        \label{fig:supp_cka}\par
        \vskip 0pt}
    \dimen0=\ht0 \ifdim\ht2>\dimen0 \dimen0=\ht2 \fi
    \hbox to\linewidth{\vbox to\dimen0{\unvbox0}\hfill\vbox to\dimen0{\unvbox2}}
\end{figure}

\noindent \textbf{Experimental Setup.}
To validate our claim, we keep our module entirely intact and change only how the STSS tensor $\mathbf{S} \in \mathbb{R}^{T \times H \times W \times L \times U \times V}$ is used.
For each temporal offset $l$, the $(U,V)$ similarity map is either transformed directly into a feature by an FC layer (ours), or used as attention weights, together with relative positional encoding over the window, to aggregate the value embeddings at the corresponding positions.
Everything downstream, \eg, the $(H,W)$ refinement and the $L$ fusion in the STSS encoder (Fig.~\ref{fig:main_b}), is identical in both.
All experiments are conducted with MOSS-S taking 8 frames on Something-Something V1.

\noindent \textbf{Results.}
As shown in Tab.~\ref{tab:supp_stss_usage}, with a single encoder, transforming the similarity into a feature lifts the baseline by 2.1\%p, whereas using it as attention weights improves over the baseline by only 0.7\%p.
This demonstrates that representing self-similarity directly as a feature is substantially more effective for temporal modeling than using it as aggregation weights.
The second encoder further highlights the difference: the attention variant saturates (+0.1\%p) while ours gains a further 1.0\%p, indicating that the gain stems from the complementarity of STSS rather than using more parameters.

\noindent \textbf{Representation Analysis.}
To understand why only our model continues to benefit from higher orders, we analyze the encoder representations of the two-level models in Tab.~\ref{tab:supp_stss_usage} using centered kernel alignment (CKA)~\cite{kornblith2019similarity}.
As shown in Fig.~\ref{fig:supp_cka}, when STSS is used as a feature, the two encoders yield representations with very low CKA similarity to the appearance features (0.03--0.05) and to one another (0.09), indicating that each order carries distinct information.
When the same STSS is used as attention weights, the outputs remain highly correlated with the appearance features (0.27--0.37) and with each other (0.33), \ie, successive levels learn largely redundant representations rather than distinct higher-order ones.
Stacking layers therefore adds capacity but not new information; higher-order STSS helps because the 1st- and 2nd-order STSS encode complementary structures, which is an inductive bias of how the similarity is represented, not a consequence of depth.

\subsection{MOSS with Longer Videos and Sparse Frame Input}
\label{sec:supp_long_video}
Although MOSS models motion over a span of only 3--4 seconds ($L=7$ at 2 FPS) in our video MLLM setting, improved local motion representations can still benefit long-video understanding when integrated over time.
Recent studies suggest that video MLLMs increasingly excel at long-context aggregation but still struggle with temporal reasoning requiring fine-grained motion understanding~\cite{yashima2026remora,imam2025can,wang2025videostr,shi2026causality}, indicating that strengthening local motion representations may provide a complementary direction.
We evaluate this hypothesis on MotionBench and FAVOR-Bench, two motion-centric VQA benchmarks with videos of diverse durations, by stratifying the evaluation by video length.

\begin{table}[t]
    \centering
    \caption{\textbf{Accuracy on MotionBench and FAVOR-Bench across (a) video durations and (b) input frame rates.} VideoLLaMA3-2B denotes the model fine-tuned on FAVOR-Train (Tab.~\ref{tab:videollm}).}
    \setlength\tabcolsep{4pt}
    \def\durbody{%
        \Xhline{0.8pt}
        benchmark & duration & \# samples & VideoLLaMA3-2B & $+$ MOSS (ours) \\
        \Xhline{0.3pt}
        \multirow{4}{*}{MotionBench~\cite{motionbench}} & 0--10\,s & 2726 & 51.7 & \cellcolor{gray!15}\textbf{54.9} \\
         & 10--30\,s & 1238 & 50.8 & \cellcolor{gray!15}\textbf{52.6} \\
         & $>$30\,s & \phantom{00}54 & 51.9 & \cellcolor{gray!15}\textbf{55.6} \\
         \hhline{~|--|--}
         & overall & 4018 & 51.4 & \cellcolor{gray!15}\textbf{54.2} \\
        \Xhline{0.3pt}
        \multirow{4}{*}{FAVOR-Bench~\cite{favorbench}} & 0--10\,s & 2921 & 43.0 & \cellcolor{gray!15}\textbf{44.3} \\
         & 10--30\,s & 2362 & 47.8 & \cellcolor{gray!15}\textbf{48.7} \\
         & $>$30\,s & 2901 & 46.2 & \cellcolor{gray!15}\textbf{47.3} \\
         \hhline{~|--|--}
         & overall & 8184 & 45.5 & \cellcolor{gray!15}\textbf{46.6} \\
        \Xhline{0.8pt}}
    \def\fpsbody{%
        \Xhline{0.8pt}
        \multirow{2}{*}{benchmark} & \multirow{2}{*}{method} & \multicolumn{4}{c}{FPS} \\
        \cline{3-6}
         & & 4 & 2 & 1 & 0.5 \\
        \Xhline{0.3pt}
        \multirow{2}{*}{MotionBench~\cite{motionbench}} & VideoLLaMA3-2B & 51.8 & 51.4 & 50.8 & 49.6 \\
         & \cellcolor{gray!15}$+$ MOSS (ours) & \cellcolor{gray!15}\textbf{55.4} & \cellcolor{gray!15}\textbf{54.2} & \cellcolor{gray!15}\textbf{53.3} & \cellcolor{gray!15}\textbf{51.5} \\
        \Xhline{0.3pt}
        \multirow{2}{*}{FAVOR-Bench~\cite{favorbench}} & VideoLLaMA3-2B & 45.6 & 45.5 & 44.7 & 43.9 \\
         & \cellcolor{gray!15}$+$ MOSS (ours) & \cellcolor{gray!15}\textbf{46.8} & \cellcolor{gray!15}\textbf{46.6} & \cellcolor{gray!15}\textbf{45.5} & \cellcolor{gray!15}\textbf{44.1} \\
        \Xhline{0.8pt}}
    \setbox0=\hbox{\begin{tabular}{l|lc|cc}\durbody\end{tabular}}
    \setbox2=\hbox{\begin{tabular}{l|l|cccc}\fpsbody\end{tabular}}
    \dimen6=\wd0 \ifdim\wd2>\dimen6 \dimen6=\wd2 \fi
    \def\colw#1#2{\settowidth{\dimen2}{#2}\edef#1{\the\dimexpr\dimen2+\dimen0\relax}}
    \dimen0=\dimexpr(\dimen6-\wd0)/2\relax
    \colw\durwA{VideoLLaMA3-2B}\colw\durwB{$+$ MOSS (ours)}
    \dimen0=\dimexpr(\dimen6-\wd2)/4\relax
    \colw\fpswA{\textbf{55.4}}\colw\fpswB{\textbf{54.2}}\colw\fpswC{\textbf{53.3}}\colw\fpswD{\textbf{51.5}}
    \begin{subtable}[t]{\linewidth}
        \centering
        \subcaption{Video durations}
        \vspace{-1mm}
        \scalebox{0.85}{%
        \begin{tabular}{l|lc|C{\durwA}C{\durwB}}\durbody\end{tabular}%
        }
        \label{tab:supp_long_video_duration}
    \end{subtable}

    \vspace{2mm}

    \begin{subtable}[t]{\linewidth}
        \centering
        \subcaption{Input frame rates}
        \vspace{-1mm}
        \scalebox{0.85}{%
        \begin{tabular}{l|l|C{\fpswA}C{\fpswB}C{\fpswC}C{\fpswD}}\fpsbody\end{tabular}%
        }
        \label{tab:supp_long_video_fps}
    \end{subtable}
    \label{tab:supp_long_video}
\end{table}

\noindent \textbf{Longer Videos.}
As shown in Tab.~\ref{tab:supp_long_video_duration}, the gains are consistent across all duration bands on both benchmarks, including videos longer than 30 seconds.
This observation is further supported by our temporal action detection results on THUMOS-14 (Sec.~\ref{sec:experiments_tad}), whose videos average 4.4 minutes; adding MOSS improves AdaTAD with both VideoMAE-B and -L backbones (Tab.~\ref{tab:tad}).
These results indicate that improving atomic motion representations can benefit long-range temporal modeling.

\noindent \textbf{Sparse Frame Input.}
We further evaluate whether MOSS remains effective under sparse temporal sampling by varying the input frame rate from 4 to 0.5 FPS (Tab.~\ref{tab:supp_long_video_fps}).
MOSS consistently improves the baseline across all sampling rates.
As expected, the gain gradually decreases as the sampling becomes sparser, since larger inter-frame displacements make reliable correspondence estimation more difficult.
Nevertheless, the consistent gains at 1 FPS and 0.5 FPS suggest that MOSS remains beneficial when local motion representations are hierarchically integrated over long videos.

\subsection{MOSS with Single-Frame VLA Backbones}
\label{sec:supp_vla_single_frame}
Many existing VLA policies, \eg, GR00T-N1.5~\cite{bjorck2025gr00t}, mainly use the current single-frame observation, whereas MOSS computes STSS over multi-frame features.
We here investigate whether MOSS can also be retrofitted into such single-frame VLA backbones.
To this end, we evaluate two variants of the GR00T-based single-frame policy with MOSS on RoboCasa-Kitchen.
The first variant (\textit{Motion-Augmented Current}) inserts MOSS into the intermediate layers of the vision encoder while keeping the original GR00T VLM and diffusion transformer (DiT) action head unchanged.
The resulting motion-augmented current-frame feature is then fed into the VLM.
The second variant (\textit{Motion-to-DiT}) also inserts MOSS into the intermediate layers of the vision encoder, but feeds the MOSS output features directly to DiT through cross-attention.
To keep the computational overhead low, the multi-frame MOSS features are spatio-temporally average-pooled into 64 tokens ($T \times H \times W = 4 \times 4 \times 4$), which are concatenated with the VLM tokens as conditioning tokens.

\begin{table}[t]
    \centering
    \caption{\textbf{Integrating MOSS into single-frame and multi-frame VLA backbones}, evaluated on RoboCasa-Kitchen. We report the success rates (\%).}
    \vspace{1mm}
    \setlength\tabcolsep{4pt}
    \scalebox{0.85}{%
    \begin{tabular}{l|l|c}
    \Xhline{0.8pt}
    VLA backbone & method & RoboCasa-Kitchen \\
    \Xhline{0.3pt}
    \multirow{3}{*}{single-frame} & GR00T-N1.5~\cite{bjorck2025gr00t} & 62.4 \\
     & \cellcolor{gray!15}$+$ Motion-Augmented Current & \cellcolor{gray!15}65.4 \\
     & \cellcolor{gray!15}$+$ Motion-to-DiT & \cellcolor{gray!15}64.8 \\
    \Xhline{0.3pt}
    \multirow{2}{*}{multi-frame} & ContextVLA~\cite{jang2025contextvla} & 64.0 \\
     & \cellcolor{gray!15}$+$ MOSS (ours) & \cellcolor{gray!15}\textbf{66.9} \\
    \Xhline{0.8pt}
    \end{tabular}%
    }
    \label{tab:supp_vla_single_frame}
\end{table}

The results are summarized in Tab.~\ref{tab:supp_vla_single_frame}.
Despite their naive implementations, both variants achieve clear gains over GR00T-N1.5 on RoboCasa-Kitchen.
Notably, these gains are larger than that achieved by ContextVLA alone, highlighting the effectiveness of low-level motion modeling in atomic manipulation tasks.
Motion-to-DiT performs slightly worse than Motion-Augmented Current, which we attribute to the naive spatio-temporal pooling used to compress MOSS features; more effective token compression strategies may further improve its performance.
Finally, both variants underperform ContextVLA$+$MOSS, indicating that MOSS's low-level motion modeling and ContextVLA's VLM-level temporal modeling are complementary and most effective together.

\subsection{Efficiency Comparison}
\label{sec:supp_efficiency}
We measure the wall-clock inference latency and throughput of MOSS in all three settings, \ie, action recognition, motion-centric VQA, and robotic tasks. The results are summarized in Tab.~\ref{tab:supp_latency}.
Across all three architectures, MOSS adds only 1.5--3.3\% latency with a correspondingly small throughput reduction, while delivering substantial accuracy gains.
For real-world robotic deployment, this overhead proved negligible in practice; each inference predicts an action chunk spanning 0.8\,s while being invoked every 0.4\,s.
Moreover, our deployed RTX 5090 workstation was 20--30\% faster than the H200 used for benchmarking, and MOSS ran without any practical latency issues during closed-loop manipulation.

\begin{table}[t]
    \centering
    \caption{\textbf{Inference latency and throughput of MOSS.} (a) and (b) are measured on a single RTX A6000 with 16 frames at 224$\times$224, with (b) including 16-token generation; (c) is measured on a single H200 with 8 frames at a single 224$\times$224 view. VideoLLaMA3-2B in (b) denotes the model fine-tuned on FAVOR-Train (Tab.~\ref{tab:videollm}). $^\dagger$ accuracies obtained with 32 frames.}
    \setlength\tabcolsep{4pt}
    \def\arbody{%
        \Xhline{0.8pt}
        method & latency (ms) & throughput (Hz) & SSV1 & SSV2 & Diving48$^\dagger$ & FineGym99$^\dagger$ \\
        \Xhline{0.3pt}
        Side4Video-B~\cite{side4video} & 39.2 & 25.5 & 60.7 & 71.5 & 88.6 & 92.3 \\
        \rowcolor{gray!15} $+$ MOSS (ours) & 40.0 (+2.1\%) & 25.0 & \textbf{61.8} & \textbf{72.4} & \textbf{91.2} & \textbf{93.9} \\
        \Xhline{0.8pt}}
    \def\vqabody{%
        \Xhline{0.8pt}
        method & latency (ms) & throughput (Hz) & MotionBench & FAVOR-Bench \\
        \Xhline{0.3pt}
        VideoLLaMA3-2B~\cite{videollama3} & 265.7 & 3.8 & 51.4 & 45.5 \\
        \rowcolor{gray!15} $+$ MOSS (ours) & 269.8 (+1.5\%) & 3.7 & \textbf{54.2} & \textbf{46.6} \\
        \Xhline{0.8pt}}
    \def\robotbody{%
        \Xhline{0.8pt}
        \multirow{2}{*}{method} & \multirow{2}{*}{latency (ms)} & \multirow{2}{*}{throughput (Hz)} & \multirow{2}{*}{\makecell{RoboCasa\\Kitchen}} & \multicolumn{2}{c}{MoveSense} & \multirow{2}{*}{\makecell{Pong\\Predict}} & \multirow{2}{*}{\makecell{Conveyor\\PnP}} \\
        \cline{5-6}
         & & & & Left/Right & CW/CCW & & \\
        \Xhline{0.3pt}
        ContextVLA~\cite{jang2025contextvla} & 78.5 & 12.7 & 64.0 & 67.5 & 42.7 & 51.9 & 54.2 \\
        \rowcolor{gray!15} $+$ MOSS (ours) & 81.1 (+3.3\%) & 12.3 & \textbf{66.9} & \textbf{95.8} & \textbf{99.0} & \textbf{81.5} & \textbf{87.5} \\
        \Xhline{0.8pt}}
    \setbox0=\hbox{\begin{tabular}{l|cc|cccc}\arbody\end{tabular}}
    \setbox2=\hbox{\begin{tabular}{l|cc|cc}\vqabody\end{tabular}}
    \setbox4=\hbox{\begin{tabular}{l|cc|ccccc}\robotbody\end{tabular}}
    \dimen6=\wd0 \ifdim\wd2>\dimen6 \dimen6=\wd2 \fi \ifdim\wd4>\dimen6 \dimen6=\wd4 \fi
    \def\colw#1#2{\settowidth{\dimen2}{#2}\edef#1{\the\dimexpr\dimen2+\dimen0\relax}}
    \dimen0=\dimexpr(\dimen6-\wd0)/4\relax
    \colw\arwA{SSV1}\colw\arwB{SSV2}\colw\arwC{Diving48$^\dagger$}\colw\arwD{FineGym99$^\dagger$}
    \dimen0=\dimexpr(\dimen6-\wd2)/2\relax
    \colw\vqawA{MotionBench}\colw\vqawB{FAVOR-Bench}
    \dimen0=\dimexpr(\dimen6-\wd4)/5\relax
    \colw\robotwA{RoboCasa}\colw\robotwB{Left/Right}\colw\robotwC{CW/CCW}\colw\robotwD{Predict}\colw\robotwE{Conveyor}
    \begin{subtable}[t]{\linewidth}
        \centering
        \subcaption{Action recognition}
        \vspace{-1mm}
        \scalebox{0.85}{%
        \begin{tabular}{l|cc|C{\arwA}C{\arwB}C{\arwC}C{\arwD}}\arbody\end{tabular}%
        }
        \label{tab:supp_latency_ar}
    \end{subtable}

    \vspace{2mm}

    \begin{subtable}[t]{\linewidth}
        \centering
        \subcaption{Motion-centric VQA}
        \vspace{-1mm}
        \scalebox{0.85}{%
        \begin{tabular}{l|cc|C{\vqawA}C{\vqawB}}\vqabody\end{tabular}%
        }
        \label{tab:supp_latency_vqa}
    \end{subtable}

    \vspace{2mm}

    \begin{subtable}[t]{\linewidth}
        \centering
        \subcaption{Robotic tasks}
        \vspace{-1mm}
        \scalebox{0.85}{%
        \begin{tabular}{l|cc|C{\robotwA}C{\robotwB}C{\robotwC}C{\robotwD}C{\robotwE}}\robotbody\end{tabular}%
        }
        \label{tab:supp_latency_robot}
    \end{subtable}
    \label{tab:supp_latency}
\end{table}

\section{Per-Class Analysis}
\label{sec:supp_perclass}
\begin{figure*}[t]
    \centering
    \includegraphics[width=0.99\linewidth]{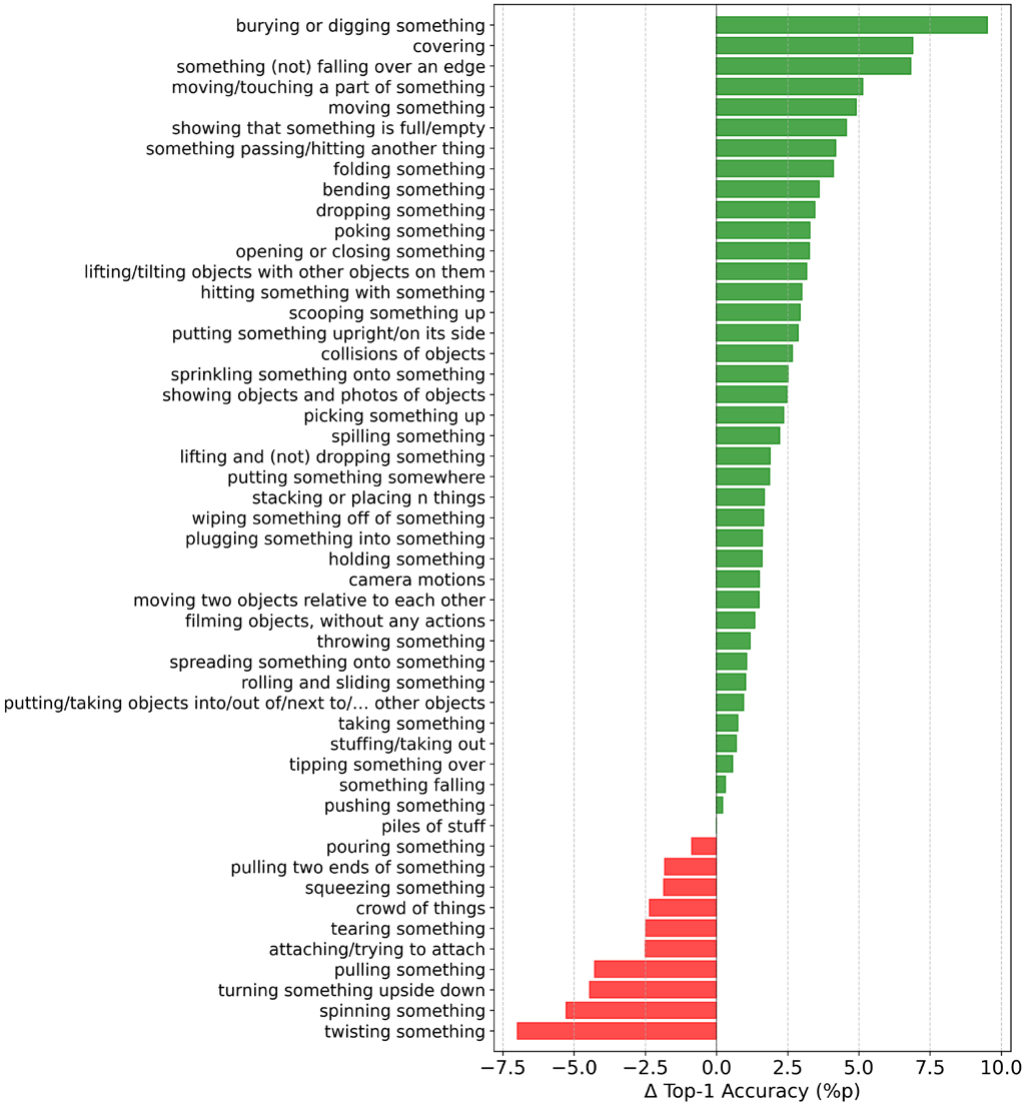}
    \vspace{-2mm}
    \caption{\textbf{Effects of 2nd-order STSS on Something-Something V1.} The figure shows the accuracy difference for each action group when incorporating the 2nd-order STSS on top of the 1st-order.}
    \label{fig:supp_perclass}
\end{figure*}

We here provide a statistical analysis across diverse action classes, offering a comprehensive understanding of when 2nd-order STSS becomes effective.
To this end, we measure the difference in accuracy of each \textit{action group} in Something-Something V1~\cite{goyal2017something} when incorporating the 2nd-order STSS on top of the 1st-order STSS.
The results are summarized in Fig.~\ref{fig:supp_perclass}.
Among the 50 action groups, we observe accuracy improvements in 39 groups and drops in 10 groups.
Specifically, we find that the 2nd-order STSS is beneficial in understanding not only basic motions, \eg, moving something or moving/touching a part of something (Fig.~\ref{fig:stss_sup1}), but also more complex object-object interactions, \eg, passing/hitting another object or moving two objects relative to each other (Fig.~\ref{fig:stss_sup2}), (dis-)appearance events, \eg, burying, covering, or dropping (Fig.~\ref{fig:stss_sup3_a}), and camera motion scenarios (Fig.~\ref{fig:stss_sup3_b}).
These improvements indicate that the 2nd-order STSS captures complementary temporal dynamics beyond what the 1st-order STSS can capture.
Meanwhile, the accuracy drops mainly in action groups such as squeezing, spinning, or twisting.
This implies that when motion blur or severe deformation makes 1st-order STSS unreliable, motion segmentation in 2nd-order STSS becomes ambiguous, resulting in limited benefits.
This analysis reveals when 2nd-order STSS provides tangible benefits and when it becomes less effective, providing practical guidance for its use.

\section{Additional Visualization Results}
\label{sec:supp_visualization}
We present additional visualization results on Something-Something V1 in Figs.~\ref{fig:stss_sup1}-\ref{fig:stss_sup5}.
These figures visualize STSS maps and their feature L2 norms across the 1st to 3rd layers for videos including: simple motions (Fig.~\ref{fig:stss_sup1}), object-object interactions (Fig.~\ref{fig:stss_sup2}), sudden object appearance (Fig.~\ref{fig:stss_sup3_a}), camera motions (Fig.~\ref{fig:stss_sup3_b}), motion changes (Fig.~\ref{fig:stss_sup4}), and background clutter (Fig.~\ref{fig:stss_sup5}).

\begin{figure*}[t]
    \centering
\begin{subfigure}{\linewidth}
\centering
    \includegraphics[width=0.95\linewidth]{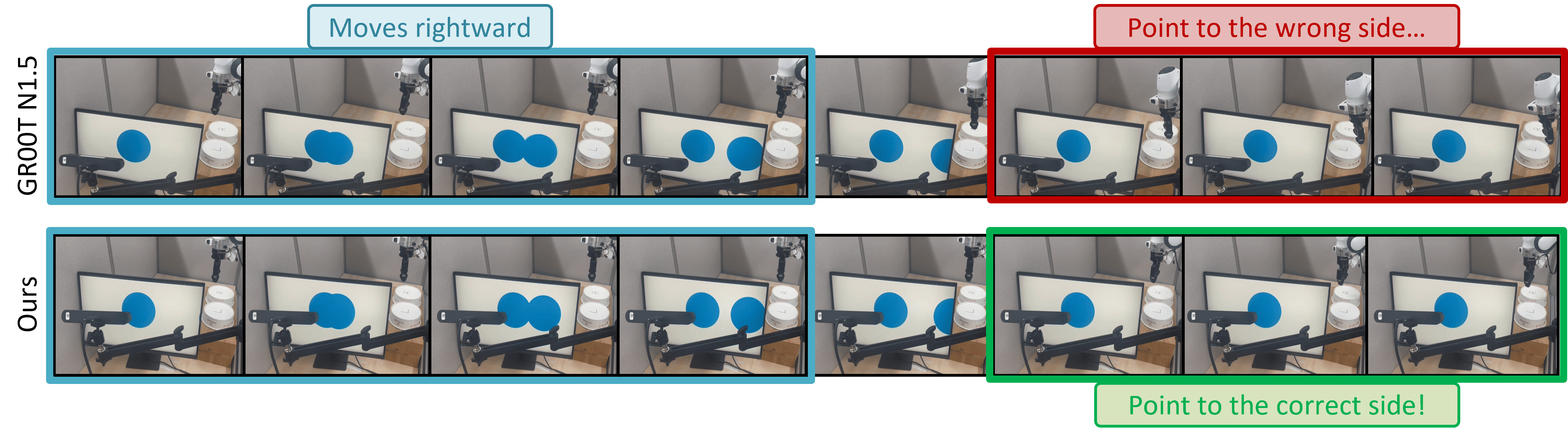}
    \vspace{-2mm}
    \caption{MoveSense (Left/Right)}
    \label{fig:robot_sup1_a}
\end{subfigure}
    \hspace{0.01\linewidth}

\vspace{2mm}

\begin{subfigure}{\linewidth}
\centering
    \includegraphics[width=0.95\linewidth]{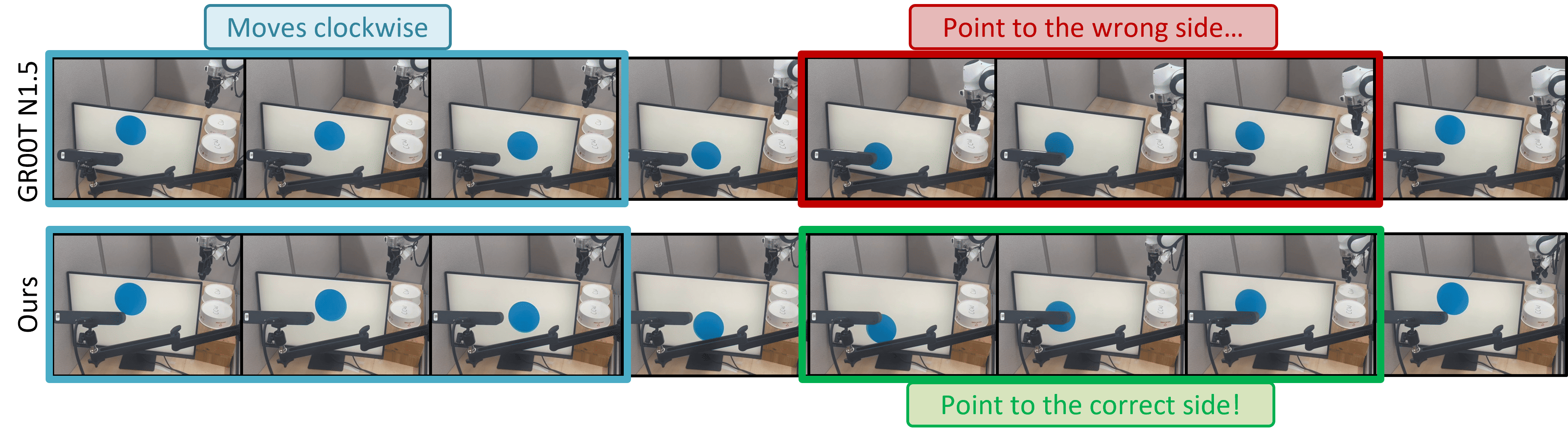}
        \vspace{-2mm}
        \caption{MoveSense (CW/CCW)}
    \label{fig:robot_sup1_b}
\end{subfigure}

\vspace{2mm}

\begin{subfigure}{\linewidth}
\centering
    \includegraphics[width=0.95\linewidth]{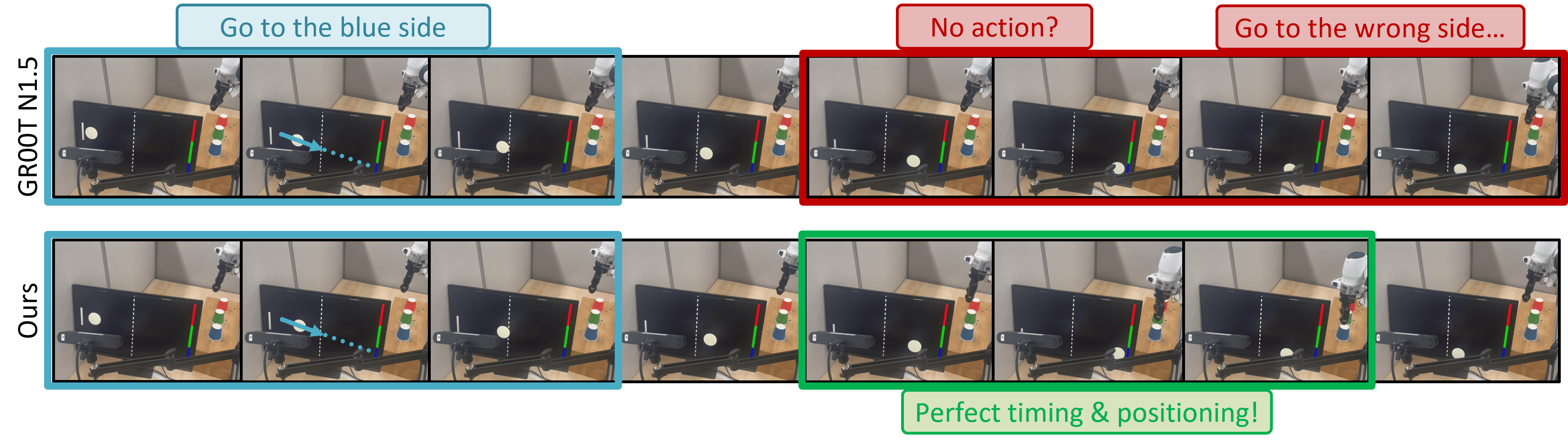}
        \vspace{-2mm}
        \caption{PongPredict}
    \label{fig:robot_sup1_c}
\end{subfigure}

\vspace{2mm}

\begin{subfigure}{\linewidth}
\centering
    \includegraphics[width=0.95\linewidth]{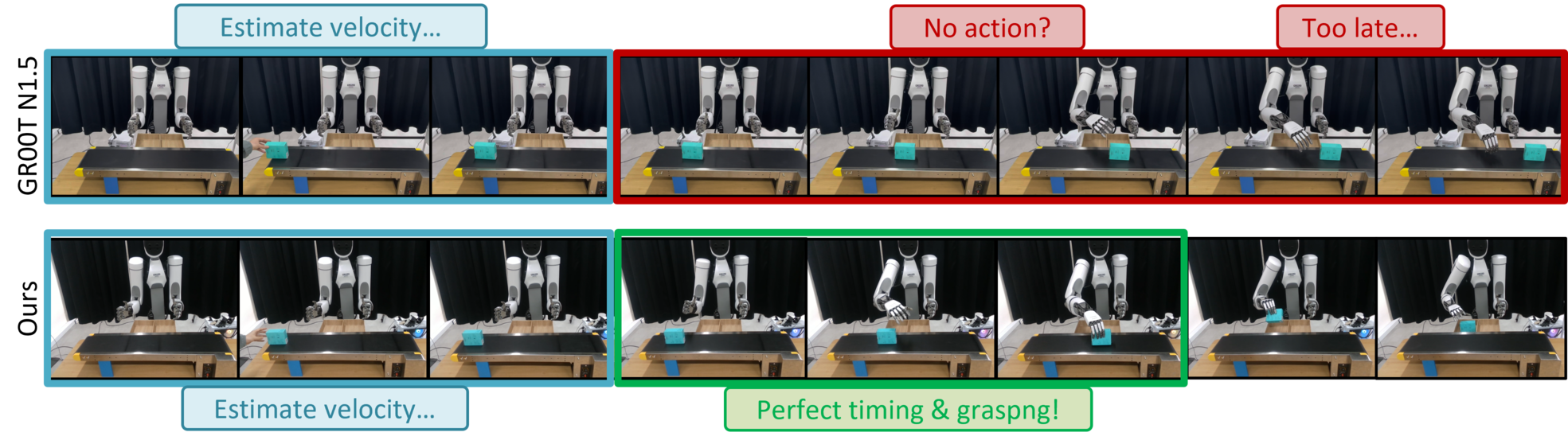}
        \vspace{-2mm}
        \caption{Conveyor PnP}
    \label{fig:robot_sup1_d}
\end{subfigure}
    \caption{\textbf{Example rollouts of real-world robot tasks}. MoveSense, PongPredict, and Conveyor PnP examples. Best viewed in PDF.}
    \label{fig:robot_sup1}
    \vspace{-1mm}
\end{figure*}

In Fig.~\ref{fig:robot_sup1}, we present qualitative results on real-world robotic tasks.
While GR00T-N1.5 often fails to distinguish basic motion directions (left vs. right) and circular movements (CW vs. CCW), our model makes correct decisions on the same examples.
For the motion prediction task (PongPredict), GR00T-N1.5 often reacts too late or becomes confused, whereas the MOSS model demonstrates strong performance.
On the Conveyor PnP task, which requires the model to precisely track and manipulate moving objects, GR00T-N1.5 either remains stationary or mistimes its grasp, failing to interact with the moving objects, whereas the MOSS model successfully grasps objects at the correct moment.
These results demonstrate that MOSS enables the VLA model to better understand dynamic environments and effectively interact with moving objects in real-world robotic tasks.
Full demonstration videos are provided in \texttt{Robotics\_demo\_videos.pptx} in the supplementary zip.

\section{Potential Broader Impact}
\label{sec:broader_impact}
Our work advances the modeling of complex temporal dynamics in video representation learning, providing a foundational technology that assists various application scenarios such as automatic surveillance, robotics, autonomous driving, and sports analytics. 
However, video-related research could potentially be misused for harmful purposes, particularly in unauthorized monitoring.
We believe that the positive impacts of our work substantially exceed the negative impacts, but we clarify that the ultimate outcome depends on how end users choose to implement this technology.

\section{Limitation and Future Work}
\label{sec:limitation}
\begin{figure*}[t]
    \centering
\begin{subfigure}{\linewidth}
\centering
    \includegraphics[width=0.8\linewidth]{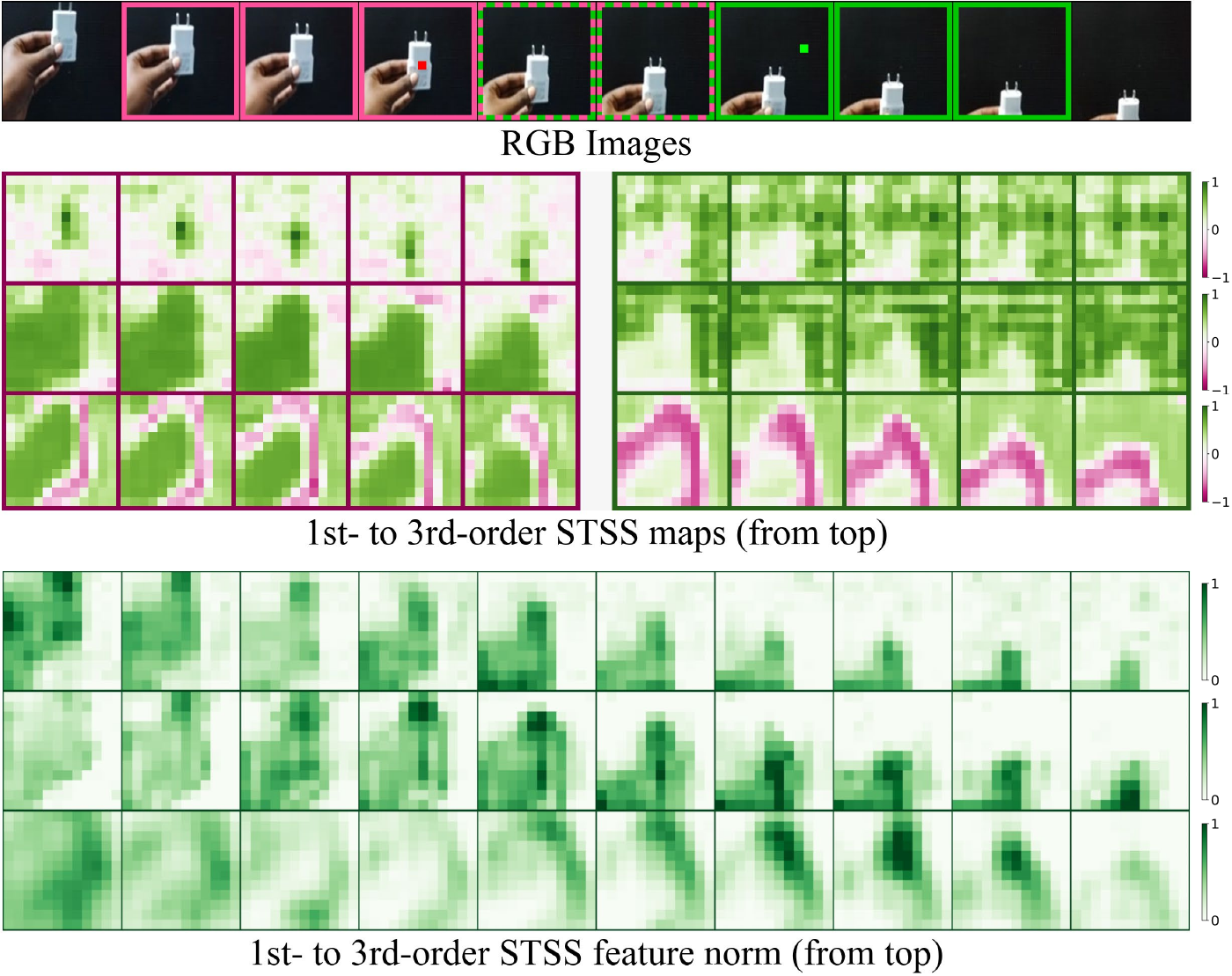}
    \caption{``Moving something down"}
    \label{fig:stss_sup1_a}
\end{subfigure}
    \hspace{0.01\linewidth}

\vspace{3mm}

\begin{subfigure}{\linewidth}
\centering
    \includegraphics[width=0.8\linewidth]{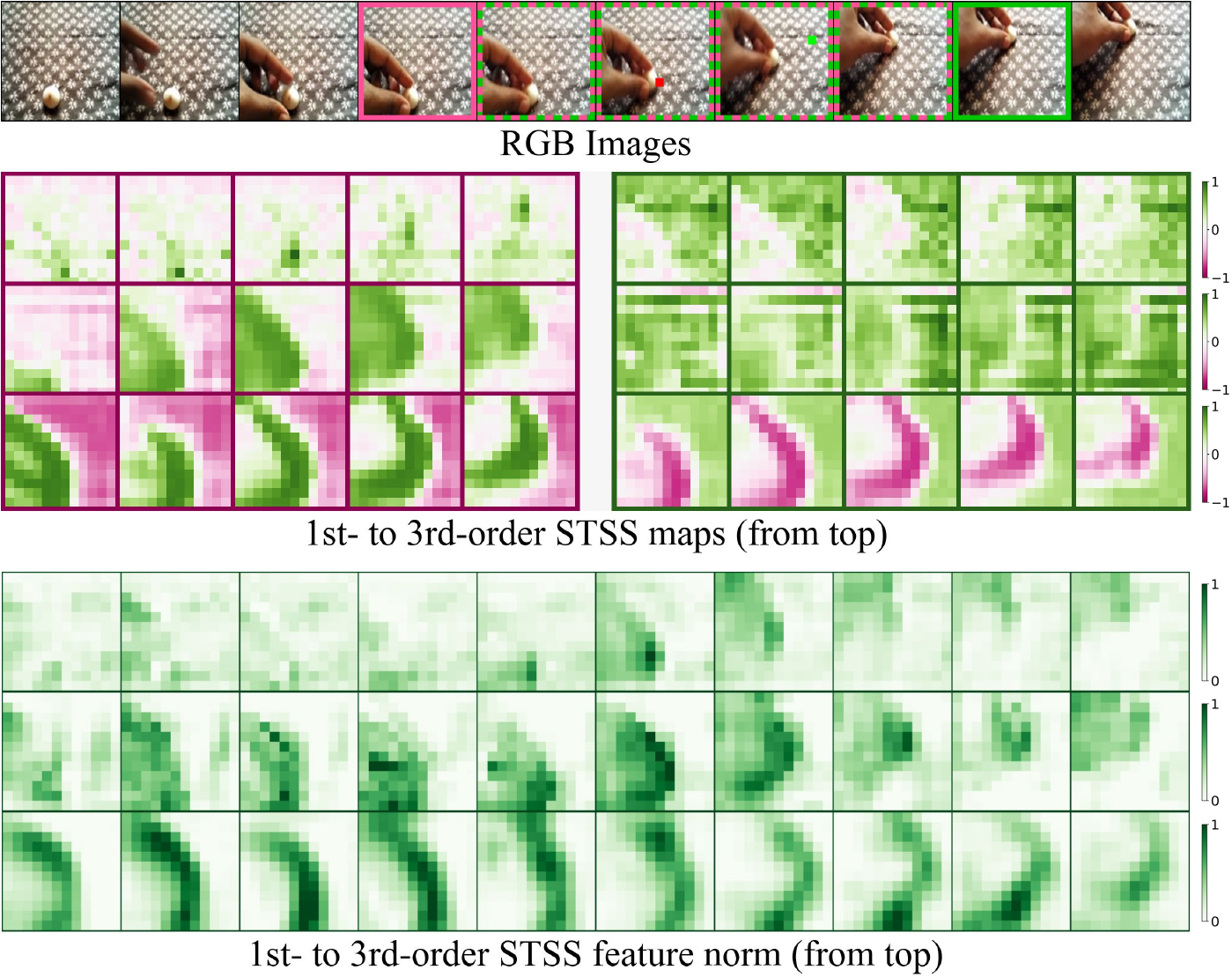}
        \caption{Moving something up}
    \label{fig:stss_sup1_b}
\end{subfigure}
    \caption{\textbf{STSS visualization}. RGB frames at the top show query locations and their spatio-temporal matching regions marked in \red{red} and \textcolor{Green}{green}, respectively. The subsequent rows show STSS maps for the two queries and the L2-norm of feature maps from 1st- to 3rd-order STSSs. Best viewed in PDF.}
    \label{fig:stss_sup1}
\end{figure*}

\begin{figure*}[t]
    \centering
\begin{subfigure}{\linewidth}
\centering
    \includegraphics[width=0.8\linewidth]{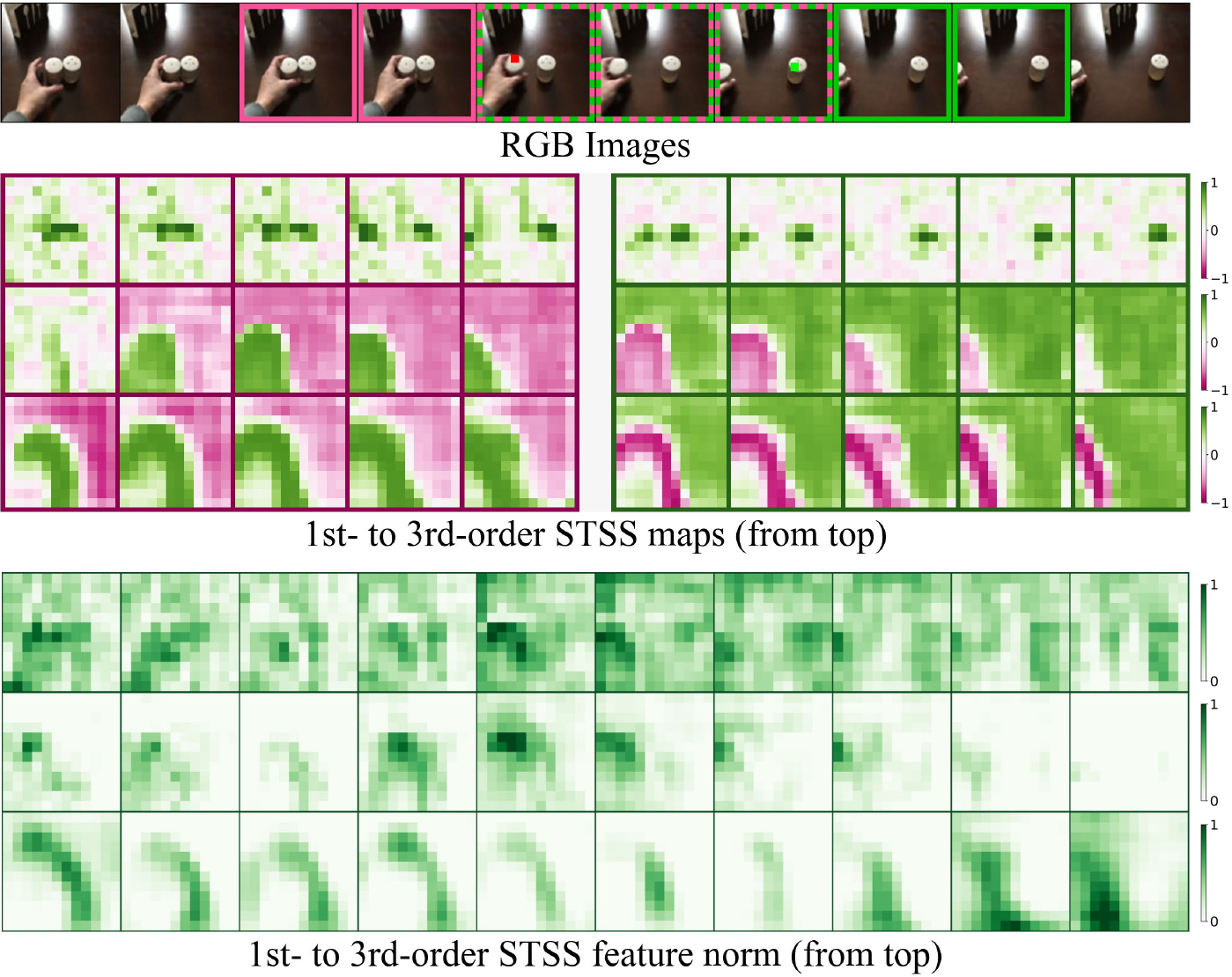}
    \caption{`Moving something away from something"}
    \label{fig:stss_sup2_a}
\end{subfigure}
    \hspace{0.01\linewidth}

\vspace{3mm}

\begin{subfigure}{\linewidth}
\centering
    \includegraphics[width=0.8\linewidth]{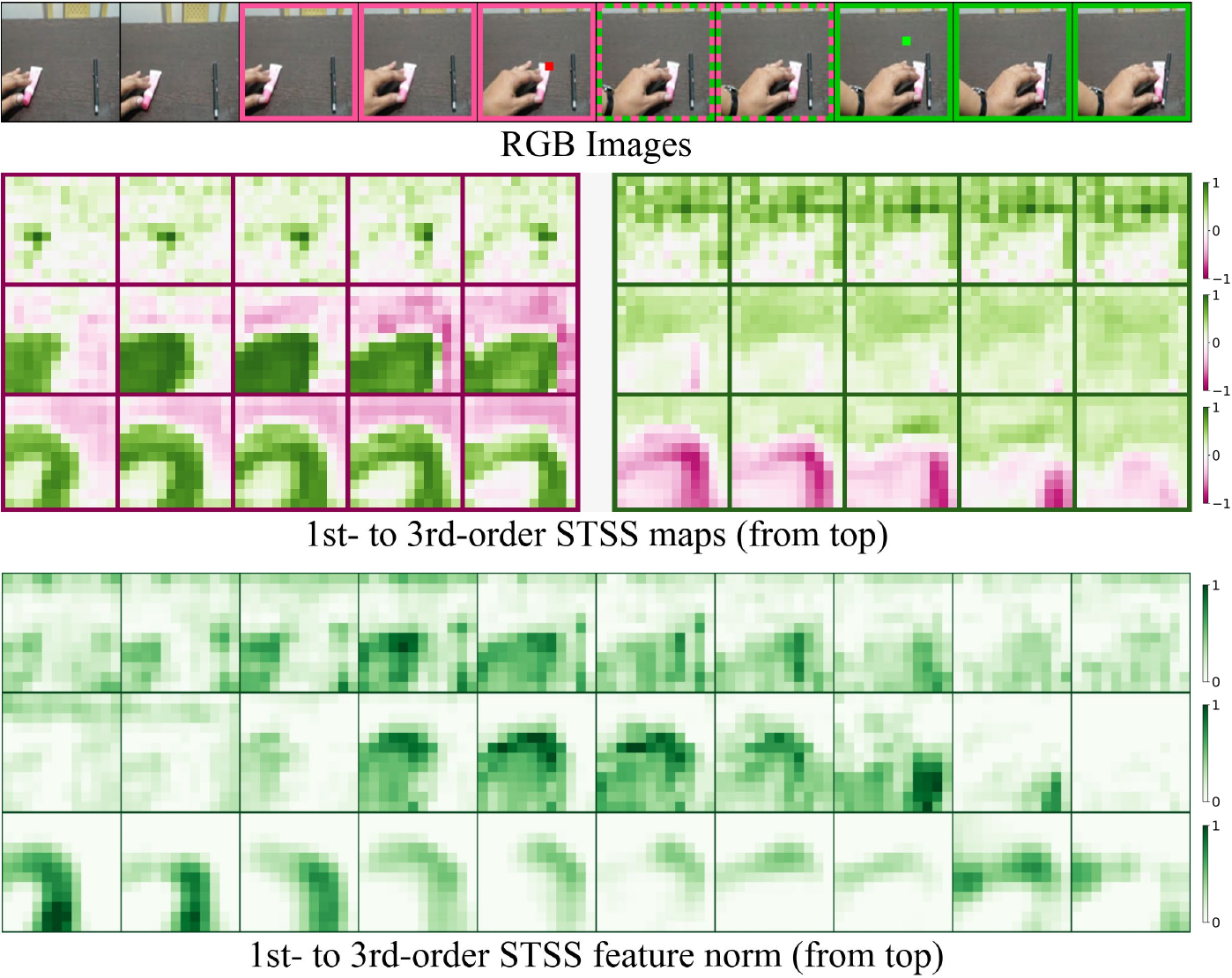}
    \caption{``Moving something closer to something"}
    \label{fig:stss_sup2_b}
\end{subfigure}
    \caption{\textbf{STSS visualization}. RGB frames at the top show query locations and their spatio-temporal matching regions marked in \red{red} and \textcolor{Green}{green}, respectively. The subsequent rows show STSS maps for the two queries and the L2-norm of feature maps from 1st- to 3rd-order STSSs. Best viewed in PDF.}
    \label{fig:stss_sup2}
\end{figure*}

\begin{figure*}[t]
    \centering
\begin{subfigure}{\linewidth}
\centering
    \includegraphics[width=0.8\linewidth]{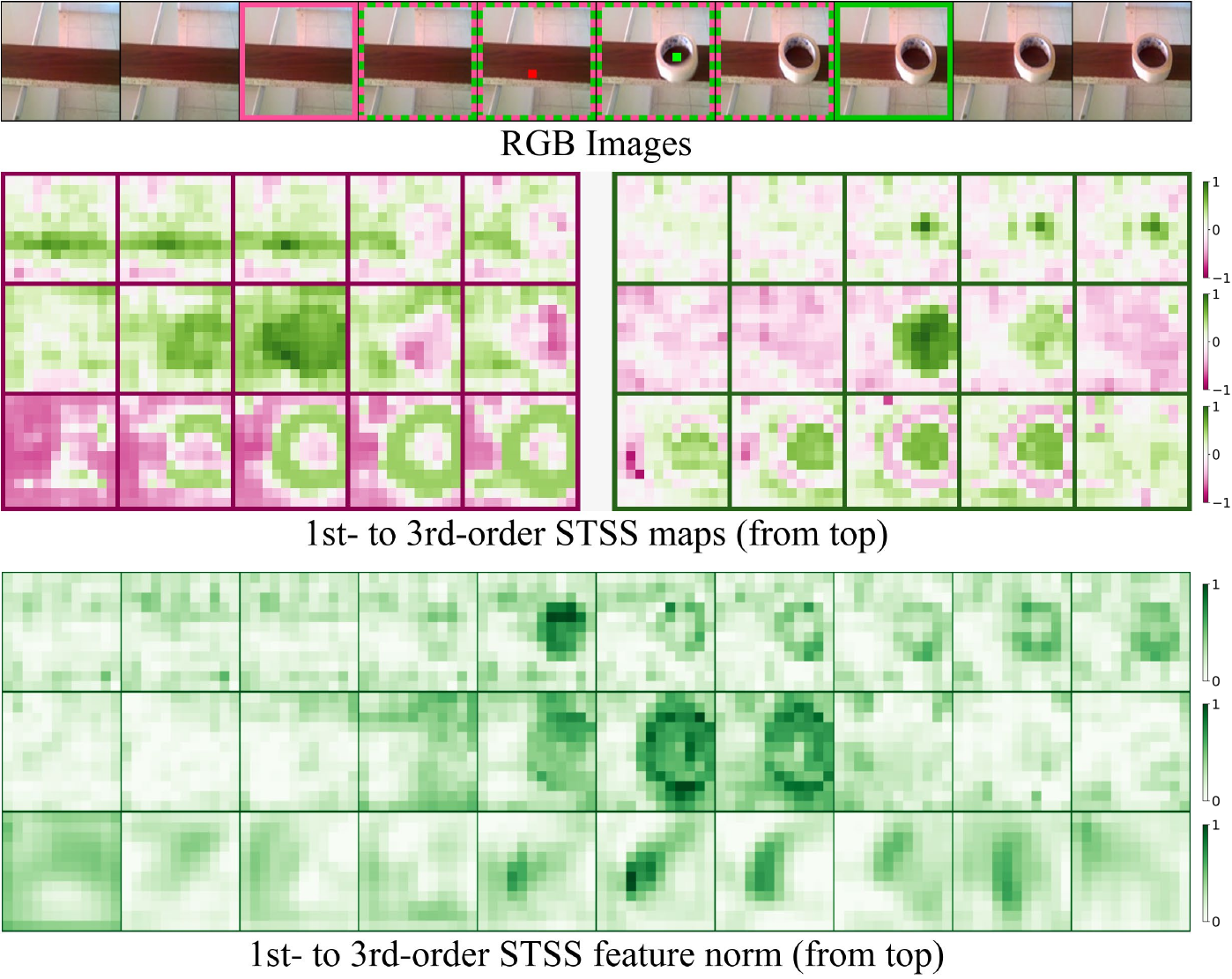}
    \caption{`Dropping something"}
    \label{fig:stss_sup3_a}
\end{subfigure}
    \hspace{0.01\linewidth}

\vspace{3mm}

\begin{subfigure}{\linewidth}
\centering
    \includegraphics[width=0.8\linewidth]{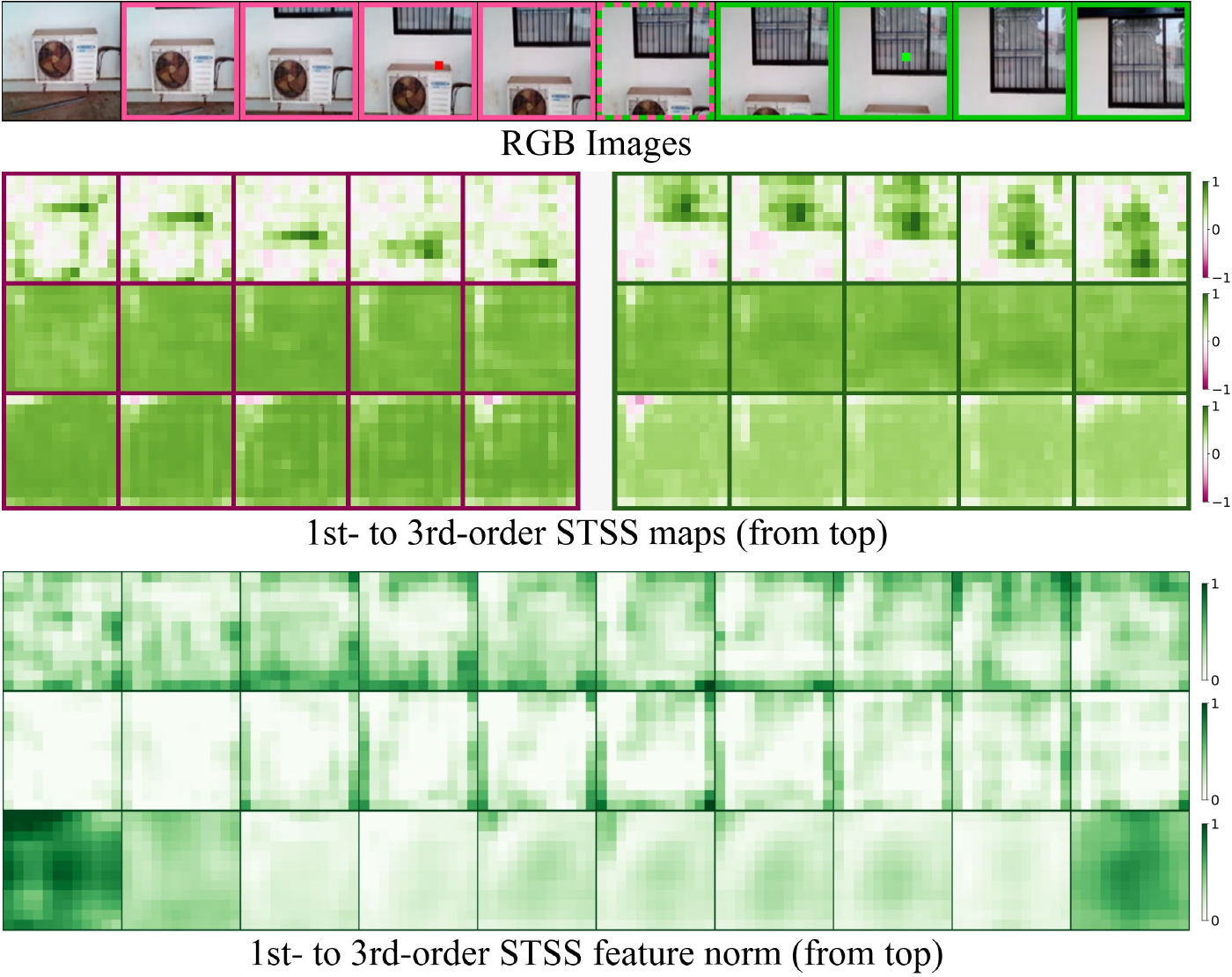}
    \caption{``Turning the camera downwards while filming something"}
    \label{fig:stss_sup3_b}
\end{subfigure}
    \caption{\textbf{STSS visualization}. RGB frames at the top show query locations and their spatio-temporal matching regions marked in \red{red} and \textcolor{Green}{green}, respectively. The subsequent rows show STSS maps for the two queries and the L2-norm of feature maps from 1st- to 3rd-order STSSs. Best viewed in PDF.}
    \label{fig:stss_sup3}
\end{figure*}

\begin{figure*}[t]
    \centering
\begin{subfigure}{\linewidth}
\centering
    \includegraphics[width=0.8\linewidth]{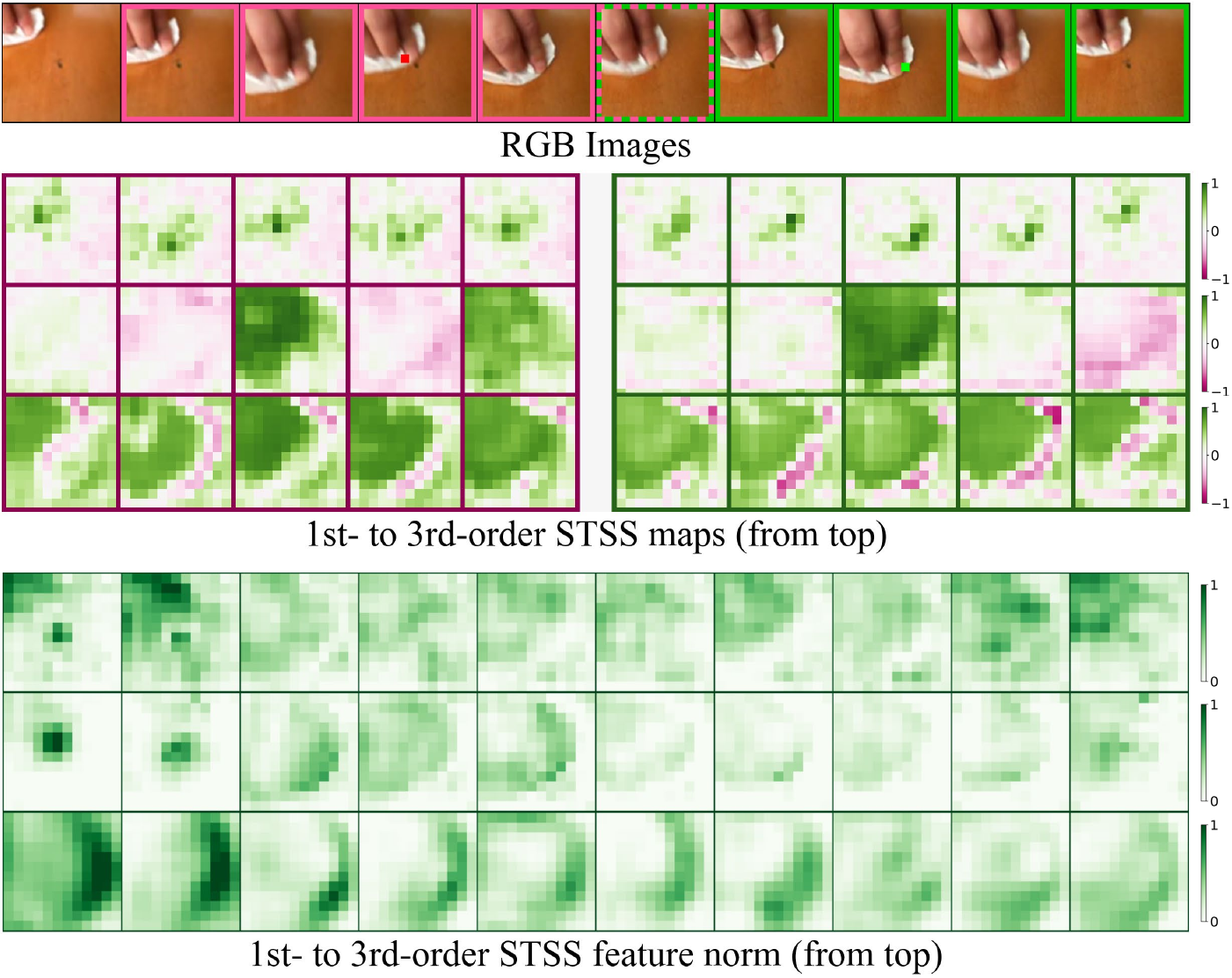}
    \caption{`Pretending or failing to wipe something off of something"}
    \label{fig:stss_sup4_a}
\end{subfigure}
    \hspace{0.01\linewidth}

\vspace{3mm}

\begin{subfigure}{\linewidth}
\centering
    \includegraphics[width=0.8\linewidth]{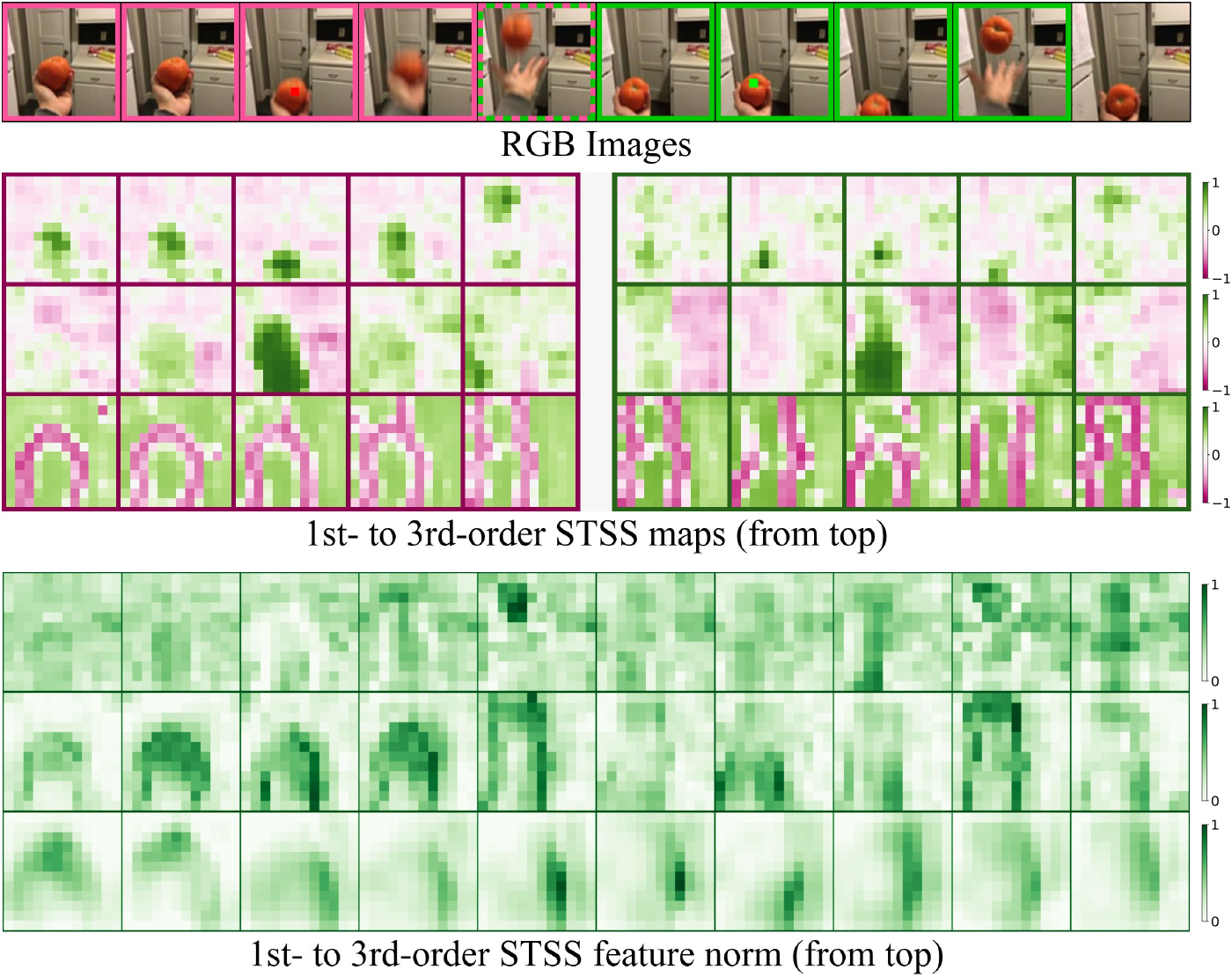}
    \caption{``Throwing something in the air and catching it"}
    \label{fig:stss_sup4_b}
\end{subfigure}
    \caption{\textbf{STSS visualization}. RGB frames at the top show query locations and their spatio-temporal matching regions marked in \red{red} and \textcolor{Green}{green}, respectively. The subsequent rows show STSS maps for the two queries and the L2-norm of feature maps from 1st- to 3rd-order STSSs. Best viewed in PDF.}
    \label{fig:stss_sup4}
\end{figure*}

\begin{figure*}[t]
    \centering
\begin{subfigure}{\linewidth}
\centering
    \includegraphics[width=0.8\linewidth]{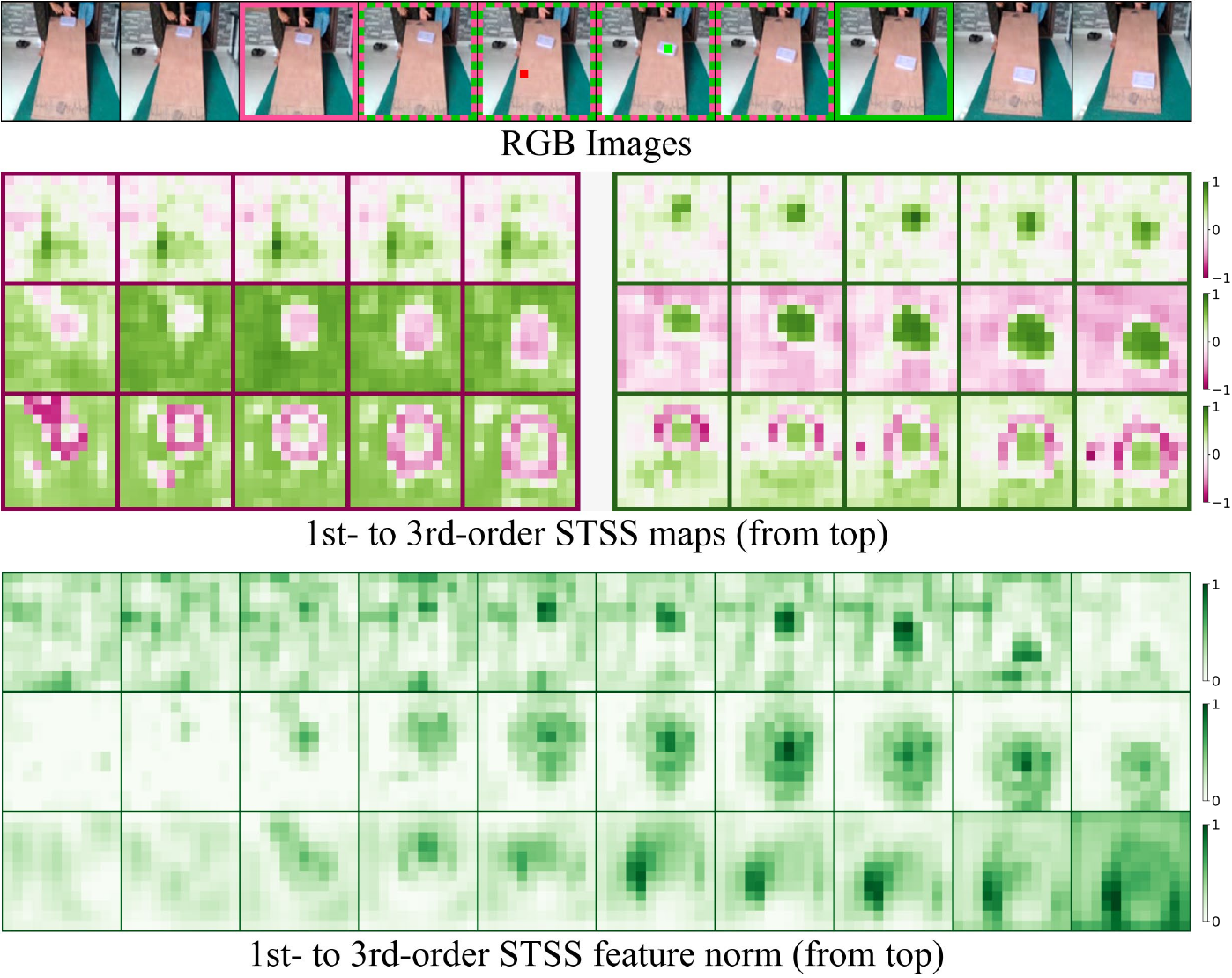}
    \caption{`Putting something that can't roll onto a slanted surface so it slides down"}
    \label{fig:stss_sup5_a}
\end{subfigure}
    \hspace{0.01\linewidth}

\vspace{3mm}

\begin{subfigure}{\linewidth}
\centering
    \includegraphics[width=0.8\linewidth]{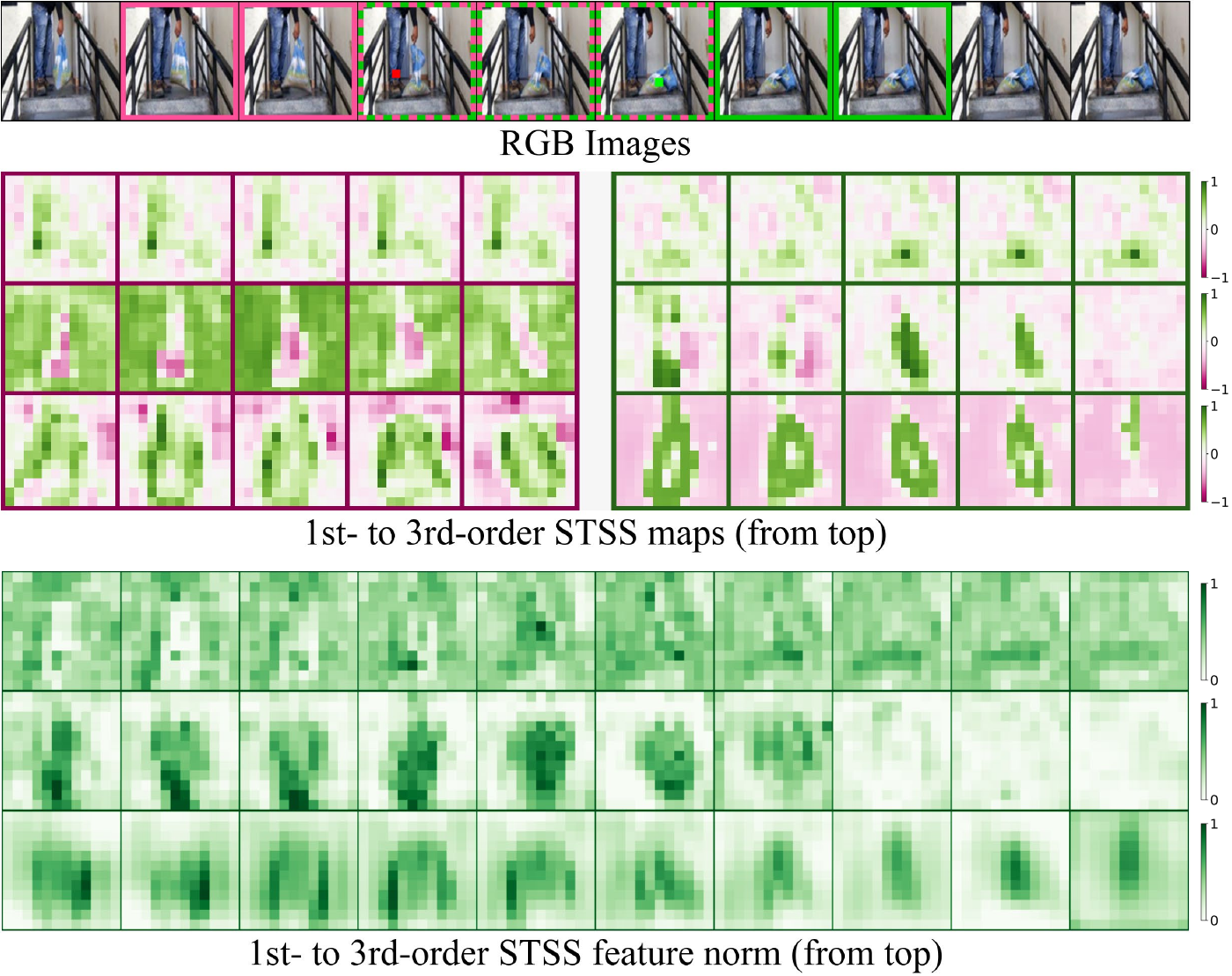}
    \caption{``Lifting something up completely then letting it drop down"}
    \label{fig:stss_sup5_b}
\end{subfigure}
    \caption{\textbf{STSS visualization}. RGB frames at the top show query locations and their spatio-temporal matching regions marked in \red{red} and \textcolor{Green}{green}, respectively. The subsequent rows show STSS maps for the two queries and the L2-norm of feature maps from 1st- to 3rd-order STSSs. Best viewed in PDF.}
    \label{fig:stss_sup5}
\end{figure*}

Our research explores the role of high-order STSS in learning video representations. While our theoretical analysis and the toy examples demonstrate that 3rd-order STSS has the potential to capture group-wise motion patterns, we observe that the learned model primarily utilizes 3rd-order STSS to capture motion boundaries for video action recognition (Figs.~\ref{fig:stss}, \ref{fig:stss_sup1}-\ref{fig:stss_sup5}). This limited utilization of 3rd-order STSS may explain why 2nd- and 3rd-order STSS are not complementary to each other since 2nd-order STSS can already provide such boundary information implicitly. We conjecture that learning group-wise motion patterns may not provide significant benefits for existing action recognition tasks. Future work should focus on developing new benchmarks where higher-order (3rd-order or beyond) temporal dynamics can demonstrate more meaningful benefits.

\begin{table}
\centering
\caption{\textbf{Full results on RoboCasa-Kitchen.} We report the success rates (\%) of VLAs fine-tuned on the training dataset using official implementations.}
  \begin{tabular}{lccc}
  \toprule
  Task & GR00T-N1.5 & ContextVLA& \textbf{ContextVLA $+$ MOSS (ours)} \\
  \midrule
  \multicolumn{4}{l}{\textit{Pick and place (8 tasks)}} \\
  \midrule
  Cabinet to Counter  & \phantom{0}58.0& \phantom{0}58.0 & \phantom{0}44.0 \\
  Counter to Cabinet  &\phantom{0}56.0 & \phantom{0}46.0& \phantom{0}54.0 \\
  Counter to Microwave & \phantom{0}30.0& \phantom{0}34.0& \phantom{0}40.0 \\
  Counter to Sink     & \phantom{0}52.0& \phantom{0}56.0& \phantom{0}56.0 \\
  Counter to Stove    & \phantom{0}60.0& \phantom{0}76.0& \phantom{0}68.0 \\
  Microwave to Counter& \phantom{0}44.0& \phantom{0}36.0& \phantom{0}40.0 \\
  Sink to Counter     & \phantom{0}54.0& \phantom{0}56.0& \phantom{0}68.0 \\
  Stove to Counter    & \phantom{0}62.0& \phantom{0}58.0& \phantom{0}58.0 \\
  \midrule
  \multicolumn{4}{l}{\textit{Open or close (6 tasks)}} \\
  \midrule
  Close Double Door & \phantom{0}76.0& \phantom{0}84.0& \phantom{0}72.0  \\
  Close Drawer      & \phantom{0}92.0& \phantom{0}96.0& \phantom{0}96.0  \\
  Close Single Door & \phantom{0}98.0& \phantom{0}94.0& \phantom{0}92.0 \\
  Open Double Door  & \phantom{0}96.0& \phantom{0}82.0& \phantom{0}86.0  \\
  Open Drawer       & \phantom{0}64.0& \phantom{0}66.0& \phantom{0}76.0  \\
  Open Single Door  & \phantom{0}56.0& \phantom{0}72.0& \phantom{0}82.0  \\
  \midrule
  \multicolumn{4}{l}{\textit{Others (10 tasks)}} \\
  \midrule
  Coffee Press Button  & \phantom{0}82.0& \phantom{0}86.0& \phantom{0}92.0  \\
  Coffee Serve Mug     & \phantom{0}60.0& \phantom{0}74.0& \phantom{0}80.0  \\
  Coffee Setup Mug     & \phantom{0}22.0& \phantom{0}26.0& \phantom{0}30.0  \\
  Turn Off Microwave   & \phantom{0}92.0&\phantom{0}98.0 & \phantom{0}92.0  \\
  Turn Off Sink Faucet & \phantom{0}86.0& \phantom{0}76.0& \phantom{0}94.0  \\
  Turn Off Stove       & \phantom{0}22.0 & \phantom{0}14.0& \phantom{0}20.0  \\
  Turn On Microwave    & \phantom{0}40.0& \phantom{0}50.0& \phantom{0}94.0  \\
  Turn On Sink Faucet  & \phantom{0}88.0& \phantom{0}82.0& \phantom{0}78.0 \\
  Turn On Stove        & \phantom{0}52.0&\phantom{0}44.0 & \phantom{0}38.0  \\
  Turn Sink Spout      & \phantom{0}56.0 & \phantom{0}62.0 & \phantom{0}66.0  \\
  \midrule
  \textbf{Total (24 tasks)} & \phantom{0}62.4&\phantom{0}64.0 & \phantom{0}\textbf{66.9} \\
  \bottomrule
  \end{tabular}
    \label{tab:robocasa}
\end{table}

\clearpage

\newpage
\section*{NeurIPS Paper Checklist}

The checklist is designed to encourage best practices for responsible machine learning research, addressing issues of reproducibility, transparency, research ethics, and societal impact. Do not remove the checklist: {\bf The papers not including the checklist will be desk rejected.} The checklist should follow the references and follow the (optional) supplemental material.  The checklist does NOT count towards the page
limit. 

Please read the checklist guidelines carefully for information on how to answer these questions. For each question in the checklist:
\begin{itemize}
    \item You should answer \answerYes{}, \answerNo{}, or \answerNA{}.
    \item \answerNA{} means either that the question is Not Applicable for that particular paper or the relevant information is Not Available.
    \item Please provide a short (1--2 sentence) justification right after your answer (even for \answerNA). 
\end{itemize}

{\bf The checklist answers are an integral part of your paper submission.} They are visible to the reviewers, area chairs, senior area chairs, and ethics reviewers. You will also be asked to include it (after eventual revisions) with the final version of your paper, and its final version will be published with the paper.

The reviewers of your paper will be asked to use the checklist as one of the factors in their evaluation. While \answerYes{} is generally preferable to \answerNo{}, it is perfectly acceptable to answer \answerNo{} provided a proper justification is given (e.g., error bars are not reported because it would be too computationally expensive'' or ``we were unable to find the license for the dataset we used''). In general, answering \answerNo{} or \answerNA{} is not grounds for rejection. While the questions are phrased in a binary way, we acknowledge that the true answer is often more nuanced, so please just use your best judgment and write a justification to elaborate. All supporting evidence can appear either in the main paper or the supplemental material, provided in appendix. If you answer \answerYes{} to a question, in the justification please point to the section(s) where related material for the question can be found.

IMPORTANT, please:
\begin{itemize}
    \item {\bf Delete this instruction block, but keep the section heading ``NeurIPS Paper Checklist"},
    \item  {\bf Keep the checklist subsection headings, questions/answers and guidelines below.}
    \item {\bf Do not modify the questions and only use the provided macros for your answers}.
\end{itemize}

\begin{enumerate}

\item {\bf Claims}
    \item[] Question: Do the main claims made in the abstract and introduction accurately reflect the paper's contributions and scope?
    \item[] Answer: \answerYes %
    \item[] Justification: The abstract and introduction accurately summarize our key contributions and experimental results.

\item {\bf Limitations}
    \item[] Question: Does the paper discuss the limitations of the work performed by the authors?
    \item[] Answer: \answerYes{} %
    \item[] Justification: We discuss the limitations in Appendix~\ref{sec:limitation}.

\item {\bf Theory assumptions and proofs}
    \item[] Question: For each theoretical result, does the paper provide the full set of assumptions and a complete (and correct) proof?
    \item[] Answer: \answerNA{} %
    \item[] Justification: The paper does not include any theorems or lemmas that require theoretical proofs.

    \item {\bf Experimental result reproducibility}
    \item[] Question: Does the paper fully disclose all the information needed to reproduce the main experimental results of the paper to the extent that it affects the main claims and/or conclusions of the paper (regardless of whether the code and data are provided or not)?
    \item[] Answer: \answerYes{} %
    \item[] Justification: We provide detailed experimental setups including datasets, model configurations, and training hyperparameters in Appendices~\ref{sec:implementation_detail_supp} and \ref{sec:experiments_supp}.

\item {\bf Open access to data and code}
    \item[] Question: Does the paper provide open access to the data and code, with sufficient instructions to faithfully reproduce the main experimental results, as described in supplemental material?
    \item[] Answer: \answerYes{} %
    \item[] Justification: We provide PyTorch implementation code in our supplementary material. The code will be publicly available at \url{http://cvlab.postech.ac.kr/research/MOSS}.

\item {\bf Experimental setting/details}
    \item[] Question: Does the paper specify all the training and test details (e.g., data splits, hyperparameters, how they were chosen, type of optimizer) necessary to understand the results?
    \item[] Answer: \answerYes{} %
    \item[] Justification: We provide detailed experimental setups including datasets, model configurations, and training hyperparameters in Appendices~\ref{sec:implementation_detail_supp} and \ref{sec:experiments_supp}.

\item {\bf Experiment statistical significance}
    \item[] Question: Does the paper report error bars suitably and correctly defined or other appropriate information about the statistical significance of the experiments?
    \item[] Answer: \answerNo{} %
    \item[] Justification: Due to the high computational cost and time required for each experiment, we report results from a single run with a fixed random seed.

\item {\bf Experiments compute resources}
    \item[] Question: For each experiment, does the paper provide sufficient information on the computer resources (type of compute workers, memory, time of execution) needed to reproduce the experiments?
    \item[] Answer: \answerYes{} %
    \item[] Justification: We specify the compute resources (8 NVIDIA RTX 6000 Ada GPUs) in Appendix~\ref{sec:implementation_detail_supp}, and report detailed computational costs (FLOPs) and memory footprints in Tabs.~\ref{tab:ablation} and \ref{tab:ablation_supp}, and inference latency in Tab.~\ref{tab:supp_latency}.
    
\item {\bf Code of ethics}
    \item[] Question: Does the research conducted in the paper conform, in every respect, with the NeurIPS Code of Ethics \url{https://neurips.cc/public/EthicsGuidelines}?
    \item[] Answer: \answerYes{} %
    \item[] Justification: We have carefully reviewed the NeurIPS Code of Ethics and confirm that our research and paper fully comply with all ethical guidelines.

\item {\bf Broader impacts}
    \item[] Question: Does the paper discuss both potential positive societal impacts and negative societal impacts of the work performed?
    \item[] Answer: \answerYes{} %
    \item[] Justification: Potential societal impacts are discussed in Appendix~\ref{sec:broader_impact}.
    
\item {\bf Safeguards}
    \item[] Question: Does the paper describe safeguards that have been put in place for responsible release of data or models that have a high risk for misuse (e.g., pre-trained language models, image generators, or scraped datasets)?
    \item[] Answer: \answerNA{} %
    \item[] Justification: This paper poses no such risks.

\item {\bf Licenses for existing assets}
    \item[] Question: Are the creators or original owners of assets (e.g., code, data, models), used in the paper, properly credited and are the license and terms of use explicitly mentioned and properly respected?
    \item[] Answer: \answerYes{} %
    \item[] Justification:  We clarify all the licenses of the codebase and datasets we used in Appendix~\ref{sec:implementation_detail_supp}.

\item {\bf New assets}
    \item[] Question: Are new assets introduced in the paper well documented and is the documentation provided alongside the assets?
    \item[] Answer: \answerNA{} %
    \item[] Justification: This paper does not release new assets.

\item {\bf Crowdsourcing and research with human subjects}
    \item[] Question: For crowdsourcing experiments and research with human subjects, does the paper include the full text of instructions given to participants and screenshots, if applicable, as well as details about compensation (if any)? 
    \item[] Answer: \answerNA{} %
    \item[] Justification: This paper does not involve crowdsourcing nor research with human subjects.

\item {\bf Institutional review board (IRB) approvals or equivalent for research with human subjects}
    \item[] Question: Does the paper describe potential risks incurred by study participants, whether such risks were disclosed to the subjects, and whether Institutional Review Board (IRB) approvals (or an equivalent approval/review based on the requirements of your country or institution) were obtained?
    \item[] Answer: \answerNA{} %
    \item[] Justification: This paper does not involve research with human subjects.

\item {\bf Declaration of LLM usage}
    \item[] Question: Does the paper describe the usage of LLMs if it is an important, original, or non-standard component of the core methods in this research? Note that if the LLM is used only for writing, editing, or formatting purposes and does \emph{not} impact the core methodology, scientific rigor, or originality of the research, declaration is not required.
    \item[] Answer: \answerNA{} %
    \item[] Justification: The core method development in this research does not involve LLMs as any important, original, or non-standard components.

\end{enumerate}

\end{document}